\documentclass[peerreview,12pt,draftcls]{IEEEtran}
\pdfoutput=1
\usepackage{mathrsfs}
\usepackage{graphicx}
\usepackage{float}
\usepackage{booktabs}
\usepackage{hyperref}
\usepackage{cite}

\usepackage{enumitem}
\usepackage{enumerate}
\usepackage{color}
\usepackage{psfrag}
\usepackage{subfigure}
\usepackage{amssymb}
\usepackage{bbding}
\usepackage{pifont}
\usepackage{threeparttable}
\usepackage{amsthm}
\usepackage{setspace}
\usepackage{epsfig}
\usepackage{pifont}
\usepackage{amsmath}
\usepackage{array}
\usepackage{multicol}
\usepackage{multirow}
\usepackage{diagbox}
\usepackage{pifont}
\usepackage{indentfirst}
\usepackage{amsfonts}
\usepackage{algorithm}
\usepackage{algorithmicx}
\usepackage{algpseudocode}
\usepackage{fancyhdr}
\usepackage{amscd}
\usepackage{bm}
\usepackage{fancyhdr}
\usepackage{enumerate}
\usepackage{color}
\usepackage{makecell}
\usepackage{extpfeil}

\renewcommand{\arraystretch}{1.2}
\newtheorem{remark}{Remark}
\newcommand{\tabincell}[2]{\begin{tabular}{@{}#1@{}}#2\end{tabular}}
\allowdisplaybreaks[4]
\IEEEoverridecommandlockouts

\begin{document}
	\title{\Large Learning Resource Allocation Policy: Vertex-GNN or Edge-GNN?}
		
\author{
	%\thanks{This work is supported by National Natural Science Foundation of China (NSFC) under Grant 61731002.}
	\IEEEauthorblockN{Yao Peng, Jia Guo and Chenyang Yang}
%	
%	\IEEEauthorblockA{Beihang University, Beijing, China\\Email: \{pengyyao,guojia,cyyang\}@buaa.edu.cn}%\vspace{-2.5em}
	}
	\maketitle
	\setcounter{page}{1}
	\thispagestyle{headings}
	
\vspace{-6mm}\begin{abstract}
Graph neural networks (GNNs) update the hidden representations of vertices (called Vertex-GNNs) or hidden representations of edges (called Edge-GNNs) by processing and pooling the information of neighboring vertices and edges and combining to exploit topology information. When learning resource allocation policies, GNNs cannot perform well if their expressive power is weak, i.e., if they cannot differentiate all input features such as channel matrices. In this paper, we analyze the expressive power of the Vertex-GNNs and Edge-GNNs for learning three representative wireless policies: link scheduling, power control, and precoding policies. We find that the expressive power of the GNNs depends on the linearity and output dimensions of the processing and combination functions. When linear processors are used, the Vertex-GNNs cannot differentiate all channel matrices due to the loss of channel information, while the Edge-GNNs can. When learning the precoding policy, even the Vertex-GNNs with non-linear processors may not be with strong expressive ability due to the dimension compression. We proceed to provide necessary conditions for the GNNs to well learn the precoding policy. Simulation results validate the analyses and show that the Edge-GNNs can achieve the same performance as the Vertex-GNNs with much lower training and inference time.

\vspace{-2mm}
\begin{IEEEkeywords}\vspace{-4mm}
	\emph{Vertex-GNN, Edge-GNN, Resource allocation, Precoding, Expressive power}	
\end{IEEEkeywords}
\end{abstract}

\vspace{-6mm}\section{Introduction}
Optimizing resource allocation such as link scheduling, power control, and precoding is important for improving the spectral efficiency of wireless communications. Various numerical algorithms have been proposed to solve these problems, such as the weighted minimal mean-square error (WMMSE) and the fractional programming algorithms \cite{WMMSE,FPLinQ_shen2017}, which are however with high computational complexity. To facilitate real-time implementation, fully-connected neural networks (FNNs) have been introduced to learn resource allocation policies, which are the mappings from environmental parameters (e.g., channels) to the optimized variables \cite{DNN_sun2017}. While significant research efforts have been devoted to intelligent communications, most existing works optimize resource allocation with FNNs or convolutional neural networks (CNNs). These DNNs are with high training complexity, not scalable, and not generalizable to problem size (say, the number of users). This hinders their practical use in dynamic wireless environments.

Encouraged by the potential for achieving good performance, reducing sample complexity and space complexity, as well as supporting scalability and size generalizability, graph neural networks (GNNs) have been introduced for learning resource allocation policies \cite{REGNN2020,LS2021Lee,2021Shen,review2021,LYsurvey,Het_PA2022GJ}. %\cite{REGNN2020,LS2021Lee,2021Shen,Het_PC/BF2021,Het_PA2022GJ,2022ZBC,phase_precoding2022,
%Access2021,GBLink2022,BF_RIS2021,Bipartite2022,IoT-supervised2022}.

These benefits of GNNs originate from exploiting topology information of graphs and permutation properties of wireless policies.
To embed the topology information, the hidden representations in each layer of a GNN are updated by first aggregating the information of neighboring vertices and edges
and then combining with the hidden representations in the previous layer, where the aggregation operation consists of processing and pooling. A policy can be learned either by a Vertex-GNN that updates the hidden representations of vertices or by an Edge-GNN that updates the hidden representations of edges, no matter if the optimization variables are defined on vertices or edges. To harness the permutation property, parameter-sharing should be introduced for each layer of a GNN. It has been shown in \cite{Het_PA2022GJ} that a GNN designed for learning a wireless policy will not perform well if the permutation property of the functions learnable by the GNN is mismatched with the policy.

However, even after satisfying the permutation property of a policy, a GNN may still not perform well due to the insufficient expressive power for learning the policy. The expressive power of a GNN is weak if the GNN cannot distinguish some pairs of graphs \cite{GNN2019power}. When learning a wireless policy, the graphs (i.e., the samples) that a GNN learns over are with different features. If a GNN maps different inputs (i.e., features) into the same output (called action), then the policy learned by the GNN may not achieve fairly good performance. Nonetheless, such a cause of the performance degradation has never been noticed until now.

\vspace{-2mm}
\subsection{Related Works} \vspace{-1mm}
\subsubsection{Vertex-GNNs and Edge-GNNs}
GNNs can either update the hidden representations of vertices (i.e., Vertex-GNNs) or edges (i.e., Edge-GNNs).
In the literature of machine learning, Vertex-GNNs were proposed for ``vertex-level tasks'' (say, vertex and graph classification) whose actions are defined on vertices \cite{GNNsurvey_Wu2021}. For ``edge-level tasks'' (say, edge classification and link prediction) whose actions are defined on edges, both Vertex-GNNs and Edge-GNNs have been designed. In early works as summarized in \cite{GNNsurvey_Wu2021}, edge-level tasks were learned by Vertex-GNNs with a read-out layer, which was used to map the representations of vertices
to the actions on edges. In \cite{CensNet2019,NeuralPS2021}, the edges of the original graph were transformed into vertices of a line-graph or hyper-graph. Then, the edge-level task on the original graph is equivalent to the vertex-level task on the line-graph or hyper-graph, which can be learned by Vertex-GNNs. In \cite{skeleton2020}, an Edge-GNN was designed for learning an edge-level task.

In the literature of intelligent communications, GNNs have been designed to learn the link scheduling  \cite{REGNN2020,LS2021Lee,IoT-supervised2022,GBLink2022}, power control \cite{2021Shen,Het_PC/BF2021} and power allocation \cite{Het_PA2022GJ} policies in device-to-device (D2D) communications and interference networks, the precoding policies in multi-user multi-input-multi-output (MIMO) system \cite{BF_RIS2021,2022ZBC,Bipartite2022,LSJ2023MGNN}, as well as the access point selection policy in cell-free MIMO systems \cite{Access2021}. Except \cite{2022ZBC,LSJ2023MGNN} where the optimized precoding matrices were defined as the actions on edges and hence Edge-GNNs were designed, the actions of all these works were defined on vertices and thereby Vertex-GNNs were designed.

\subsubsection{Structures of GNNs}
When using GNNs for a learning task, graphs need to be constructed and the structures of GNNs need to be designed or selected. For a resource allocation task, more than one graph can be constructed, say homogeneous graph or heterogeneous graph, and the action can be defined either on vertex or on edge. The structure of a GNN is captured by its update equation, which can either be vertex-representation update or edge-representation update, with a variety of choices for the processing, pooling, and combination functions. The commonly used pooling functions are sum, mean, and max functions, and the processing and combination functions can be either linear or nonlinear (say using FNNs). The three functions used in the GNNs for wireless policies are provided in Table \ref{table: MPGNN}, which were designed empirically without explaining the rationality.

\begin{table}[!htb]
	\centering
	\vspace{-5mm}
	\small
	\caption{Processing, pooling and combination functions of GNNs for resource allocation.}\label{table: MPGNN}\vspace{-3mm}
	\begin{tabular}{|c|c|c|c|}
		\hline
		\textbf{Processing Function}&\textbf{Pooling Function}&\textbf{Combination Function}&\textbf{Literature}\\
		\hline
		Linear function&sum&FNN without hidden layer&\cite{LS2021Lee,Het_PA2022GJ,2022ZBC,LSJ2023MGNN}\\
		\hline
		FNN&max&FNN without hidden layer&\cite{Access2021}\\
		\hline
		FNN&max,mean&FNN&\cite{GBLink2022}\\
		\hline
		FNN&max&FNN&\cite{2021Shen,Het_PC/BF2021,BF_RIS2021}\\
		\hline
		FNN&sum&FNN&\cite{Bipartite2022}\\
		\hline
		FNN&mean&FNN&\cite{BF_RIS2021}\\
		\hline
		CNN &mean&FNN&\cite{IoT-supervised2022}\\
	%	\hline
	%	FNN&sum&Recurrent Neural Network&\cite{Detection2020}\\
		\hline
	\end{tabular}
\end{table}
\vspace{-5mm}

GNNs can be designed to satisfy permutation properties, which widely exist in wireless policies \cite{SCJ,LSJ2023MGNN}. According to the problem, a policy can be with different properties such as one-dimension (1D)-permutation equivariance (PE), two-dimension (2D)-PE, and joint-PE \cite{Het_PA2022GJ} property. A GNN learning over a homogenous graph can automatically learn the policies with 1D-PE property if the read-out layer (if needed) also satisfies the 1D-PE property, since the permutation of the vertices in the graph does not affect the output of the GNN. If a GNN is designed for learning a policy with 2D-PE, joint-PE, or more complicated PE property, then the parameter-sharing in the update equation of the GNN should be judiciously designed, as detailed in \cite{LSJ2023MGNN}. In \cite{REGNN2020,2021Shen,Het_PA2022GJ,2022ZBC,LSJ2023MGNN}, the PE properties of their considered wireless tasks were analyzed, and the GNNs were designed to satisfy the same properties. A GNN with matched PE property to a policy is possible to be sample efficient, scalable, and size generalizable, but the GNN may still not perform well due to insufficient expressive power to the policy.
% Besides, some previous works have validated that the GNN has the size generalization ability via simulations \cite{REGNN2020,2021Shen,Het_PA2022GJ,LS2021Lee,IoT-supervised2022,GBLink2022,Het_PC/BF2021,BF_RIS2021,Bipartite2022}.

\subsubsection{Expressive power of GNNs}
The parameter-sharing in GNNs restricts their expressive power \cite{expressive-GNN-Li2022}. In the literature of machine learning, the expressive power of Vertex-GNNs has been investigated for vertex classification, link prediction, and graph classification tasks, where the samples for training and testing a GNN are graphs with different topologies \cite{expressive-GNN-Li2022,GNN2019power,survey2020expressive,jegelka2022theory}.
The expressive power is characterized by the capability to distinguish non-isomorphic graphs.\footnote{
According to the definition in  \cite{expressive-GNN-Li2022}, we know that two graphs are isomorphic if a graph can be obtained by reordering the vertices in another graph, otherwise they are non-isomorphic.}

The 1-dimensional Weisfeiler-Lehman (1-WL) test is a widely used algorithm to distinguish non-isomorphic graphs, which consists of an injective aggregation process. Both the 1-WL test and Vertex-GNNs iteratively update the representation of a vertex by aggregating the representations of its neighboring vertices. Inspired by such a finding, the expressive power of Vertex-GNNs was characterized by whether their aggregating functions were injective in \cite{GNN2019power}. To build a GNN that is as powerful as the 1-WL test, a graph isomorphic network (GIN) was developed in \cite{GNN2019power}, which is a Vertex-GNN whose processing and combination functions are FNNs and the pooling function is sum function.
Under the assumption that the features of vertices were from a countable set, e.g., all vertices are with identical feature, the aggregating functions of the GIN were proved to be injective.
%, which was proved to be as powerful as the 1-WL test.
When replacing the processing functions in the GIN with linear functions or replacing the pooling functions with mean or max function, the aggregating functions were proved to be non-injective,
hence the resultant GNN is with weaker expressive power than the GIN.
It was empirically shown that the less powerful Vertex-GNNs perform worse than the GIN on a number of graph classification tasks.
% where the features of vertices are from a countable set.

In \cite{survey2020expressive,expressive-GNN-Li2022,jegelka2022theory},
the expressive power of Vertex-GNNs and the techniques to improve their expressive power were reviewed.
%According to the results in \cite{GNN2019power}, the 1-WL algorithm is the upper bound of the expressive power of GNNs.
%By comparing the updating procedure of the GNN with the 1-WL test as shown in \cite{GNN2019power},
%since the aggregating functions in the GNN are not necessarily injective,
%the GNN does not surpass the expressive power of the 1-WL test.
According to the analyses in \cite{GNN2019power}, a GNN as powerful as the 1-WL test is with the strongest expressive power among all Vertex-GNNs. Since the 1-WL test cannot distinguish some non-isomorphic graphs, e.g., $k$-regular graphs with the same size and same vertex features, the Vertex-GNNs also cannot distinguish them.
%However, there are still non-isomorphic graphs that the GNN fails to distinguish, e.g., $k$-regular graphs with the same size and same vertex features.
To design Vertex-GNNs with stronger expressive power for these graphs, one can provide each vertex a unique feature to make vertices more distinguishable \cite{expressive-GNN-Li2022,jegelka2022theory} or use a high-order GNN that updates the representation of $k$-tuple of vertices to maintain more structural information of the graph \cite{survey2020expressive,expressive-GNN-Li2022,jegelka2022theory}. However, these techniques incur computational costs.
%, which indicates their capabilities of distinguishing whether or not two graphs are isomorphic \cite{GNN2019power,survey2020expressive}.
%\cite{GNN2019power,Aggreration2020Cambridge}.

Due to the considered tasks in \cite{expressive-GNN-Li2022,jegelka2022theory,GNN2019power,survey2020expressive} and the references therein, these works did not consider the features of edges. When learning wireless policies, the edges of the constructed graphs may have features, and the graphs with the same topology but with different features usually corresponds to different actions. The 1-WL test was proposed to distinguish non-isomorphic graphs without edge features that are different from the graphs for wireless problems, hence the analyses for the expressive power of GNNs in \cite{expressive-GNN-Li2022,jegelka2022theory,GNN2019power,survey2020expressive} are not applicable to most wireless problems. The expressive power of a GNN for learning wireless policies is captured by its capability to distinguish input features (e.g., channel matrices), which has never been investigated so far.

\vspace{-3mm}\subsection{Motivation and Major Contributions}\vspace{-1mm}
In this paper, we strive to analyze the impact of the structure of a GNN for optimizing resource allocation or precoding on its expressive power, aiming to provide useful insight into the design of GNNs for learning wireless policies.

We take the link scheduling and power control problems in D2D communications and the precoding problem in multi-user MIMO system as examples, each representing a class of policies. The link scheduling and power control policies are the mappings from the channel matrices to the optimized vectors in a lower dimensional space, where the optimization variables are respectively discrete and continuous scalars. The precoding policy is the mapping from the channel matrix to the optimized precoding matrix, where the optimization variables are vectors.
%To demonstrate that a wireless policy can be learned by both Vertex-GNN and Edge-GNN, we consider different graphs for each policy.
To demonstrate that the graph constructed for a wireless policy is not unique and the policy can be learned by both Vertex-GNN and Edge-GNN, we consider different graphs and GNNs for each policy.
%When using a Vertex-GNN for learning the precoding policy,
%%the number of the vertices are less than the number of optimized variables, hence
%the low dimensional vertex representation will restrict its expressive power.
%, which is also applicable to improve the expressive power of Vertex-GNNs for learning wireless policies. Unlike the Vertex-GNNs, the Edge-GNNs with linear processing functions can differentiate different channel matrices. For the high dimensional problem, such as the precoding problem, when using the Vertex-GNNs, the number of the vertexes are less than the number of optimized variables. The low representation dimensions of vertexes can restrict the expressive power of Vertex-GNNs with non-linear processing functions.

To the best of our knowledge, this is the first attempt to analyze the expressive power of the GNNs for learning wireless policies. The major contributions are summarized as follows.
\begin{itemize}
\item We find that Vertex-GNNs with linear processing functions cannot differentiate specific channel matrices due to the loss of channel information after aggregation, which leads to poor learning performance. Their expressive power can be improved by introducing non-linear processers. By contrast, the update equations of Edge-GNNs with linear processors do not incur the information loss.
\item We find that the output dimensions of the processing and combination functions also restrict the expressive power of the Vertex-GNN for learning the precoding policy, in addition to the linear processors. We provide a lower bound on the dimensions for a Vertex-GNN or Edge-GNN without the dimension compression.
\item We validate the analyses and compare the performance of Vertex-GNNs and Edge-GNNs for learning the three policies via simulations. Our results show that both the training time and inference time of the Edge-GNNs with linear processors are much lower than the Vertex-GNNs with FNN-processors to achieve the same performance.
\end{itemize}

The rest of this paper is organized as follows.
In section \ref{sec: system model}, we introduce three resource allocation policies. In section \ref{sec: GNN}, we present Vertex-GNNs and Edge-GNNs to learn the policies over the directed homogeneous or undirected heterogeneous graph, and analyze the expressive power of the GNNs with linear processors. In section \ref{sec:Influence}, we analyze the impacts of the linearity and output dimensions of processing and combination functions on the expressive power of GNNs. In sections \ref{sec: simulation} and  \ref{sec: conclusion}, we provide simulation results and conclusions.

\emph{Notations:}
$(\cdot)^{\sf T}$ and $(\cdot)^{\sf H}$ denote transpose and Hermitian transpose, respectively. $|\cdot|$ denotes the absolute value of a real number or the magnitude of a complex number. ${\sf Tr}(\cdot)$ denotes the trace of a matrix. ${\bf X}=[x_{ij}]_{m\times n}$ denotes a matrix with $m$ rows and $n$ columns where $x_{ij}$ is the element in the $i$th row and the $j$th column. $\Vert{\bf X}\Vert\triangleq\sum_{i=1}^{m}\sum_{j=1}^{n}x_{ij}$, and $|{\bf X}|_{max}\triangleq \max_{i=1}^{m}\max_{j=1}^{n}|x_{ij}|$. $\bf \Pi$, $\bf \Pi_1$ and $\bf \Pi_2$ denote permutation matrices.
%${\rm Re}\{\cdot\}$ and ${\rm Im}\{\cdot\}$ respectively denote the real and imaginary parts of a complex number,
$\mathbb R$, $\mathbb C$, and $\mathbb I$ denote the sets of real, complex, and integer numbers, respectively. ${\mathbb R}^{n}$ denotes $n$-dimensional vector space.

\vspace{-2mm}
\section{Resource Allocation Problems and Policies}\label{sec: system model} \vspace{-2mm}

In this section, we present three representative resource allocation problems.

\vspace{-1mm}
\subsubsection{Link scheduling}
\begin{figure}[!htb]
	\centering
	\begin{minipage}[t]{0.42\linewidth}	
		\subfigure[D2D communication system]{
			\includegraphics[width=\textwidth]{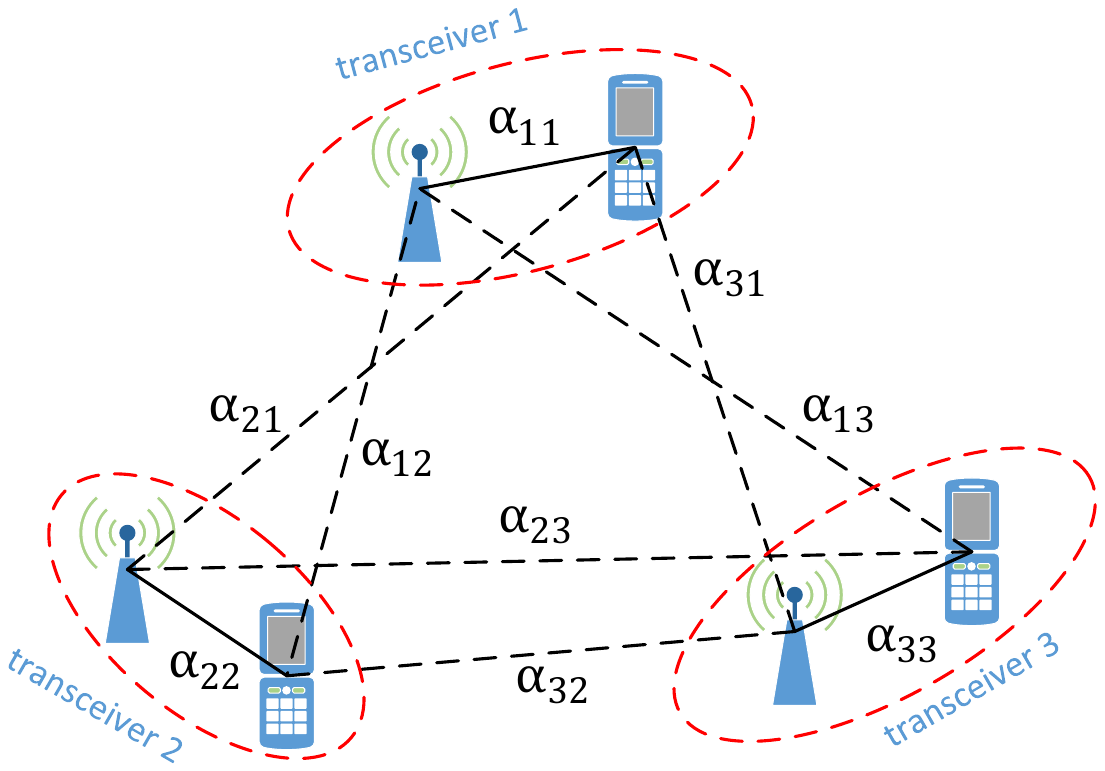}}
	\end{minipage}
	\begin{minipage}[t]{0.35\linewidth}	
		\subfigure[Multi-user MIMO system]{
			\includegraphics[width=\textwidth]{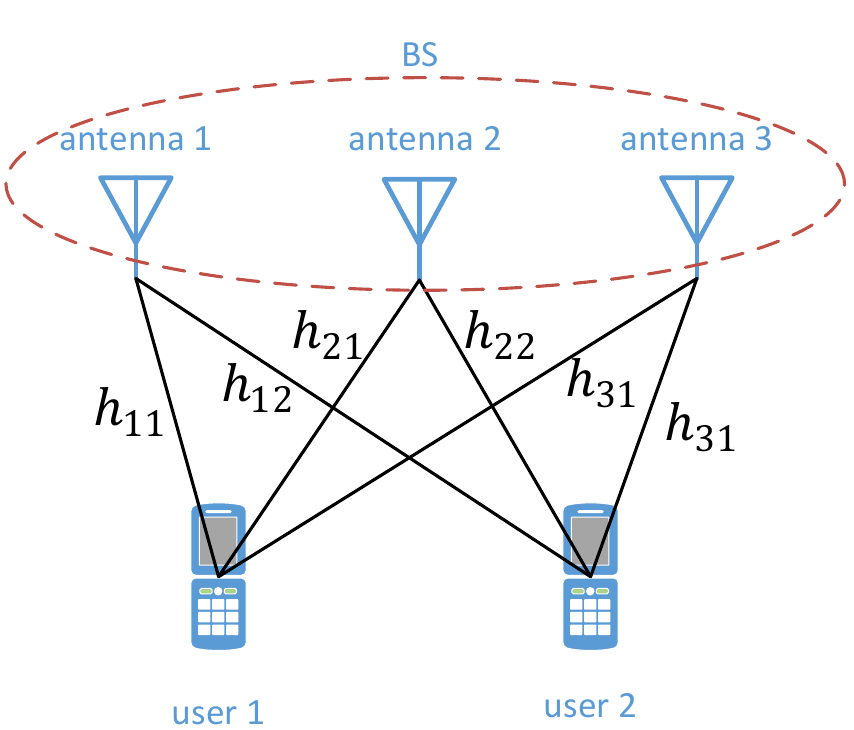}}
	\end{minipage}
	\vspace{-2mm}
	\caption{The considered two systems. (a): $K=3$, solid and dashed lines represent the communication and interference links, respectively. (b): $N=3$ and $K=2$.}\label{fig:system-structure}
	\vspace{-4mm}
\end{figure}

Consider a D2D communication system with $K$ pairs of transceivers. Every transmitter sends data to a receiver, and all the transmissions share the same spectrum. Hence, there exist interference among the transceiver pairs, as illustrated in Fig. \ref{fig:system-structure}(a). To coordinate the interference, not all the D2D links are activated.  A {link scheduling} problem that maximizes the sum rate of active links is \cite{FPLinQ_shen2017,LS2021Lee,IoT-supervised2022},
\begin{equation}\label{eq: max sum-rate}
   \max_{x_1,\cdots,x_K} \sum_{k=1}^K  \log_2\left(1+\frac{x_k p_k \alpha_{kk}} {\sum_{j=1,j \neq k}^K x_j p_j \alpha_{jk}+\sigma_0^2}\right) ~~
   {\rm s.t.} ~~ x_k\in\{0,1\}, k=1,\cdots,K,
\end{equation}
where $x_k$ is the active state of the $k$th D2D link, $x_k=1$ when the $k$th D2D link is active, $x_k=0$ otherwise, $\alpha_{jk}$ is the composite large- and small-scale channel gain from the $j$th transmitter to the $k$th receiver, $p_k$ is the power of the $k$th transmitter, and $\sigma_0^2$ is noise power.

The link scheduling policy is denoted as ${\bf x}^{*}= F_{\sf ls}({\bm \alpha})$,
where ${\bf x}^{*}= [x_1^*,\cdots,x_K^*]^{\sf T}$ is the optimized solution of the problem in \eqref{eq: max sum-rate} for a given channel matrix ${\bm \alpha}=[\alpha_{ij}]_{K\times K}$, and $F_{\sf ls}(\cdot)$ is a function that maps ${\bm \alpha} \in \mathbb{R}^{K\times K}$ into ${\bf x}^* \in \mathbb{I}^{K\times 1}$. This policy is joint-PE to ${\bm \alpha}$ \cite{Het_PA2022GJ}, i.e., ${\bf \Pi}^{\sf T}{{\bf x}^*}=F_{\sf ls}({\bf \Pi}^{\sf T}{\bm \alpha}{\bf \Pi})$.

\subsubsection{Power control}\vspace{-1mm}
The interference in the D2D system can also be coordinated by adjusting the transmit power of every transmitter. The {power control} problem that maximizes the sum rate under power constraint is \cite{DNN_sun2017,2021Shen,Het_PC/BF2021},
\begin{equation}\label{eq: max sum-ratepc}
   \max_{p_1,\cdots,p_K} \sum_{k=1}^K  \log_2\left(1+\frac{p_k \alpha_{kk}} {\sum_{j=1,j \neq k}^K p_j \alpha_{jk}+\sigma_0^2}\right) ~~
   {\rm s.t.} ~~ 0\leq p_k \leq P_{max}, k=1,\cdots,K,
\end{equation}
where $P_{max}$ is the maximal transmit power.

The power control policy is denoted as ${\bf p}^{*}= F_{\sf pc}({\bm \alpha})$,
where ${\bf p}^{*}= [p_1^*,\cdots,p_K^*]^{\sf T}$ is the optimized solution of the problem in \eqref{eq: max sum-ratepc} for a given channel matrix ${\bm \alpha}$, and $F_{\sf pc}(\cdot)$ is a function  that maps ${\bm \alpha} \in \mathbb{R}^{K\times K}$ into ${\bf p}^* \in \mathbb{R}^{K\times 1}$.
This policy is also joint-PE to ${\bm \alpha}$ \cite{2021Shen}, i.e., ${\bf \Pi}^{\sf T}{{\bf p}^*}=F_{\sf pc}({\bf \Pi}^{\sf T}{\bm \alpha}{\bf \Pi})$.

\vspace{-4mm}

%\vspace{-5mm}
\subsubsection{Precoding}\vspace{-1mm}
Consider a multi-user multi-antenna system, where a base station (BS) equipped with $N$ antennas transmits to $K$ users each with a single antenna, as shown in Fig. \ref{fig:system-structure}(b). The {precoding} problem that maximizes the sum rate of users subject to power constraint is \cite{2022ZBC,Bipartite2022}
\begin{equation}\label{eq: max sum-rate precoding}
   \max_{{\bf v}_1,\cdots,{\bf v}_K} \sum_{k=1}^K  \log_2\left(1+\frac{|{\bf h}_k^{\sf H}{\bf v}_k|^2} {\sum_{j=1,j \neq k}^K |{\bf h}_k^{\sf H}{\bf v}_j|^2+\sigma_0^2}\right) ~~
   {\rm s.t.} ~~ {\rm Tr}({\bf V}{\bf V}^{\sf H})\leq P_{max},
\end{equation}
where ${\bf V}=[{\bf v}_1,\cdots,{\bf v}_K]=[v_{nk}]_{N\times K}$, ${\bf v}_k\in \mathbb{C}^{N\times 1}$ is the precoding vector for the $k$th user, and ${\bf h}_k\in \mathbb{C}^{N\times 1}$ is the channel vector from the BS to the $k$th user.

Denote the precoding policy as ${\bf V}^{*}= F_{\sf p}({\bf H})$, where ${\bf V}^{*}$ is the optimized solution of the problem in \eqref{eq: max sum-rate precoding} for a given channel matrix ${\bf H}=[{\bf h}_1,\cdots,{\bf h}_K]=[h_{nk}]_{N\times K}$, and $F_{\sf p}(\cdot)$ is a function  that maps ${\bf H} \in \mathbb{C}^{N\times K}$ into ${\bf V}^{*} \in \mathbb{C}^{N\times K}$.
The precoding policy is 2D-PE to ${\bf H}$ \cite{2022ZBC}, i.e., ${\bm \Pi}_1^{\sf T}{\bf V}^*{\bm \Pi}_2=F_{\sf p}({\bm \Pi}_1^{\sf T}{\bf H}{\bm \Pi}_2)$.

The link scheduling problem is a combinatorial optimization problem, which can be solved by using iterative algorithms \cite{FPLinQ_shen2017} or exhaustive searching.
Both the problems in \eqref{eq: max sum-ratepc} and \eqref{eq: max sum-rate precoding} are non-convex, which can be solved by numerical algorithms such as the WMMSE algorithm \cite{WMMSE}.

\vspace{-3mm}
\section{GNNs for Learning the Policies}\label{sec: GNN}\vspace{-1mm}
In this section, we introduce several GNNs to learn the three policies. Since constructing appropriate graphical models is the premise of applying GNNs, we first introduce graphs for each policy. Then, we introduce Vertex-GNN and Edge-GNN. Finally, we analyze the expressive power of the GNNs with simple pooling, processing, and combination functions.

\vspace{-5mm}
\subsection{Homogeneous and Heterogeneous Graphs}\vspace{-2mm}
A graph consists of vertices and edges. Each vertex or edge may be associated with features and actions. Learning a resource allocation policy over a graph is to learn the actions defined on vertices or edges based on their features. The inputs and outputs of a GNN are the features and actions of a graph, respectively.

A graph may consist of vertices and edges that belong to different types.
If a graph consists of vertices or edges with more than one type, then it is a heterogeneous graph. Otherwise, it is a homogeneous graph, which can be regarded as a special heterogeneous graph.

More than one graph can be constructed for a resource allocation problem.
%We only describe the topologies of the graphs in this subsection. The features and actions of the graphs for specific GNNs are defined in the next subsection.
\subsubsection{Link scheduling/Power control}  For learning the link scheduling and power control policies, the topologies and features of the graphs and the structures of the GNNs are the same, and only the actions of the GNNs and the loss functions for training the GNNs are different.
Hence, we focus on the GNNs for learning the link scheduling policy in the sequel.

\begin{figure}[!htb]
	\centering
	\begin{minipage}[t]{0.28\linewidth}	
		\subfigure[Homogeneous graph, ${\cal G}_{\sf{ls}}^{\sf hom}$]{
			\includegraphics[width=\textwidth]{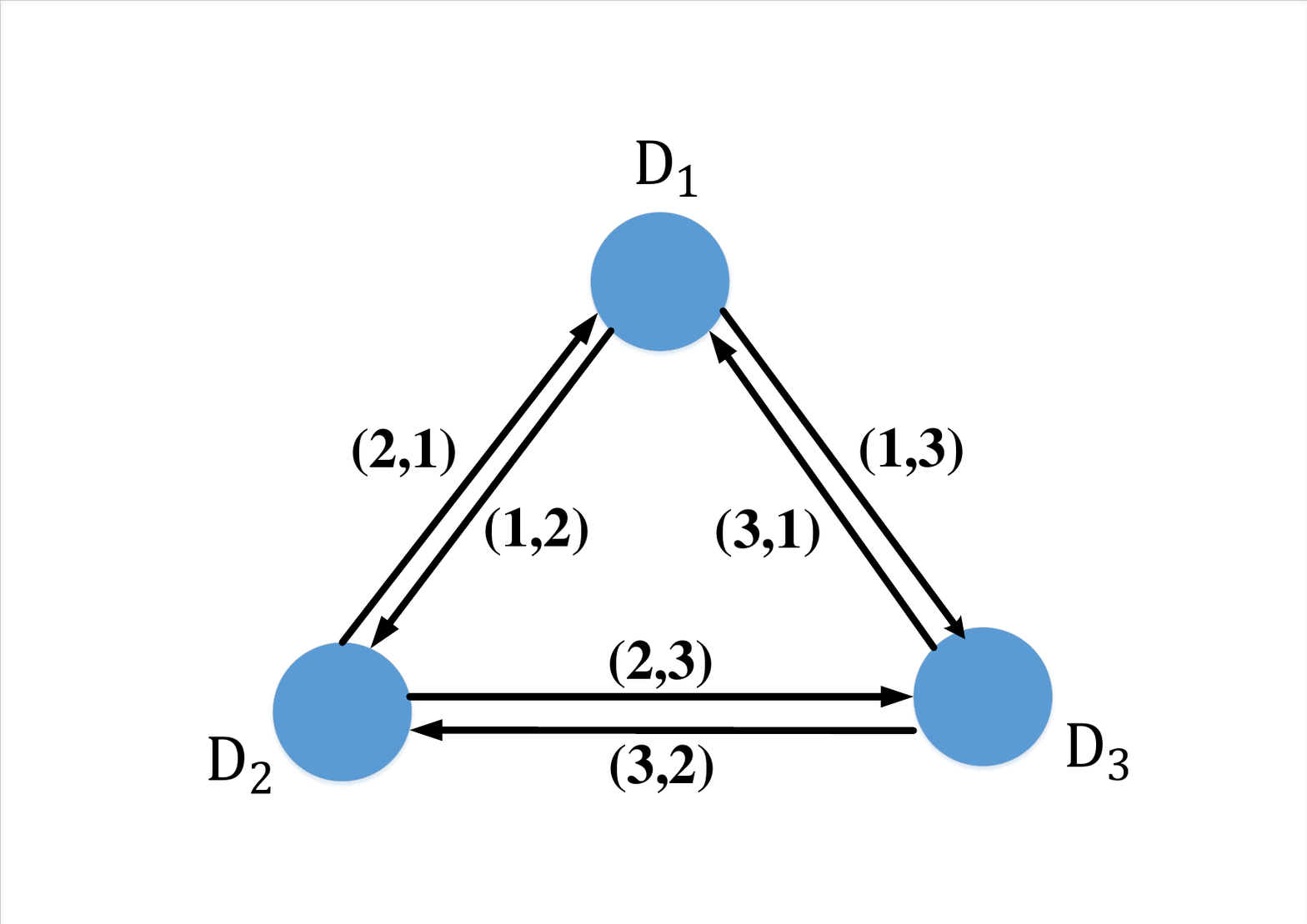}}
	\end{minipage}
	\begin{minipage}[t]{0.28\linewidth}	
		\subfigure[Heterogeneous graph, ${\cal G}_{\sf{ls}}^{\sf{het}}$]{
			\includegraphics[width=\textwidth]{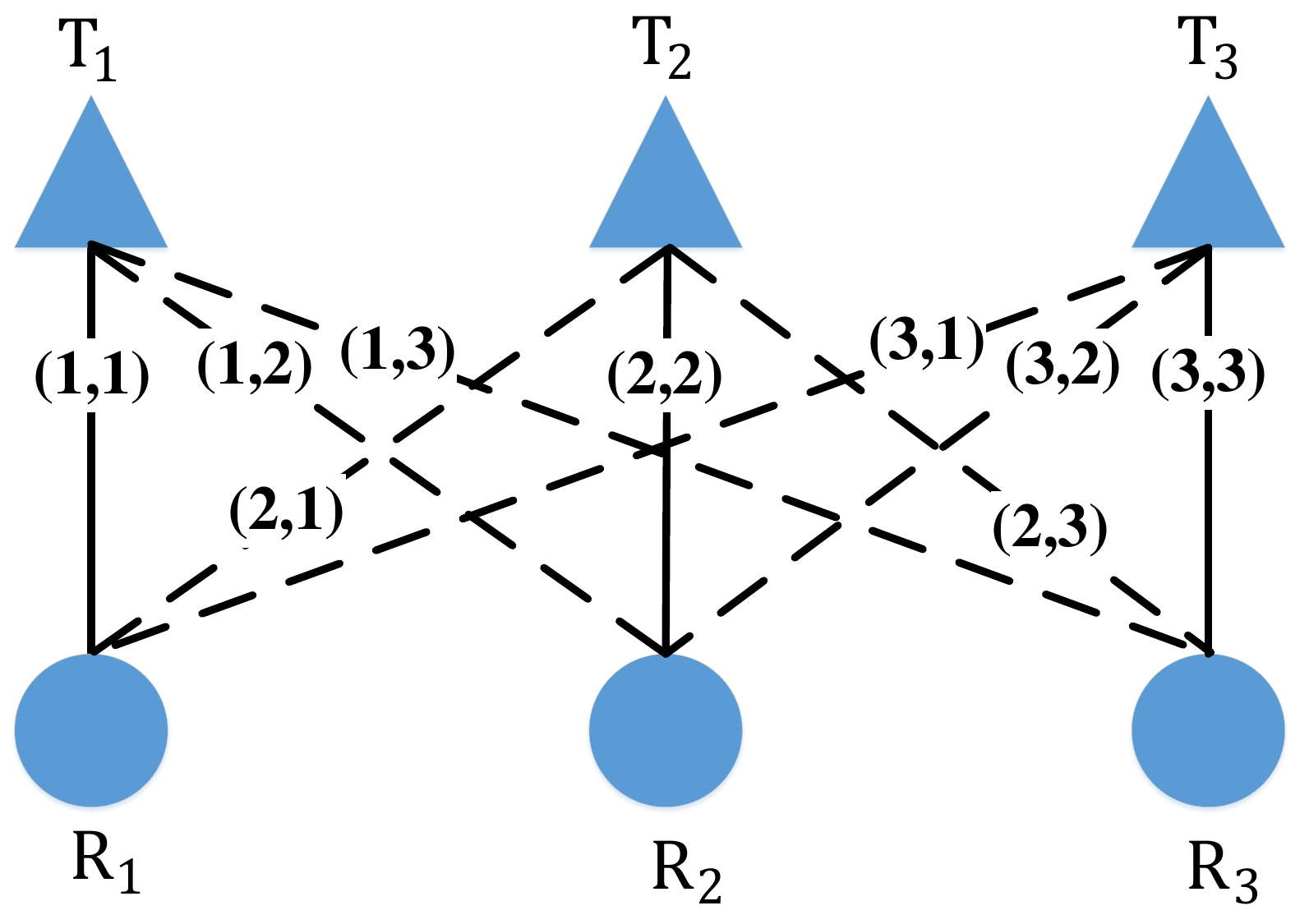}}
	\end{minipage}
	\begin{minipage}[t]{0.3\linewidth}	
		\subfigure[Heterogeneous graph, ${\cal G}_{\sf{p}}^{\sf{het}}$]{
			\includegraphics[width=\textwidth]{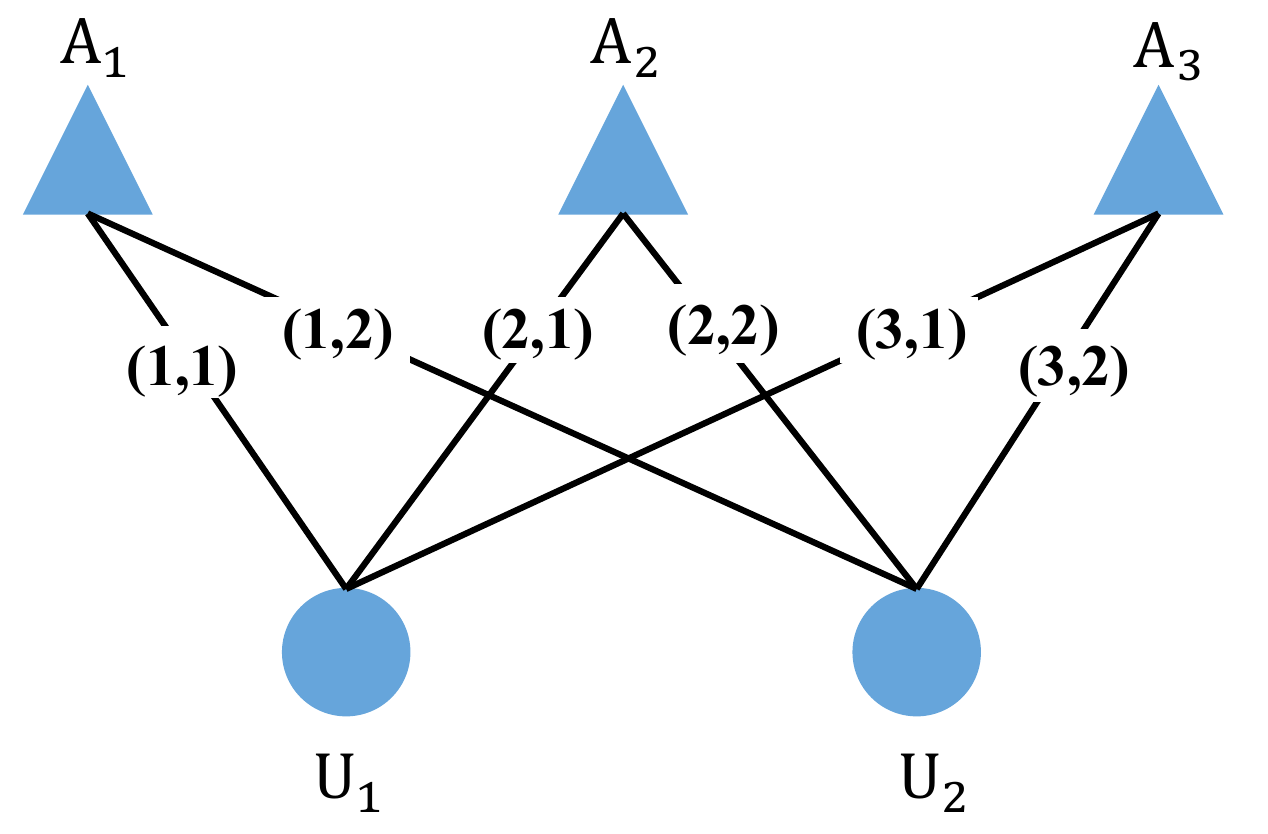}}
	\end{minipage}
	\vspace{-2mm}
	\caption{The topologies of the graphs. (a) and (b): link scheduling/power control, $K=3$. (c): precoding, $N=3$, $K=2$. The circles and triangles represent vertices, solid and dashed lines represent the two types of edges.}\label{fig:graph}
	\vspace{-4mm}
\end{figure}

In \cite{LS2021Lee, IoT-supervised2022}, a directed homogeneous graph ${\cal G}_{\sf ls}^{\sf hom}$, as shown in Fig. \ref{fig:graph}(a),  was constructed for learning the link scheduling policy. In ${\cal G}_{\sf{ls}}^{\sf hom}$, each D2D pair is a vertex, and the interference links among the D2D pairs are directed edges. Denote the $i$th vertex as $D_i$, and the edge from $D_i$ to $D_j$ as edge $(i,j)$.
The feature of vertex $D_i$ is $\alpha_{ii}$, and the feature of edge $(i,j)$ is $\alpha_{ij}$. The features of all vertices and edges can be represented as $\bm \alpha$. The action of vertex $D_i$ is the active state of the $i$th link $x_i$. In \cite{2021Shen}, the graph constructed for optimizing power control only has one difference from ${\cal G}_{\sf {ls}}^{\sf hom}$: the action of vertex $D_i$ is $p_i$.

We can also construct an undirected heterogeneous graph ${\cal G}_{\sf{ls}}^{\sf{het}}$ as shown in Fig. \ref{fig:graph}(b) for learning the link scheduling policy. In ${\cal G}_{\sf{ls}}^{\sf{het}}$, there are two types of vertices and two types of edges. Each transmitter and each receiver are respectively defined as a transmitter vertex and a receiver vertex (respectively called $\rm tx$ vertex and $\rm rx$ vertex for short), and the link between them is an undirected edge. Denote the $i$th $\rm tx$ vertex and  the $i$th $\rm rx$ vertex as $T_i$ and $R_i$, respectively, and the edge between $T_i$ and $R_j$ as edge $(i,j)$. Edge $(i,i)$ is referred to as {signal edge}, and edge $(i,j)$ $(i\neq j)$ is referred to as {interference edge} (respectively called $\rm sig$ edge and $\rm int$ edge for short).
The vertices have no features. The feature of edge $(i,j)$ is $\alpha_{ij}$, and the features of all the edges can be represented as $\bm \alpha$. The active state $x_i$ can either be defined as the action of vertex $T_i$ or the action of $\rm sig$ edge $(i,i)$.

\vspace{-5mm}

\subsubsection{Precoding}\vspace{-1mm} In \cite{2022ZBC}, the precoding policy was learned over a heterogeneous graph ${\cal G}_{\sf{p}}^{\sf{het}}$ as shown in Fig. \ref{fig:graph}(c). In ${\cal G}_{\sf{p}}^{\sf{het}}$, there are two types of vertices and one type of edges. Each antenna at the BS and each user are respectively antenna vertex and user vertex, and the link between them is an undirected edge. Denote the $i$th antenna vertex and the $j$th user vertex as $A_i$ and $U_j$, respectively, and the edge between $A_i$ and $U_j$ as edge $(i,j)$.
The vertices have no features. The feature of edge $(i,j)$ is $h_{ij}$, and the features of all the edges can be represented as $\bf H$. The action $v_{ij}$ is defined on edge $(i,j)$.

One can also construct a homogeneous graph for learning the precoding policy in a similar way as in \cite{BF_RIS2021}. In particular, each user (say the $k$th user) and all antennas at the BS are defined as a vertex, ${\bf h}_k$ and ${\bf v}_k$ are the feature and action of the vertex, respectively. Every two vertices are connected by an edge, which has no feature and action. However, a GNN learning over such a graph is only permutation equivariant to users, losing the property of permutation equivariance to antennas and hence incurring higher training complexity.

\vspace{-4mm}
\subsection{Vertex-GNN and Edge-GNN}%\vspace{-3mm}
GNNs can be classified into Vertex-GNNs and Edge-GNNs, which respectively update the hidden representations of vertices and edges. Each class of GNNs can learn over either homogeneous or heterogeneous graphs. The directed homogeneous graph can be transformed into an undirected heterogeneous graph by regarding the edges of different directions as two types of undirected edges and the two vertices connected by a directed edge as two types of vertices.\footnote{The basic idea is to transform the direction information in a directed homogeneous graph into the type information in an undirected heterogeneous graph, which is applicable to any problem.} Hence, we focus on undirected heterogeneous graph in the following (called heterogeneous graph for short).

For conciseness, we take the GNNs for learning the link scheduling policy over the heterogeneous graph ${\cal G}_{\sf ls}^{\sf het}$ in Fig. \ref{fig:graph}(b) as an example. We discuss the GNN for learning the link scheduling policy over the converted heterogeneous graph from the directed homogeneous graph and the GNN for learning the precoding policy in remarks.

\subsubsection{Vertex-GNN} \label{sec:Vertex-GNNs}
\begin{figure}[!htb]
	\centering
	\begin{minipage}[t]{0.4\linewidth}	
		\subfigure[Neighboring vertices and edges of $T_1$]{
			\includegraphics[width=0.8\textwidth]{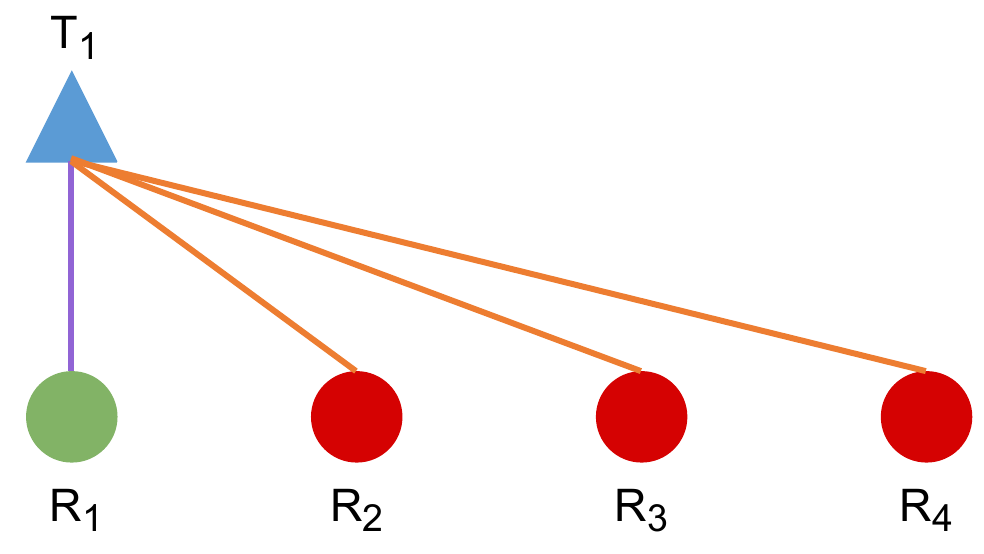}}
	\end{minipage}
	\begin{minipage}[t]{0.4\linewidth}	
		\subfigure[Neighboring vertices and edges of $R_1$]{
			\includegraphics[width=0.8\textwidth]{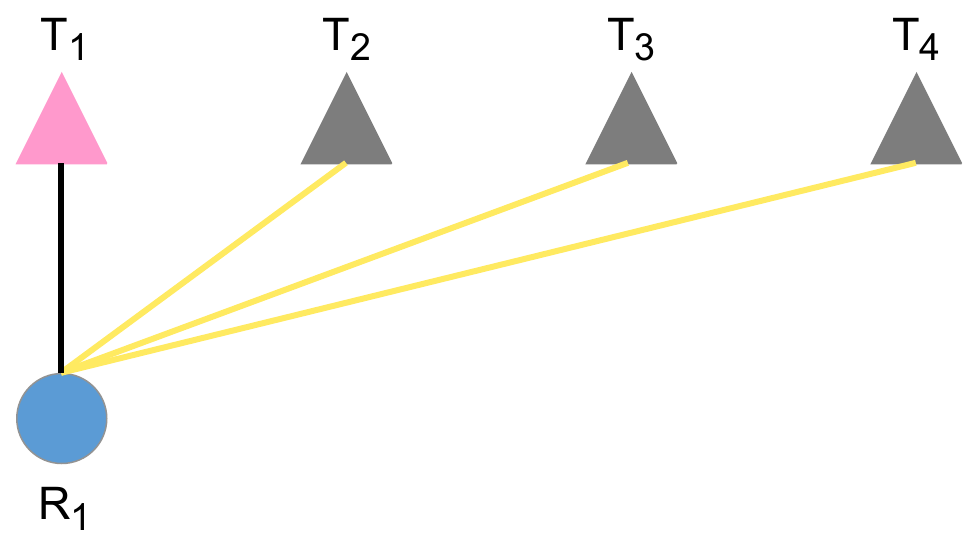}}
	\end{minipage}\\
	\begin{minipage}[t]{0.4\linewidth}	
		\subfigure[Neighboring edges of edge $(1,1)$]{
			\includegraphics[width=0.8\textwidth]{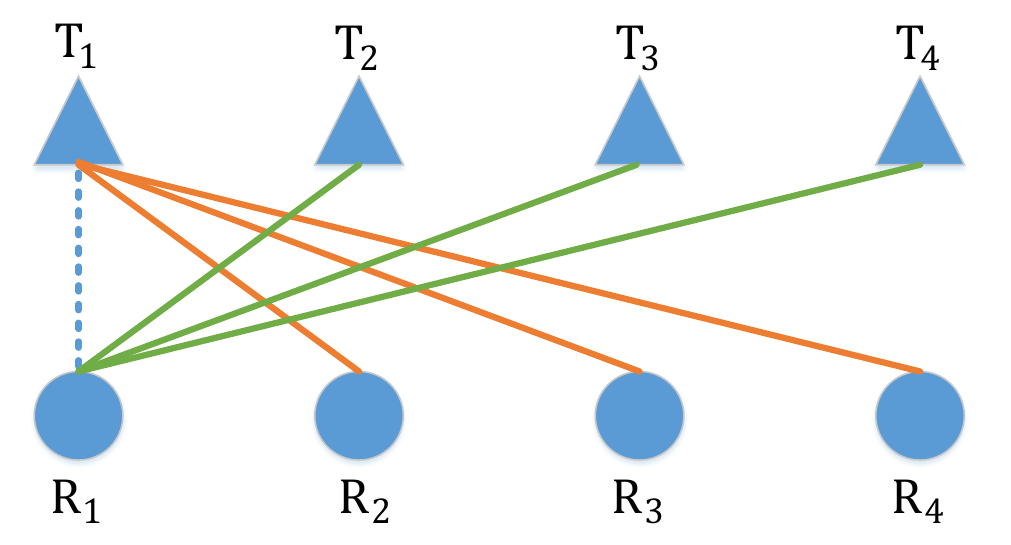}}
	\end{minipage}
	\begin{minipage}[t]{0.4\linewidth}	
		\subfigure[Neighboring edges of edge $(1,2)$]{
			\includegraphics[width=0.8\textwidth]{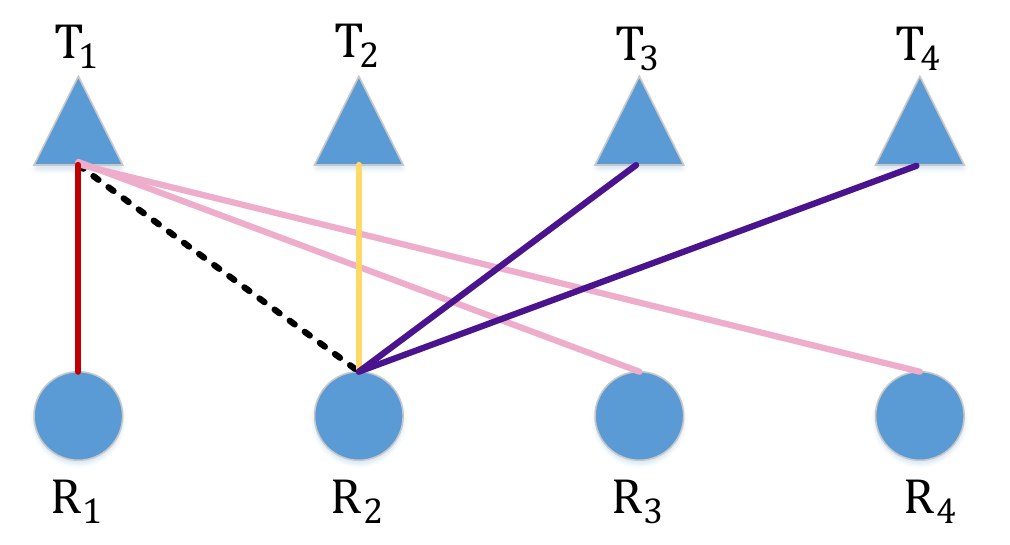}}
	\end{minipage}
    \vspace{-1mm}
	\caption{Vertex-GNN (a) and (b) (When updating the representation of a vertex in blue color, the information of the vertices and edges with the same color is aggregated with the same weight.) and Edge-GNN (c) and (d) (When updating  the representation of an edge with dashed lines, the information of the edges with the same color is aggregated with the same weight.)}\label{fig:Vertex-HetGNN}
	\vspace{-12mm}
\end{figure}

In Vertex-GNN, the hidden representation of each vertex is updated in each layer, by first aggregating information from its neighboring vertices and edges, and then combining the aggregated information with its own information in the previous layer.
For each vertex, its neighboring vertices are the vertices connected to it with edges, and its neighboring edges are the edges connected to it. As illustrated in Fig. \ref{fig:Vertex-HetGNN}(a)(b) where $K=4$,
for $T_1$, $R_1 \sim R_4$ are its neighboring vertices and edge (1,1) $\sim$ edge (1,4) are its neighboring edges, while for $R_1$, $T_1 \sim T_4$ are its neighboring vertices and edge (1,1) $\sim$ edge (4,1) are its neighboring edges.
%for transmitter vertex $T_i$ in Fig. \ref{fig:graph}(b), $R_i$ is its neighboring vertex connected by its neighboring signal edge $(i,i)$, and $R_j (j\neq i)$  is its neighboring vertex connected by its neighboring interference edge $(i,j)$, as illustrated in Fig. \ref{fig:Vertex-HetGNN}(a).
%For receive vertex $R_i$, $T_i$ is its neighboring vertex connected by its neighboring signal edge $(i,i)$, and  $T_j$ is its neighboring vertex connected by its neighboring interference edge $(j,i)$, as illustrated in Fig. \ref{fig:Vertex-HetGNN}(b).
%Since edge $(i,i)$ and edge $(j,i)$ are of different types, the weight matrices for processing the information from $T_i$, edge $(i,i)$ are different from the weight matrices for processing the information from $T_j$ and the edge $(j,i)$.
\vspace{-1mm}

For the Vertex-GNN learning over a heterogeneous graph,
the hidden representation of the $i$th vertex with the $\tau_i$th type in the $l$th layer, ${\bf d}_{i,\tau_i}^{(l)}$, is updated as follows\cite{zhang_heterogeneous_2019},
\begin{equation} \label{eq: node-GNN}
	\small
	\begin{split}
   \textbf{Aggregate:~}&{\bf a}_{i,\tau_i}^{(l)}={\sf PL}_{j \in {\cal N}{(i)}} \Big(q({\bf d}_{j,\tau_j}^{(l-1)},{\bf e}_{ij};{\bf W}_{\tau_i\tau_j\tau_{ij}}^{(l)})\Big), \\
   \textbf{Combine:~}&{\bf d}_{i,\tau_i}^{(l)}={\sf CB}\Big( {\bf d}_{i,\tau_i}^{(l-1)}, {\bf a}_{i,\tau_i}^{(l)};{\bf U}_{\tau_i}^{(l)} \Big),
   \end{split}
%\vspace{-8mm}
\end{equation}
where ${\bf a}_{i,\tau_i}^{(l)}$ is the aggregated output at the $i$th vertex, $\tau_i$ and $\tau_{ij}$ are respectively the types of the $i$th vertex and edge $(i,j)$, ${\cal N}(i)$ is the set of neighboring vertices of the $i$th vertex, ${\bf e}_{ij}$ is the feature of edge $(i,j)$, $q(\cdot)$, ${\sf PL}(\cdot)$ and ${\sf CB}(\cdot)$ are respectively the processing, pooling, and combination functions, and ${\bf W}_{\tau_i\tau_j\tau_{ij}}^{(l)}$ and ${\bf U}_{\tau_i}^{(l)}$ are weight matrices.
%and ${\sf CB}_{\tau(i)}(\cdot)$ is the combination function for updating the hidden representations of vertices of the $\tau(i)$th type.

%\vspace{-5mm}
The choices for the processing and combination functions are flexible \cite{GNNsurvey_Wu2021}. To ensure a GNN satisfying the PE property of a learned policy, the pooling function should satisfy the commutative law, e.g., sum-pooling $\sum_{k=1}^K(\cdot)$, mean-pooling $\frac{1}{K}\sum_{k=1}^K(\cdot)$, and max-pooling $\max_{k=1}^K(\cdot)$.
%sum-pooling (i.e., $\sum_{k=1}^K(\cdot)$), mean-pooling (i.e., $\frac{1}{K}\sum_{k=1}^K(\cdot)$) and max-pooling (i.e., $\max_{k=1}^K(\cdot)$).

%\vspace{-6mm}
\begin{remark}\label{remark: gnn funcs}
	 \emph{We consider simple pooling, processing, and combination functions as in \cite{LS2021Lee,Het_PA2022GJ,2022ZBC,LSJ2023MGNN} for easy analysis, where ${\sf PL}(\cdot)$ is the sum-pooling function, $q(\cdot)$ is a linear function, and ${\sf CB}(\cdot)$ is a FNN without hidden layer (i.e., a linear function cascaded with an activation function), unless otherwise specified. The GNNs with the three functions are called {vanilla GNNs}.}
\end{remark}\vspace{-3mm}

The vanilla GNN updating vertex representations for learning the link scheduling policy over ${\cal G}_{\sf ls}^{\sf het}$ is referred to as {vanilla Vertex-GNN$_{\sf ls}^{\sf het}$}, where the active state $x_i$ is defined as the action of vertex $T_i$.
%where the actions are defined on $\rm tx$ vertices.
Since there are two types of vertices in ${\cal G}_{\sf ls}^{\sf het}$, the GNN respectively updates the hidden representations of the vertices of each type in each layer.

From \eqref{eq: node-GNN},
the hidden representations of $T_i$ and $R_i$ in the $l$th layer of the {vanilla Vertex-GNN$_{\sf ls}^{\sf het}$}, ${\bf{d}}_{i,T}^{(l)}$ and ${\bf{d}}_{i,R}^{(l)}$, are updated as,
\begin{subequations} \label{eq:Node-HetGNN-update}
	\small
	\vspace{-2mm}
  \begin{align}
  \begin{split} \label{eq: Node-HetGNN Tx}
  &\textbf{Update hidden representations of $\rm \bf tx$ vertices:} \\
&\textbf{Aggregate:}~
  {\bf a}_{i,T}^{(l)}={\bf Q}_{RS}^{(l)} {\bf d}_{i,R}^{(l-1)}+{\bf P}_{RS}^{(l)}\alpha_{ii}+\sum_{j=1,j\neq i}^K \Big( {\bf Q}_{RI}^{(l)} {\bf d}_{j,R}^{(l-1)}+{\bf P}_{RI}^{(l)}\alpha_{ij}\Big),\\
&\textbf{Combine:}~~
  {\bf d}_{i,T}^{(l)}=\sigma\Big( {\bf U}_T^{(l)}{\bf d}_{i,T}^{(l-1)}+{\bf a}_{i,T}^{(l)}\Big),
  \end{split}\\
  \begin{split} \label{eq: Node-HetGNN Rx}
  &\textbf{Update hidden representations of $\rm \bf rx$ vertices:} \\
&\textbf{Aggregate:}~
  {\bf a}_{i,R}^{(l)}={\bf Q}_{TS}^{(l)} {\bf d}_{i,T}^{(l-1)}+{\bf P}_{TS}^{(l)}\alpha_{ii}+\sum_{j=1,j\neq i}^K \Big( {\bf Q}_{TI}^{(l)} {\bf d}_{j,T}^{(l-1)}+{\bf P}_{TI}^{(l)}\alpha_{ji}\Big),\\
&\textbf{Combine:}~~
  {\bf d}_{i,R}^{(l)}=\sigma\Big( {\bf U}_R^{(l)}{\bf d}_{i,R}^{(l-1)}+{\bf a}_{i,R}^{(l)}\Big),
  \end{split}
\vspace{-2mm}
  \end{align}
\end{subequations}
where ${\bf a}_{i,T}^{(l)}$ and ${\bf a}_{i,R}^{(l)}$ are respectively the aggregated outputs at $T_i$ and $R_i$, ${\bf Q}_{RS}^{(l)}$, ${\bf P}_{RS}^{(l)}$, ${\bf Q}_{RI}^{(l)}$, and ${\bf P}_{RI}^{(l)}$ are respectively the weight matrices for processing the information of $R_i$, edge $(i,i)$, $R_j, j\neq i$, and edges $(i,j), j\neq i$,
%the neighboring receiver vertex of $T_i$ connected by the signal edge,
%${\bf Q}_{RI}^{(l)}$ and ${\bf P}_{RI}^{(l)}$ are weight matrices for respectively processing the information of the neighboring receiver vertices and the connected interference edges,
${\bf U}_T^{(l)}$ is the weight matrix for combining when updating $T_i$, $\sigma(\cdot)$ is an activation function, ${\bf Q}_{TS}^{(l)}$, ${\bf P}_{TS}^{(l)}$, ${\bf Q}_{TI}^{(l)}$, ${\bf P}_{TI}^{(l)}$, and ${\bf U}_{R}^{(l)}$ in \eqref{eq: Node-HetGNN Rx} are respectively the weight matrices used for processing and combining when updating $R_i$.

In the input layer (i.e., $l=0$), we set ${d}_{i,T}^{(0)}=0$ and ${d}_{i,R}^{(0)}=0$, because the vertices in ${\cal G}_{\sf {ls}}^{\sf het}$ have no features. The input of the GNN is $\bm \alpha$, which is composed of the features of all edges. The output of the GNN is $[{ d}_{1,T}^{(L)},\cdots,{d}_{K,T}^{(L)}]^{\sf T} \triangleq[\hat x_{1,T},\cdots, \hat x_{K,T}]^{\sf T} \triangleq \hat{\bf x}_T $, which is composed of the learned actions taken on all the $\rm tx$ vertices, where $L$ is the number of layers of the GNN.
Denote the policy learned by the vanilla Vertex-GNN$_{\sf ls}^{\sf het}$ as $\hat {\bf x}_T=G_{\sf {ls,v}}^{\sf{het}}(\bm \alpha)$. It is not hard to show that the learned policy is joint-PE to ${\bm \alpha}$.

\vspace{-4mm}\begin{remark}\label{remark: Vertex-D-GNN}
	 \emph{In \cite{LS2021Lee,2021Shen,IoT-supervised2022}, Vertex-GNNs were used to learn the link scheduling and power control policies over directed homogeneous graphs with the same topology as ${\cal G}_{\sf {ls}}^{\sf hom}$ in Fig. \ref{fig:graph}(a).
To apply the update equation in \eqref{eq: node-GNN}, one can convert ${\cal G}_{\sf {ls}}^{\sf hom}$ into a heterogeneous graph, denoted as ${\cal G}_{\sf {ls,v}}^{\sf undir}$.
%which consists of one type of vertices and two types of edges.
When updating the representation of vertex $D_i$, its neighboring edges $(i,j)$ and $(j,i)$ $(j\neq i)$ are two types of edges, and its neighboring vertices are with two different types. Then, the representation of $D_i$ in the $l$th layer, ${\bf d}_{i,V}^{(l)}$, can be obtained from \eqref{eq: node-GNN}, where ${d}_{i,V}^{(0)}=\alpha_{ii}$.
	 The input of the GNN is  $\bm \alpha$, which is composed of the features of all vertices and edges.
	 The output of the GNN is $[{d}_{1,V}^{(L)},\cdots,{d}_{K,V}^{(L)}]^{\sf T}  \triangleq [{\hat x_{1,V}},\cdots,{\hat x_{K,V}}]^{\sf T} \triangleq \hat{\bf x}_V $, which includes the learned actions on all vertices.
When the three functions in Remark \ref{remark: gnn funcs} are used, the Vertex-GNN learning over ${\cal G}_{\sf {ls,v}}^{\sf{undir}}$ is referred to as {vanilla Vertex-GNN$_{\sf ls}^{\sf undir}$}. It is not hard to show that the learned policy is joint-PE to ${\bm \alpha}$.}
\end{remark}
\vspace{-6mm}\begin{remark}\label{remark: Vertex-GNN-precoding}
\emph{When using a Vertex-GNN to learn the precoding policy over ${\cal G}_{\sf {p}}^{\sf het}$,
the hidden representations of antenna vertex $A_i$ and user vertex $U_j$ in the $l$th layer, ${\bf d}_{i,A}^{(l)}$ and ${\bf d}_{j,U}^{(l)}$, can be obtained from \eqref{eq: node-GNN}, where ${d}_{i,A}^{(0)}={d}_{j,U}^{(0)}=0$ because the vertices have no features.
Since the actions are defined on edges, a read-out layer is required to map the vertex representations in the output layer into the actions. In particular, to map ${\bf d}_{i,A}^{(L)}$ and ${\bf d}_{j,U}^{(L)}$ into the action on edge $(i,j)$, a FNN can be designed as $\hat v_{ij,{V}}={\sf FNN}_{\sf read}^{\sf het}({\bf d}_{i,A}^{(L)},{\bf d}_{j,U}^{(L)})$, which is the same for $i=1,\cdots, N, j=1,\cdots, K$.
%which can be expressed as $\hat w_{ij,{\sf v}}={\sf FNN}_{\sf read}({\bf d}_{i,A}^{(L)},{\bf d}_{j,U}^{(L)}) $, where ${\sf FNN}_{\sf read}(\cdot)$ denotes a FNN.
The input and output of the GNN are respectively $\bf H$ and $[{\hat v}_{ij,V}]_{N\times K}\triangleq \hat {\bf V}_V$, which are composed of the features and the learned actions of all edges.
%The learned precoding policy is denoted as ${\hat {\bf V}_{\sf v}}=G_{\sf{p,v}}^{\sf{het}} (\bf H)$.
When the three functions in Remark \ref{remark: gnn funcs} are used, the GNN is referred to as {vanilla Vertex-GNN$_{\sf p}^{\sf het}$}. It is not hard to show that the learned policy is 2D-PE to {\bf H}.}
\end{remark}\vspace{-2mm}

\vspace{-1mm}
\subsubsection{Edge-GNN}\label{sec: Edge-GNN}
In Edge-GNN, the hidden representation of each edge is updated in
each layer, by first aggregating information from its neighboring edges and neighboring vertices, and then combining with its own hidden representation in the previous layer.
For edge $(i,j)$, the $i$th and $j$th vertices are its neighboring vertices, and the edges connected by the $i$th and $j$th vertices are its neighboring edges.
%For each edge, its neighboring vertices are two vertices connected by it, and its neighboring edges are the edges connected to its neighboring vertices.
%as illustrate in Fig. \ref{fig:Vertex-HetGNN}(c)(d).
%For example, for the edge $(1, 1)$ in Fig. \ref{fig:graph}(b), $T_1$ and $R_1$ are its neighboring vertexes, and the edge $(1,2)$, $(1,3)$, $(2,1)$ and $(3,1)$ are its neighboring edges.

The update equation of an Edge-GNN can be obtained from the update equation of a Vertex-GNN simply by switching the roles of the edges and vertices \cite{Hypergraphs-S2021}.
For the Edge-GNN learning over a heterogeneous graph, the hidden representation of edge $(i,j)$ with the $\tau_{ij}$th type in the $l$th layer, denoted as ${\bf d}_{ij,\tau_{ij}}^{(l)}$, is updated as follows,
\begin{equation}\label{eq:edge-GNN}
  \small
  \begin{split}
  \textbf{Aggregate:}~{\bf a}_{ij,\tau_{ij}}^{(l)}&={\sf PL}_{\substack{m\in {\cal N}(i)/j\\n\in {\cal N}(j)/i} } \Big( q({\bf d}_{im,\tau_{im}}^{(l-1)},{\bf e}_{v_i};{\bf W}_{\tau_{ij}\tau_{im}\tau_i}^{(l)}),q({\bf d}_{nj,\tau_{nj}}^{(l-1)},{\bf e}_{v_j};{\bf W}_{\tau_{ij}\tau_{nj}\tau_j}^{(l)})\Big), \\
  \textbf{Combine:}~~{\bf d}_{ij,\tau_{ij}}^{(l)}&={\sf CB}\Big( {\bf d}_{ij,\tau_{ij}}^{(l-1)}, {\bf a}_{ij,\tau_{ij}}^{(l)};{\bf U}_{\tau_{ij}}^{(l)}\Big),
  \end{split}
\end{equation}
where ${\bf a}_{ij,\tau_{ij}}^{(l)}$ is the aggregated output at edge $(i,j)$, the first and second processors are respectively used to process the information from the $i$th vertex and its connected edges and the $j$th vertex and its connected edges, ${\cal N}(i)/j$ is a set of neighboring vertices of the $i$th vertex except the $j$th vertex,
% where ${\cal N}_{tv}{(i,j)}$ is a set of edges and vertexs (say the edge $(m,n)$ and the $k$th vertex, the edge $(m,n)$ with the $t$th type neighbors to the edge $(i,j)$ through the $k$th vertex with the $v$th type),
${\bf e}_{v_j}$ denotes the feature of the $j$th vertex, and ${\bf W}_{\tau_{ij}\tau_{nj}\tau_j}^{(l)}$ and ${\bf U}_{\tau_{ij}}^{(l)}$ are the weight matrices.
%and ${\sf CB}_{\tau(ij)}(\cdot)$ is the combination function for updating the hidden representations of edges with the $\tau(ij)$th type.

When an Edge-GNN is used for learning the link scheduling policy over ${\cal G}_{\sf ls}^{\sf het}$ in Fig. \ref{fig:graph}(b), the actions are defined on the $\rm sig$ edges.
When the pooling, processing, and combination functions in Remark \ref{remark: gnn funcs} are used, the Edge-GNN is referred to as {vanilla Edge-GNN$_{\sf ls}^{\sf het}$}.

Since there are two types of edges in ${\cal G}_{\sf {ls}}^{\sf het}$, the GNN respectively updates the hidden representations of the edges of each type in each layer.
Since the vertices have no features in ${\cal G}_{\sf {ls}}^{\sf het}$, when updating the representation of each edge, only the information of its neighboring edges is aggregated.
For edge $(i,i)$ (say, edge $(1,1)$ in Fig. \ref{fig:Vertex-HetGNN}(c)), edge $(i,j)$ and edge $(j,i)$  $(j\neq i)$ respectively connected by $T_i$ and $R_i$ are its neighboring edges.
%Since $T_i$ and $R_i$ are of different types, the weight matrices for processing the information from edge $(i,k)$ and edge $(k,i)$ differ.
For edge $(i,j)$  (say, edge $(1,2)$ in Fig. \ref{fig:Vertex-HetGNN}(d)), edge $(i,i)$ and edge $(i,k)$ $(k\neq \{i,j\})$  connected by $T_i$ are respectively its neighboring $\rm sig$ and $\rm int$ edges, and edge $(j,j)$ and edge $(k,j)$  connected by $R_j$ are respectively its neighboring $\rm sig$ and $\rm int$ edges.

From \eqref{eq:edge-GNN}, the hidden representations of edge $(i,i)$ and edge $(i,j)$ in the $l$th layer of the {vanilla Edge-GNN$_{\sf ls}^{\sf het}$}, ${\bf d}_{i,S}^{(l)}$ and ${\bf d}_{ij,I}^{(l)}$, are updated as follows,
\begin{subequations} \label{eq:Edge_HetGNN_update}
  \small
  \begin{align}
  \begin{split} \label{eq:Edge_HetGNN com}
  &\textbf{Update hidden representations of ${\rm\bf sig}$ edges:}\\
  &\textbf{Aggregate:}
  ~{\bf a}_{i,S}^{(l)}=\sum_{k=1,k\neq i}^K {\bf Q}_{T}^{(l)}{\bf d}_{ik,I}^{(l-1)}+\sum_{k=1,k\neq i}^K {\bf Q}_{R}^{(l)}{\bf d}_{ki,I}^{(l-1)}, \\
  &\textbf{Combine:}~~{\bf d}_{i,S}^{(l)}=\sigma_S\Big( {\bf U}_S^{(l)}{\bf d}_{i,S}^{(l-1)}+{\bf a}_{i,S}^{(l)}\Big),
  \end{split} \\
  \begin{split} \label{eq:Edge_HetGNN inter}
  &\textbf{Update hidden representations of ${\rm\bf int}$ edges:}\\
  &\textbf{Aggregate:}~{\bf a}_{ij,I}^{(l)}={\bf Q}_{ST}^{(l)}{\bf d}_{i,S}^{(l-1)}+\sum_{k=1,k\neq \{i,j\}}^K {\bf Q}_{IT}^{(l)}{\bf d}_{ik,I}^{(l-1)}+{\bf Q}_{SR}^{(l)}{\bf d}_{j,S}^{(l-1)}+\sum_{k=1,k\neq \{i,j\}}^K {\bf Q}_{IR}^{(l)}{\bf d}_{kj,I}^{(l-1)},\\
  &\textbf{Combine:}~~{\bf d}_{ij,I}^{(l)}=\sigma_I \Big( {\bf U}_I^{(l)}{\bf d}_{ij,I}^{(l-1)}+{\bf a}_{ij,I}^{(l)}\Big),\\
  \end{split}
  \end{align}
\end{subequations}
where ${\bf a}_{i,S}^{(l)}$ and ${\bf a}_{ij,I}^{(l)}$ are respectively the aggregated outputs at edge $(i,i)$ and edge $(i,j)$,  ${\bf Q}_{T}^{(l)}$ and ${\bf Q}_{R}^{(l)}$ are respectively the weight matrices for  processing the information of the neighboring $\rm int$ edges of the $\rm sig$ edge $(i,i)$ connected by $T_i$ and $R_i$, ${\bf Q}_{ST}^{(l)}$ and ${\bf Q}_{IT}^{(l)}$ are respectively the weight matrices for processing the information of the neighboring $\rm sig$ and $\rm int$ edges of the $\rm int$ edge $(i,j)$ connected by $T_i$, ${\bf Q}_{SR}^{(l)}$ and ${\bf Q}_{IR}^{(l)}$ are respectively used for processing the information from neighboring $\rm sig$ and $\rm int$ edges of the $\rm int$ edge $(i,j)$ connected by $R_j$, ${\bf U}_S^{(l)}$ and ${\bf U}_I^{(l)}$ are respectively the weight matrices for combining the information of the $\rm sig$ and $\rm int$ edges, and $\sigma_S(\cdot)$ and $\sigma_I(\cdot)$ are activation functions.
%Again, we do not distinguish activation functions $\sigma_S(\cdot)$ and $\sigma_I(\cdot)$ in different layers.

In the input layer, ${d}_{i,S}^{(0)}=\alpha_{ii}$ and ${d}_{ij,I}^{(0)}=\alpha_{ij}$. The input of the GNN is $\bm \alpha$, which consists of the features of all edges. The output of the GNN is $[{ d}_{1,S}^{(L)},\cdots,{ d}_{K,S}^{(L)}]^{\sf T}  \triangleq [{\hat x_{1,S}},\cdots,{\hat x_{K,S}}]^{\sf T}  \triangleq \hat {\bf x}_S$, which consists of the learned actions on all the $\rm sig$ edges.
The policy learned by the vanilla Edge-GNN$_{\sf ls}^{\sf het}$ is denoted as $\hat {\bf x}_S= G_{\sf ls,e}^{\sf het}(\bm \alpha)$, which is easily shown as joint-PE to ${\bm \alpha}$.

%It is noteworthy that the information of the neighboring edges of an edge is aggregated at the edge in \eqref{eq:Edge_HetGNN_update}. For example, ${\bf a}_{i,S}^{(l)}$ in \eqref{eq:Edge_HetGNN com} is the aggregated output at edge $(i,j)$. In fact, the aggregated information in \eqref{eq:Edge_HetGNN_update} can also be expressed as the aggregation at the neighboring vertexes. For example, the first term in \eqref{eq:Edge_HetGNN com} is the aggregated output at vertex $T_i$, and the second term is the aggregated output at vertex $R_i$. Hence, it does not matter whether the aggregated outputs are at vertexes or edges.

\vspace{-2mm}\begin{remark}\label{remark: Edge-D-GNN}
      \emph{When using the Edge-GNN to learn the link scheduling policy over ${\cal G}_{\sf {ls}}^{\sf hom}$ with the update equations in \eqref{eq:edge-GNN}, ${\cal G}_{\sf {ls}}^{\sf hom}$ needs to be converted into a heterogeneous graph, denoted as ${\cal G}_{\sf {ls,e}}^{\sf{undir}}$. For the directed edge $(i,j)$, its neighboring vertices $D_i$ and $D_j$ are two types of vertices, its neighboring edges $(i,k)$ and $(j,k)$ are one type of edges while edges $(k,i)$ and $(k,j)$ are another type of edges. Then, the hidden representation of edge $(i,j)$ in the $l$th layer, ${\bf d}_{ij,{\sf ls}}^{(l)}$, can be obtained from \eqref{eq:edge-GNN}.
      	Since the actions are defined on vertices in ${\cal G}_{\sf {ls}}^{\sf hom}$, a read-out layer is required to map the representations of edges in the output layer to the action on each vertex. For example, a FNN layer can be used that is shared among $D_i, i=1,\cdots,K$, which can be designed as $\hat {x}_{i,E}={\sf FNN}_{\sf read}^{\sf undir}\Big(\sum_{k=1,k\neq i}^K {\bf d}_{ik,{\sf ls}}^{(L)},\sum_{k=1,k\neq i}^K{\bf d}_{ki,{\sf ls}}^{(L)}\Big)$.
      The input of the GNN is $\bm \alpha$,  which is composed of the features of all the vertices and edges. The output of the GNN is $[\hat {x}_{1,E},\cdots,\hat {x}_{K,E}]\triangleq{\bf\hat x}_{E}$, consisting of the learned actions on all vertices.
      When using the processing, pooling, and combination functions in Remark \ref{remark: gnn funcs}, the GNN is referred to as {vanilla Edge-GNN$_{\sf ls}^{\sf undir}$}. It is not hard to show that the learned policy is joint-PE to ${\bm \alpha}$.}
\end{remark}

\vspace{-4mm}\begin{remark}\label{remark: Edge-GNN-precoding}
\emph{In \cite{2022ZBC}, the precoding policy was learned with a vanilla Edge-GNN over ${\cal G}_{\sf{p}}^{\sf{het}}$, where the hidden representation of edge $(i,j)$ in the $l$th layer, ${\bf d}_{ij,{E}}^{(l)}$, can be obtained from \eqref{eq:edge-GNN} with the three functions in Remark 1.
This GNN is referred to as {vanilla Edge-GNN$_{\sf p}^{\sf het}$}. The input and output of the GNN are respectively $\bf H$ and $[{\bf d}_{ij,E}^{(L)}]_{N\times K}$, which are composed of the features and the learned actions of all edges. It was shown that the learned policy is 2D-PE to ${\bf H}$.
}
\end{remark}\vspace{-2mm}

The above-mentioned GNNs and the features and actions of graphs for each GNN are summarized in Table \ref{table: name-GNN} and Table \ref{table: define-feature-action}, respectively.
\begin{table}[!htb]
	\centering
	\small
	\vspace{-4mm}
	\caption{Vertex- and edge-GNNs learned over the graphs in Fig.\ref{fig:graph}}\label{table: name-GNN}\vspace{-3mm}
	\begin{tabular}{|c|c|c|c|c|}
		\hline
		\multicolumn{3}{|c|}{\textbf{GNNs}}&\textbf{Policies}&\textbf{Graphs}\\
		\hline
		\multirow{4}{*}[+1.5ex]{Vertex-GNNs}&\multirow{2}{*}{Vertex-GNN$_{\sf ls}$}&Vertex-GNN$_{\sf ls}^{\sf undir}$&\multirow{2}{*}{Link scheduling/power control}& \makecell{${\cal G}_{\sf{ls,v}}^{\sf{undir}}$, converted from \\${\cal G}_{\sf{ls}}^{\sf{hom}}$ in Fig.\ref{fig:graph}(a)} \\
		\cline{3-3}\cline{5-5}
		&&Vertex-GNN$_{\sf ls}^{\sf het}$&& ${\cal G}_{\sf{ls}}^{\sf{het}}$ in Fig.\ref{fig:graph}(b) \\
		\cline{2-5}
		&\multicolumn{2}{c|}{Vertex-GNN$_{\sf p}^{\sf het}$}&Precoding &${\cal G}_{\sf{p}}^{\sf{het}}$ in Fig.\ref{fig:graph}(c) \\
		\hline
		\multirow{4}{*}[+1.5ex]{Edge-GNNs}&\multirow{2}{*}{Edge-GNN$_{\sf ls}$}&Edge-GNN$_{\sf ls}^{\sf undir}$&\multirow{2}{*}{Link scheduling/power control}& \makecell{${\cal G}_{\sf{ls,e}}^{\sf{undir}}$, converted from \\${\cal G}_{\sf{ls}}^{\sf{hom}}$ in Fig.\ref{fig:graph}(a)}\\
		\cline{3-3}\cline{5-5}
		&&Edge-GNN$_{\sf ls}^{\sf het}$&& ${\cal G}_{\sf{ls}}^{\sf{het}}$ in Fig.\ref{fig:graph}(b) \\
		\cline{2-5}
		&\multicolumn{2}{c|}{Edge-GNN$_{\sf p}^{\sf het}$}&Precoding& ${\cal G}_{\sf{p}}^{\sf{het}}$ in Fig.\ref{fig:graph}(c) \\
		\hline
	\end{tabular}\vspace{-8mm}
\end{table}

\begin{table}[H]
	\centering %\vspace{-6mm}
	\small
	\caption{Features and actions of the graphs for Each GNN}\label{table: define-feature-action}
	%\footnotesize
	\renewcommand\arraystretch{0.95}\vspace{-2mm}
\begin{threeparttable}
	\begin{tabular}{|c|c|c|c|c|c|}
		\hline
		\multirow{2}{*}{\textbf{Graphs}} & \multicolumn{2}{c|}{\textbf{Features}}& \multicolumn{2}{c|}{\textbf{Actions}}&\multirow{2}{*}{\textbf{GNNs}} \\
		\cline{2-5}
      &{\textbf {Vertex}}&{\textbf {Edge}}&{\textbf {Vertex}}&{\textbf {Edge}}&\\
     \hline
		${\mathcal G}_{\sf ls,v}^{\sf undir}$\tnote{*}& \multirow{2}{*}{vertex $D_i$: $\alpha_{ii}$ } &\multirow{2}{*}{edge $(i,j)$: $\alpha_{ij}$ }&\multirow{2}{*}{\tabincell{c}{vertex $D_i$: $x_i$ or $p_i$ }}&\multirow{2}{*}{-}&\tabincell{c}{Vertex-GNN$_{\sf ls}^{\sf undir}$} \\
		\cline{1-1}\cline{6-6}
		${\mathcal G}_{\sf ls,e}^{\sf undir}\tnote{*}$&& & & &\tabincell{c}{Edge-GNN$_{\sf ls}^{\sf undir}$} \\
		\hline
		\multirow{2}{*}{${\mathcal G}_{\sf ls}^{\sf het}$}& \multirow{2}{*}{-} &\multirow{2}{*}{edge $(i,j)$: $\alpha_{ij}$ }&{vertex $T_i$: $x_i$ or $p_i$ }&{-}&\tabincell{c}{Vertex-GNN$_{\sf ls}^{\sf het}$} \\
		\cline{4-5}\cline{6-6}
		&& &{-}&{$\rm sig$ edge $(i,i)$: $x_i$ or $p_i$}&\tabincell{c}{Edge-GNN$_{\sf ls}^{\sf het}$} \\
		\hline
		\multirow{2}{*}{${\mathcal G}_{\sf p}^{\sf het}$}& \multirow{2}{*}{-} &\multirow{2}{*}{edge $(i,j)$: $h_{ij}$ }&\multirow{2}{*}{-}&\multirow{2}{*}{ edge $(i,j)$: $v_{ij}$}&\tabincell{c}{Vertex-GNN$_{\sf p}^{\sf het}$} \\
		\cline{6-6}
		& & & & &\tabincell{c}{Edge-GNN$_{\sf p}^{\sf het}$}\\
		\hline
	\end{tabular}
\begin{tablenotes}
	\footnotesize
   \item[*] ${\mathcal G}_{\sf ls,v}^{\sf undir}$ and ${\mathcal G}_{\sf ls,e}^{\sf undir}$ are the converted heterogeneous graphs from ${\mathcal G}_{\sf ls}^{\sf hom}$, as explained in Remark \ref{remark: Vertex-D-GNN} and Remark \ref{remark: Edge-D-GNN}.
\end{tablenotes}\vspace{-7mm}
\end{threeparttable}
\vspace{-2mm}
\end{table}

\vspace{-2mm}
\subsection{Expressive Power of the Vanilla GNNs} \label{sec:differences in theory}
In what follows, we analyze the expressive power of the vanilla Vertex-GNNs and vanilla Edge-GNNs, by observing whether or not a GNN can output different actions when inputting different channel matrices.
%From the input-output relation of a GNN,
%if the GNN can learn different actions from different channel matrices.
Without loss of generality, we assume that $L\geq 3$.

We first define several notations to be used in the sequel.

%\begin{itemize}
%\item
${\alpha}_{\rm R/i} \triangleq \sum_{j=1,j\neq i}^K \alpha_{ij}$ and ${\alpha}_{\rm C/i} \triangleq \sum_{j=1,j\neq i}^K \alpha_{ji}$, which are the summations of the $i$th row of the channel matrix ${\bm \alpha}$ without $\alpha_{ii}$ and the $i$th column of ${\bm \alpha}$ without $\alpha_{ii}$, respectively.

%\item
${ H}_{\rm R,i} \triangleq \sum_{j=1}^K h_{ij}$ and ${H}_{\rm C,i} \triangleq \sum_{j=1}^N h_{ji}$, which are the summations of the $i$th row and column of the channel matrix ${\bf H}$, respectively.

%\item
$\mathcal{A}_{\rm{CR}}\triangleq\{{ \alpha}_{\rm{R/1}},\cdots,{\alpha}_{\rm{R/K}},{\alpha}_{\rm{C/1}},\cdots,{\alpha}_{\rm{C/K}}\}$, which is the set composing of ${\alpha}_{\rm R/i}$ and ${\alpha}_{\rm C/i}, i=1,\cdots,K$.

%\item
$\mathcal{A}_{\rm{diag}}\triangleq\{\alpha_{11},\cdots,\alpha_{KK}\}$, which is the set composing of all the diagonal elements in ${\bm \alpha}$.

%\item
$\mathcal{A}_{\rm ind}\triangleq\{\alpha_{ij}, i,j =1, \cdots,K\}$, which is the set composing of all the elements in $\bm \alpha$.

$\mathcal{H}_{\rm ind}\triangleq\{h_{ij}, i=1,\cdots,N ,j=1, \cdots,K\}$, which is the set composing of all elements in ${\bf H}$.

%\item
$f(\cdot)$ with different super- and sub-scripts are non-linear functions.
%\end{itemize}

\vspace{1mm}
\subsubsection{Link Scheduling Policy} \label{sec: Differ-GNN-LS}
For the vanilla Vertex-GNN$_{\sf ls}^{\sf het}$, by respectively substituting ${\bf a}_{i,T}^{(l)}$ and ${\bf a}_{i,R}^{(l)}$ into ${\bf d}_{i,T}^{(l)}$ and ${\bf d}_{i,R}^{(l)}$ in \eqref{eq:Node-HetGNN-update}, the hidden representations of $T_i$ and $R_i$ in the $l$th layer can be respectively updated as, \vspace{-2mm}
\begin{subequations} \label{eq:HetGNN_detail}
\small
\vspace{-2mm}
\begin{align}
\begin{split} \label{eq:HetGNN_Tx_detail}
{\bf d}_{i,T}^{(l)}= \sigma\Big({\bf U}_T^{(l)}{\bf d}_{i,T}^{(l-1)}+{\bf Q}_{RS}^{(l)}{\bf d}_{i,R}^{(l-1)}+{\bf Q}_{RI}^{(l)}\sum_{j=1,j\neq i}^{K}{\bf d}_{j,R}^{(l-1)}+{\bf P}_{RS}^{(l)}\alpha_{ii}+{\bf P}_{RI}^{(l)}{\alpha}_{\rm{R/i}}\Big),
\end{split}\\
\begin{split} \label{eq:HetGNN_Rx_detail}
{\bf d}_{i,R}^{(l)}= \sigma\Big({\bf U}_R^{(l)}{\bf d}_{i,R}^{(l-1)}+{\bf Q}_{TS}^{(l)}{\bf d}_{i,T}^{(l-1)}+{\bf Q}_{TI}^{(l)}\sum_{j=1,j\neq i}^{K}{\bf d}_{j,T}^{(l-1)}+{\bf P}_{TS}^{(l)}\alpha_{ii}+{\bf P}_{TI}^{(l)}{\alpha}_{\rm{C/i}}\Big).
\end{split}
\end{align}
\end{subequations}\vspace{-8mm}

Since ${d}_{i,T}^{(0)}={d}_{i,R}^{(0)}=0$, when $l=1$, from \eqref{eq:HetGNN_detail} we have
\begin{subequations} \label{eq:Vertex-HetGNN-non-linear-1}
\small
\vspace{-4mm}
\begin{align}
\begin{split} \label{eq:Tx-non-linear-1}
{\bf d}_{i,T}^{(1)}&=\sigma\Big({\bf P}_{RS}^{(1)}\alpha_{ii}+{\bf P}_{RI}^{(1)}{\alpha}_{\rm{R/i}}\Big)\triangleq f_T^{(1)}(\alpha_{ii},{\alpha}_{\rm{R/i}}),
\end{split}\\
\begin{split} \label{eq:Rx-non-linear-1}
{\bf d}_{i,R}^{(1)}&=\sigma\Big({\bf P}_{TS}^{(1)}\alpha_{ii}+{\bf P}_{TI}^{(1)}{\alpha}_{\rm{C/i}}\Big)\triangleq f_R^{(1)}(\alpha_{ii},{\alpha}_{\rm{C/i}}).
\end{split}
\end{align}
\end{subequations}

When $l=2$, by substituting ${\bf d}_{i,T}^{(1)}$ and ${\bf d}_{i,R}^{(1)}$ in \eqref{eq:Vertex-HetGNN-non-linear-1} into \eqref{eq:HetGNN_detail}, it is not hard to derive
\begin{subequations} \label{eq:Vertex-HetGNN-non-linear-2}
\small
\vspace{-4mm}
\begin{align}
\begin{split} \label{eq:Tx-non-linear-2}
{\bf d}_{i,T}^{(2)}\triangleq
 f_{i,T}^{(2)}(\alpha_{11},\cdots,\alpha_{KK},{\alpha}_{\rm{C/1}},\cdots,{\alpha}_{\rm{C/K}},{\alpha}_{\rm{R/i}}),
\end{split}\\
\begin{split} \label{eq:Rx-non-linear-2}
{\bf d}_{i,R}^{(2)}\triangleq
f_{i,R}^{(2)}(\alpha_{11},\cdots,\alpha_{KK},{\alpha}_{\rm{R/1}},\cdots,{\alpha}_{\rm{R/K}},{\alpha}_{\rm{C/i}}).
\end{split}
\end{align}
\end{subequations}\vspace{-10mm}

Similarly, the action taken over $T_i$ can be derived as,
\vspace{-4mm}
\begin{equation} \label{eq:Vertex-HetGNN-ft}
\small
\vspace{-4mm}
\begin{split}
{\hat x}_{i,T}&\triangleq f_{i,T}(\alpha_{11},\cdots,\alpha_{KK},{\alpha}_{\rm{C/1}},\cdots,{\alpha}_{\rm{C/K}},{\alpha}_{\rm{R/1}},\cdots,{\alpha}_{\rm{R/K}})
= f_{i,T}(\mathcal{A}_{\rm{diag}},\mathcal{A}_{\rm{CR}}).
\end{split}
\end{equation}
It is shown that the information of interference channel gains $\alpha_{ij}, i\neq j$ is lost after the aggregation in the update equation of the vanilla Vertex-GNN$_{\sf ls}^{\sf het}$.

%Denote $f_{i,V}(\cdot)$, $f_{i,S}(\cdot)$, and $f_{i,E}(\cdot)$ as non-linear functions.
Analogously, for the vanilla Vertex-GNN$_{\sf ls}^{\sf undir}$, the vanilla Edge-GNN$_{\sf ls}^{\sf het}$, and the vanilla Edge-GNN$_{\sf ls}^{\sf undir}$, the actions taken over the $i$th vertex or the $i$th $\rm sig$ edge can respectively be expressed as,
\vspace{-2mm}
\begin{equation} \label{eq:Vertex-D-GNN-fv}
\small
\vspace{-4mm}
{\hat x}_{i,V}\triangleq f_{i,V}(\mathcal{A}_{\rm{diag}}, \mathcal{A}_{\rm{CR}}),
~~{\hat x}_{i,S}\triangleq f_{i,S}(\mathcal{A}_{ind}),
~~{\hat x}_{i,E}\triangleq f_{i,E}(\mathcal{A}_{ind}).
\end{equation}

Denote the outputs of the vanilla Vertex-GNN$_{\sf ls}$ (i.e., Vertex-GNN$_{\sf ls}^{\sf het}$ or Vertex-GNN$_{\sf ls}^{\sf undir}$) as $\hat{\bf x}_{[1]}$ and $\hat{\bf x}_{[2]}$ with two different inputs ${\bm \alpha}_{[1]}=[\alpha_{[1],ij}]_{K\times K}$ and ${\bm \alpha}_{[2]}=[\alpha_{[2],ij}]_{K\times K}$, respectively.
From \eqref{eq:Vertex-HetGNN-ft} and \eqref{eq:Vertex-D-GNN-fv}, we can obtain the following observation.

{ \bf Observation 1:}
$\hat{\bf x}_{[1]}=\hat{\bf x}_{[2]}$ if the elements in ${\bm \alpha}_{[1]}$ and ${\bm \alpha}_{[2]}$ satisfy the following conditions:  {\sf (1)}~$\alpha_{[1],11}=\alpha_{[2],11},\cdots,\alpha_{[1],KK}=\alpha_{[2],KK}$, {\sf (2)}~${\alpha}_{[1],\rm{C/1}}={\alpha}_{[2],\rm{C/1}},\cdots,{\alpha}_{[1],\rm{C/K}}={\alpha}_{[2],\rm{C/K}}$, and {\sf (3)}~${\alpha}_{[1],\rm{R/1}}={\alpha}_{[2],\rm{R/1}},\cdots,{\alpha}_{[1],\rm{R/K}}={\alpha}_{[2],\rm{R/K}}$.
\vspace{1mm}

%We first discuss the existence of channel matrices ${\bm \alpha}_{[1]}=[\alpha_{[1],ij}]_{K\times K}$ and ${\bm \alpha}_{[2]}=[\alpha_{[2],ij}]_{K\times K}$ that satisfy the conditions in {\bf{Observation 1}}, where the channel gains in ${\bm \alpha}_{[1]}$ and ${\bm \alpha}_{[2]}$ satisfy,
These conditions can be rewritten as a group of linear equations
\begin{equation} \label{eq:variable-equation}
\small
\begin{split}
  \alpha_{[1],ii}=\alpha_{[2],ii},
  ~~\sum_{j=1,j\neq i}^K \alpha_{[1],ji}=\sum_{j=1,j\neq i}^K \alpha_{[2],ji},
  ~~\sum_{j=1,j\neq i}^K \alpha_{[1],ij}=\sum_{j=1,j\neq i}^K \alpha_{[2],ij}, ~~i=1,\cdots,K,
\end{split}
\end{equation}
which consists of $3K$ equations and $2K^2$ variables. When $K \geq 3$, the number of variables is larger than the number of equations, and hence there are infinite solutions to these equations, i.e., there are infinite numbers of ${\bm \alpha}_{[1]}$ and ${\bm \alpha}_{[2]}$ satisfying the conditions.

The observation indicates that the vanilla Vertex-GNN$_{\sf ls}$ for learning the link scheduling policy ${\bf x}^{*}=F_{\sf ls}({\bm \alpha})$ is unable to differentiate all channel matrices. When the pooling function in a vanilla Vertex-GNN$_{\sf ls}$ is replaced by mean-pooling or max-pooling, we can also find channel matrices that the GNN  cannot differentiate.

Recalling that ${\bm \alpha} \in \mathbb{R}^{K\times K}$ and ${\bf x}^* \in \mathbb{I}^{K\times 1}$, the scheduling policy ${\bf x}^*=F_{\sf ls}({\bm \alpha})$ is a many-to-one mapping where the channel matrix is compressed by the mapping. However, the mappings learned by a vanilla Vertex-GNN$_{\sf ls}$ may not be the same as the scheduling policy, because $F_{\sf ls}({\bm \alpha}_{[1]})=F_{\sf ls}({\bm \alpha}_{[2]})$ does not necessarily hold when ${\bm \alpha}_{[1]}$ and ${\bm \alpha}_{[2]}$ satisfy the three conditions, as to be shown via simulations later. As a consequence, the vanilla Vertex-GNN$_{\sf ls}$ cannot well learn the link scheduling policy due to the information loss.

By contrast, the vanilla Edge-GNN$_{\sf ls}$ (i.e., Edge-GNN$_{\sf ls}^{\sf het}$ or Edge-GNN$_{\sf ls}^{\sf undir}$) does not incur the information loss, since it can differentiate the channel matrices resulting in different optimization variables.
This can be seen from \eqref{eq:Vertex-D-GNN-fv}, where the outputs of the vanilla Edge-GNN$_{\sf ls}$ (i.e., ${\hat x}_{i,S}$ and ${\hat x}_{i,E}$) depend on each individual channel gain (say $\alpha_{ij}$) in $\bm \alpha$.

\vspace{1mm}
\subsubsection{Precoding Policy} \label{sec: Differ-GNN-precoding}\vspace{-2mm}
With similar derivations, we can show that $\bf H$ is compressed into ${ H}_{\rm R,i}$ and ${ H}_{\rm C,j}, ~i=1,\cdots, N, ~j=1,\cdots, K$ after the
aggregation of the vanilla Vertex-GNN$_{\sf p}^{\sf het}$ where the information in individual channel coefficients loses. As a result, the GNN is unable to differentiate the channel matrices ${\bf H}_{[1]}=[h_{[1],ij} ]_{N\times K} \neq {\bf H}_{[2]}=[h_{[2],ij}]_{N\times K}$ that satisfy the following conditions:
{\sf (1)}~${H}_{[1],\rm{C,1}}={H}_{[2],\rm{C,1}},\cdots,{H}_{[1],\rm{C,K}}={H}_{[2],\rm{C,K}}$,  {\sf (2)}~${H}_{[1],\rm{R,1}}={H}_{[2],\rm{R,1}},\cdots,{H}_{[1],\rm{R,N}}={H}_{[2],\rm{R,N}}$, which can be expressed as the following group of equations,
\begin{equation} \label{eq:variable-equation-precoding}
\small
\begin{split}
  \sum_{n=1}^N h_{[1],nj}=\sum_{n=1}^N h_{[2],nj}, ~~j=1,\cdots,K,
  ~~\sum_{k=1}^K h_{[1],ik}=\sum_{k=1}^K h_{[2],ik},~~i=1,\cdots,N.
\end{split}
\end{equation}
In other words, when inputting ${\bf H}_{[1]}$ and ${\bf H}_{[2]}$, the outputs of the vanilla Vertex-GNN$_{\sf p}^{\sf het}$ are identical.
When the pooling function is mean-pooling or max-pooling, we can also find channel matrices that the vanilla Vertex-GNN$_{\sf p}^{\sf het}$ cannot differentiate.

However, the precoding matrix depends on every channel coefficient $h_{ij}$. For example, when the signal-to-noise ratio (SNR) is very low, the optimal precoding matrix degenerates into $K$ vectors each for maximal-ratio transmission (i.e., $v_{ij}^{*}/|v_{ij}^{*}|=h_{ij}^{\sf H}/|h_{ij}|$) \cite{2022ZBC}. As a result, the vanilla Vertex-GNN$_{\sf p}^{\sf het}$ is unable to learn the optimal precoding policy.

By contrast, the vanilla Edge-GNN$_{\sf p}^{\sf het}$ can differentiate all channel matrices. The update equation of the {vanilla Edge-GNN$_{\sf p}^{\sf het}$} designed in \cite{2022ZBC} is
\begin{equation} \label{eq:precoding-update}
\small
	{\bf d}_{ij,{E}}^{(l)}=\sigma({\bf U}_{E}^{(l)}{\bf d}_{ij,{E}}^{(l-1)}+\sum_{m=1,m\neq j}^{K}{\bf Q}_{A}^{(l)}{\bf d}_{im,{E}}^{(l-1)}+\sum_{n=1,n\neq i}^{N}{\bf Q}_{U}^{(l)}{\bf d}_{nj,{E}}^{(l-1)}),\vspace{-2mm}
\end{equation}
where ${\bf U}_{E}^{(l)}$ is the weight matrix for the combination, ${\bf Q}_{A}^{(l)}$ and ${\bf Q}_{U}^{(l)}$ are respectively the weight matrices for processing the information of the neighboring edges connected by $A_i$ and $U_j$.
In the input layer, ${d}_{ij,{E}}^{(0)}=h_{ij}$.
The inputs of the GNN are the features of all edges, i.e., $[d_{ij,{E}}^{(0)}]_{N\times K}=[h_{ij}]_{N\times K}=\bf H$.
%In the output layer, ${d}_{ij,{\sf E}}^{(L)}=\hat v_{ij}$.
The outputs of the GNN are the learned actions taken on all the edges, i.e., $[d_{ij,{ E}}^{(L)}]_{N\times K}\triangleq [\hat v_{ij,{E}}]_{N\times K}\triangleq {\hat {\bf V}}_E$.
%The learned precoding policy is denoted as ${\hat {\bf V}}_{\sf e}=G_{\sf {p,e}}^{\sf het}(\bf H)$.

It can be derived from \eqref{eq:precoding-update} that $\hat v_{ij,E}\triangleq f_{ij,E}({\mathcal H}_{ind})$, since every channel coefficient (say $h_{ij}$) is combined individually when updating the edge representation with $l=1$.%

\vspace{-2mm}\begin{remark}\label{remark: constraint}
\emph{When considering other typical constraints, we can use the same way to analyze the expressive power of GNNs, but the input samples that GNNs cannot distinguish may differ.
}
\end{remark}\vspace{-2mm}

\vspace{-3mm}
\section{Impact of Processing and Combination Functions on expressive power} \label{sec:Influence}
\vspace{-1mm}
In this section, we analyze the impact of processing and combination functions on the expressive power of the Vertex-GNNs and Edge-GNNs for learning the policies.

We first analyze the impact of the linearity of processing and combination function, and
then analyze the impact of the output dimensions of the two functions on Vertex-GNNs and Edge-GNNs.

Without the loss of generality, we assume that $L\geq 3$.

Denote ${\alpha}_{\sf{Tr}}\triangleq \sum_{j=1}^K \alpha_{jj}$. Denote $g(\cdot)$ with different super- and sub-scripts as linear functions.

\vspace{-5mm}
\subsection{Impact of Linearity}\label{sec:Influence-q-CB}
%\vspace{-3mm}
As analyzed in section \ref{sec:differences in theory}, the vanilla Vertex-GNNs (i.e., vanilla Vertex-GNN$_{\sf ls}^{\sf het}$, Vertex-GNN$_{\sf ls}^{\sf undir}$, and Vertex-GNN$_{\sf p}^{\sf het}$) cannot while the vanilla Edge-GNNs (i.e., vanilla Edge-GNN$_{\sf ls}^{\sf het}$, Edge-GNN$_{\sf ls}^{\sf undir}$, and Edge-GNN$_{\sf p}^{\sf het}$) can differentiate the channel matrices resulting in different optimization variables, where both classes of vanilla GNNs are with linear processors and non-linear combiners. In the sequel, we show that the expressive power of the Vertex-GNNs can be enhanced by using non-linear processors, and the strong expressive power of the Edges-GNNs comes from the non-linear combiners. We take the GNNs for learning the link scheduling policy as an example. The impact is the same on the GNNs for learning the power control and precoding policies.
\subsubsection{Vertex-GNNs} We start by analyzing the expressive power of a degenerated vanilla Vertex-GNN$_{\sf ls}^{\sf het}$ where the combination function becomes linear.

\emph{Linear processor and linear combinator:}
For Vertex-GNN$_{\sf ls}^{\sf het}$, ${d}_{i,T}^{(0)}={d}_{i,R}^{(0)}=0$. Then, after omitting the activation functions in \eqref{eq:HetGNN_detail}, the hidden representations of $T_i$ and $R_i$ become,
\vspace{-2mm}
\begin{subequations}\label{eq:Vertex-HetGNN-linear-1}
\small
\begin{align}
\begin{split}\label{eq:Tx-linear-1}
{\bf d}_{i,T}^{(1)}&={\bf P}_{RS}^{(l)}\alpha_{ii}+{\bf P}_{RI}^{(1)}{\alpha}_{\rm {R/i}}\triangleq g_T^{(1)}(\alpha_{ii},{\alpha}_{\rm {R/i}}),
\end{split}\\
\begin{split}\label{eq:Rx-linear-1}
{\bf d}_{i,R}^{(1)}&={\bf P}_{TS}^{(l)}\alpha_{ii}+{\bf P}_{TI}^{(1)}{\alpha}_{\rm {C/i}}\triangleq g_R^{(1)}(\alpha_{ii},{\alpha}_{\rm {C/i}}).
\end{split}
\end{align}
\end{subequations}

By substituting ${\bf d}_{i,T}^{(1)}$ and ${\bf d}_{i,R}^{(1)}$ into \eqref{eq:HetGNN_detail}, and again omitting the activation functions, we have,
\vspace{-2mm}
\begin{subequations} \label{eq:Vertex-HetGNN-linear-2}
\small
\begin{align}
\begin{split}
{\bf d}_{i,T}^{(2)}=&{\bf U}_T^{(2)}g_T^{(1)}(\alpha_{ii},{\alpha}_{\rm{R/i}})+{\bf Q}_{RS}^{(2)}g_R^{(1)}(\alpha_{ii},{\alpha}_{\rm{C/i}})+{\bf Q}_{RI}^{(2)}\sum_{j=1,j\neq i}^{K}g_R^{(1)}(\alpha_{jj},{\alpha}_{\rm{C/j}})+{\bf P}_{RS}^{(2)}\alpha_{ii}+{\bf P}_{RI}^{(2)}{\alpha}_{\rm{R/i}}\\
\overset{(a)}{=}&{\bf U}_T^{(2)}g_T^{(1)}(\alpha_{ii},{\alpha}_{\rm{R/i}})+{\bf Q}_{RS}^{(2)}g_R^{(1)}(\alpha_{ii},{\alpha}_{\rm{C/i}})+{\bf Q}_{RI}^{(2)}g_R^{(1)}({\alpha}_{\rm{Tr}}-\alpha_{ii},\Vert{\bm \alpha}\Vert-{\alpha}_{\rm{C/i}}-{\alpha}_{\rm{Tr}})\\&+{\bf P}_{RS}^{(2)}\alpha_{ii}+{\bf P}_{RI}^{(2)}{\alpha}_{\rm{R/i}} \triangleq   g_T^{(2)}(\alpha_{ii},{\alpha}_{\rm{Tr}},{\alpha}_{\rm{R/i}},{\alpha}_{\rm{C/i}}, \Vert{\bm \alpha}\Vert),\label{eq:Vertex-HetGNN-linear-2a}
\end{split}\\
\begin{split}
{\bf d}_{i,R}^{(2)}=&{\bf U}_R^{(2)}g_R^{(1)}(\alpha_{ii},{\alpha}_{\rm{C/i}})+{\bf Q}_{TS}^{(2)}g_T^{(1)}(\alpha_{ii},{\alpha}_{\rm{R/i}})+{\bf Q}_{IT}^{(2)}\sum_{j=1,j\neq i}^{K}g_T^{(1)}(\alpha_{jj},{\alpha}_{\rm{R/j}})+{\bf P}_{TS}^{(2)}\alpha_{ii}+{\bf P}_{TI}^{(2)}{\alpha}_{\rm{C/i}}\\
\overset{(a)}{=}&{\bf U}_R^{(2)}g_R^{(1)}(\alpha_{ii},{\alpha}_{\rm{C/i}})+{\bf Q}_{TS}^{(2)}g_T^{(1)}(\alpha_{ii},{\alpha}_{\rm{R/i}})+{\bf Q}_{TI}^{(2)}g_T^{(1)}({\alpha}_{\rm{Tr}}-\alpha_{ii},\Vert{\bm \alpha}\Vert-{\alpha}_{\rm{R/i}}-{\alpha}_{\rm{Tr}})\\&+{\bf P}_{TS}^{(2)}\alpha_{ii}+{\bf P}_{TI}^{(2)}{\alpha}_{\rm{C/i}}
\triangleq   g_R^{(2)}(\alpha_{ii},{\alpha}_{\rm{Tr}},{\alpha}_{\rm{R/i}},{\alpha}_{\rm{C/i}}, \Vert{\bm \alpha}\Vert),\label{eq:Vertex-HetGNN-linear-2b}
\end{split}
\end{align}
\end{subequations}
where $(a)$ in both \eqref{eq:Vertex-HetGNN-linear-2a} and \eqref{eq:Vertex-HetGNN-linear-2b} is obtained by exchanging the operation order of the linear function and the summation function, i.e., $\sum g_T^{(1)}(\cdot)=g_T^{(1)}(\sum(\cdot))$ and $\sum g_R^{(1)}(\cdot)=g_R^{(1)}(\sum(\cdot))$.

With similar derivations, we can obtain that the action of vertex $T_i$ is a linear function of the input features, i.e.,
\vspace{-4mm}\begin{equation} \label{eq:Vertex-HetGNN-linear-output}
\small
\vspace{-4mm}
{\hat x}_{i,T}\triangleq g_T(\alpha_{ii},{\alpha}_{\rm{Tr}},{\alpha}_{\rm{R/i}},{\alpha}_{\rm{C/i}}, \Vert{\bm \alpha}\Vert),
\end{equation}
which does not depend on $\alpha_{ij}~(i\neq j)$.
%. Hence, when inputting ${\bm \alpha}_{[1]}$ or ${\bm \alpha}_{[2]}$, the outputs of the Vertex-GNN are identical.
This indicates that the degenerated vanilla Vertex-GNN$_{\sf ls}^{\sf het}$ cannot distinguish the individual interference channel gains in ${\bm \alpha}$. As a result, the GNN may yield the same action for different input features ${\bm \alpha}_{[1]}$ and ${\bm \alpha}_{[2]}$.
\vspace{-4mm}
\begin{remark} \label{remark:linear-GNNs}
	\emph{
		Analogously, when the combination functions are linear, the action of vertex $D_i$ of the degenerated vanilla Vertex-GNN$_{\sf ls}^{\sf undir}$ (i.e., ${\hat x}_{i,V}$), the action of vertex $D_i$ of the degenerated vanilla Edge-GNN$_{\sf ls}^{\sf undir}$ (i.e., ${\hat x}_{i,E}$), and the action of $\rm sig$ edge $(i,i)$ of the degenerated vanilla Edge-GNN$_{\sf ls}^{\sf het}$ (i.e., ${\hat x}_{i,S}$) can also be expressed in the form as \eqref{eq:Vertex-HetGNN-linear-output}.}
\end{remark}

\vspace{-2mm}
\emph{Linear processor and non-linear combinator:} From \eqref{eq: node-GNN} and \eqref{eq:Node-HetGNN-update}, we can see that four linear functions with different weight matrices are required in the vanilla Vertex-GNN$_{\sf ls}^{\sf het}$. In particular, the processing functions are respectively used for (a) $\rm tx$ vertex aggregating information from $\rm rx$ vertex and $\rm sig$ edge, (b) $\rm tx$ vertex aggregating information from $\rm rx$ vertex and $\rm inf$ edge, (c) $\rm rx$ vertex aggregating information from $\rm tx$ vertex and $\rm sig$ edge, and (d) $\rm rx$ vertex aggregating information from $\rm tx$ vertex and $\rm inf$ edge, which are respectively
\begin{equation}\label{linearpro}
\small
\begin{split}
q({\bf d}_{i,R}^{(l-1)},\alpha_{ii};{\bf W}_{RS}^{(l)})={\bf Q}_{RS}^{(l)}{\bf d}_{i,R}^{(l-1)}+{\bf P}_{RS}^{(l)}\alpha_{ii},~~
q({\bf d}_{j,R}^{(l-1)},\alpha_{ij};{\bf W}_{RI}^{(l)})={\bf Q}_{RI}^{(l)}{\bf d}_{j,R}^{(l-1)}+{\bf P}_{RI}^{(l)}\alpha_{ij},\\
q({\bf d}_{i,T}^{(l-1)},\alpha_{ii};{\bf W}_{TS}^{(l)})={\bf Q}_{TS}^{(l)}{\bf d}_{i,T}^{(l-1)}+{\bf P}_{TS}^{(l)}\alpha_{ii},~~
q({\bf d}_{j,T}^{(l-1)},\alpha_{ji};{\bf W}_{TI}^{(l)})={\bf Q}_{TI}^{(l)}{\bf d}_{j,T}^{(l-1)}+{\bf P}_{TI}^{(l)}\alpha_{ji}.
\end{split}
\end{equation}
Then, the aggregated outputs of $T_i$ and $R_i$ in \eqref{eq:Node-HetGNN-update} can be respectively rewritten as,
\begin{subequations} \label{eq:processing-1}
\small
\begin{align}
\begin{split}
{\bf a}_{i,T}^{(l)}&=q({\bf d}_{i,R}^{(l-1)},\alpha_{ii};{\bf W}_{RS}^{(l)})+\sum_{j=1,j\neq i}^{K}q({\bf d}_{j,R}^{(l-1)},\alpha_{ij};{\bf W}_{RI}^{(l)})\\
&={\bf Q}_{RS}^{(l)}{\bf d}_{i,R}^{(l-1)}+{\bf Q}_{RI}^{(l)}\sum_{j=1,j\neq i}^{K}{\bf d}_{j,R}^{(l-1)}+{\bf P}_{RS}^{(l)}\alpha_{ii}+{\bf P}_{RI}^{(l)}{\alpha}_{\rm{R/i}},
\end{split}\\
\begin{split}
{\bf a}_{i,R}^{(l)}&=q({\bf d}_{i,T}^{(l-1)},\alpha_{ii};{\bf W}_{TS}^{(l)})+\sum_{j=1,j\neq i}^{K}q({\bf d}_{j,T}^{(l-1)},\alpha_{ji};{\bf W}_{TI}^{(l)})\\
&={\bf Q}_{TS}^{(l)}{\bf d}_{i,T}^{(l-1)}+{\bf Q}_{TI}^{(l)}\sum_{j=1,j\neq i}^{K}{\bf d}_{j,T}^{(l-1)}+{\bf P}_{TS}^{(l)}\alpha_{ii}+{\bf P}_{TI}^{(l)}{\alpha}_{\rm{C/i}}.
\end{split}
\end{align}
\end{subequations}

Since ${d}_{k,T}^{(0)}={d}_{k,R}^{(0)}=0 $,
%${\bf a}_{i,T}^{(1)}$, ${\bf a}_{i,R}^{(1)}$ as well as ${\bf d}_{i,T}^{(1)}$ and ${\bf d}_{i,R}^{(1)}$ also depend on $\mathcal{A}_{\rm{diag}},\mathcal{A}_{\rm{CR}}$, respectively.
${\bf a}_{i,T}^{(1)}$ and ${\bf d}_{i,T}^{(1)}$ depend on ${\alpha}_{ii}$ and ${\alpha}_{\rm R/i}$, and ${\bf a}_{i,R}^{(1)}$ and ${\bf d}_{i,R}^{(1)}$ depend on ${\alpha}_{ii}$ and ${\alpha}_{\rm C/i}$.
With similar derivations, it can be shown that the outputs of the GNN depend on $\mathcal{A}_{\rm{diag}}$ and $\mathcal{A}_{\rm{CR}}$, which are respectively composed of ${\alpha}_{ii}$, ${\alpha}_{\rm R/i}$ and ${\alpha}_{\rm C/i}, i=1, \cdots, K$. If $\sigma(\cdot)$ in the combination function is replaced by a FNN,
then it is not hard to show that ${\hat x}_{i,T}$ has the same form as in \eqref{eq:Vertex-HetGNN-ft}. This means that the information of interference channel gains $\alpha_{ij}, i\neq j$ is lost after the
linear processing. As a consequence, the GNN cannot distinguish  ${\bm\alpha}_{[1]}$ and  ${\bm\alpha}_{[2]}$.

\vspace{-0.1mm}
\emph{Non-linear processor and linear/non-linear combiner:} When the processors in the vanilla Vertex-GNN$_{\sf ls}^{\sf het}$ are replaced by non-linear functions (say FNNs), the aggregated outputs of $T_i$ and $R_i$ after passing through the sum-pooling become,\vspace{-2mm}
\begin{subequations} \label{eq:processing-2}
\small
\begin{align}
\begin{split}
{\bf a}_{i,T}^{(l)}&={\sf FNN}_{RS}\Big({\bf d}_{i,R}^{(l-1)},\alpha_{ii}\Big)+\sum_{j=1,j\neq i}^{K}{\sf FNN}_{RI}\Big({\bf d}_{j,R}^{(l-1)},\alpha_{ij}\Big),
\end{split}\\
\begin{split}
{\bf a}_{i,R}^{(l)}&={\sf FNN}_{TS}\Big({\bf d}_{i,T}^{(l-1)},\alpha_{ii}\Big)+\sum_{j=1,j\neq i}^{K}{\sf FNN}_{TI}\Big({\bf d}_{j,T}^{(l-1)},\alpha_{ji}\Big).
\end{split}
\end{align}
\end{subequations}
Since $\sum {\sf FNN}(\cdot)\neq{\sf FNN}(\sum(\cdot))$ and ${\bf d}_{k,T}^{(0)}={\bf d}_{k,R}^{(0)}=0$, ${\bf a}_{i,T}^{(1)}$ depends on $\mathcal{A}_{i*}\triangleq\{\alpha_{i1},\cdots,\alpha_{iK}\}$, and ${\bf a}_{i,R}^{(1)}$ depends on $\mathcal{A}_{*i}\triangleq\{\alpha_{1i},\cdots,\alpha_{Ki}\}$. After the combiner (no matter linear or non-linear), ${\bf d}_{i,T}^{(1)}$ and ${\bf d}_{i,R}^{(1)}$  respectively depend on  $\mathcal{A}_{i*}$ and $\mathcal{A}_{*i}$.  It can be shown with similar derivations that the outputs of the GNN depend on $\mathcal{A}_{\sf ind}$, which is the set of all the elements in ${\bm \alpha}$. In other words, the outputs of the GNN depend on $\alpha_{ij}$, and hence the GNN can distinguish  ${\bm\alpha}_{[1]}$ and  ${\bm\alpha}_{[2]}$.
%This indicates that the Vertex-GNN$_{\sf ls}^{\sf het}$ is more expressive than the vanilla Vertex-GNN$_{\sf ls}^{\sf het}$ by using FNNs as processing functions.

Similarly, we can show that the expressive power of Vertex-GNN$_{\sf ls}^{\sf undir}$ is able to be improved by using FNN for processing, but cannot be improved by using FNN for combining.

\subsubsection{Edge-GNNs}
Since the outputs of all GNNs for link scheduling depend on $\alpha_{ii}, i=1,\cdots, K$ when the processing and combination functions are linear
as in Remark \ref{remark:linear-GNNs}, we only analyze whether they depend on individual interference channel gains in the following. According to Remark \ref{remark:linear-GNNs} and the analysis in Section \ref{sec:differences in theory}, it is the non-linear combination functions that help the vanilla Edge-GNN$_{\sf ls}$ distinguish individual interference channels. Since there are two types of combination functions in each layer of the vanilla Edge-GNN$_{\sf ls}^{\sf het}$ as shown in \eqref{eq:Edge_HetGNN_update}, in the following we analyze which combiner helps the vanilla Edge-GNN$_{\sf ls}^{\sf het}$  distinguish $\alpha_{ij}, i\neq j$.

Since ${d}_{i,S}^{(0)}=\alpha_{ii}$ and ${d}_{ij,I}^{(0)}=\alpha_{ij}$, the combination functions of the vanilla Edge-GNN$_{\sf ls}^{\sf het}$ for updating the hidden representations of $\rm sig$ edge $(i,i)$ and $\rm int$ edge $(i,j)$ in the first layer can be obtained from \eqref{eq:Edge_HetGNN com} and \eqref{eq:Edge_HetGNN inter} as,
\begin{subequations} \label{eq:Edge-HetGNN-CB-1}
\small
\begin{align}
\begin{split}\label{eq:Edge-HetGNN-CB-S}
{\bf d}_{i,S}^{(1)}&=\sigma_S\Big( {\bf U}_S^{(1)}{d}_{i,S}^{(0)}+{\bf a}_{i,S}^{(1)} \Big) \triangleq {\sf CB}_{S}^{(1)}\Big({d}_{i,S}^{(0)},{\bf a}_{i,S}^{(1)}\Big)\\
&\overset{(a)}{=} {\sf CB}_{S}^{(1)}\Big(\alpha_{ii},\sum_{k=1,k\neq i}^K{\bf Q}_{T}^{(1)}\alpha_{ik}+\sum_{k=1,k\neq i}^K{\bf Q}_{R}^{(1)}\alpha_{ki}\Big)
\overset{(b)}{=}{\sf CB}_{S}^{(1)}\Big(\alpha_{ii},{\bf Q}_T^{(1)}{\alpha}_{\rm{R/i}}+{\bf Q}_R^{(1)}{\alpha}_{\rm{C/i}}\Big),
\end{split}\\
\begin{split} \label{eq:Edge-HetGNN-CB-I}
{\bf d}_{ij,I}^{(1)}&=\sigma_I \Big( {\bf U}_I^{(1)}{d}_{ij,I}^{(0)}+{\bf a}_{ij,I}^{(1)} \Big) \triangleq {\sf CB}_{I}^{(1)}\Big({d}_{ij,I}^{(0)},{\bf a}_{ij,I}^{(1)}\Big)\\
&\overset{(a)}{=}{\sf CB}_{I}^{(1)}\Big(\alpha_{ij},{\bf Q}_{ST}^{(1)}\alpha_{ii}+ \sum_{k=1,k\neq\{i,j\}}^K {\bf Q}_{IT}^{(1)}\alpha_{ik}+{\bf Q}_{SR}^{(1)}\alpha_{jj}+\sum_{k=1,k\neq\{i,j\}}^K {\bf Q}_{IR}^{(1)}\alpha_{kj}\Big)\\
&\overset{(b)}{=}{\sf CB}_{I}^{(1)}\Big(\alpha_{ij},{\bf Q}_{ST}^{(1)}\alpha_{ii}+ {\bf Q}_{IT}^{(1)}{\alpha}_{\rm{R/i}}+{\bf Q}_{SR}^{(1)}\alpha_{jj}+{\bf Q}_{IR}^{(1)}{\alpha}_{\rm{C/j}}-({\bf Q}_{IT}^{(1)}+{\bf Q}_{IR}^{(1)})\alpha_{ij}\Big)
\triangleq J_{ij}({\alpha_{ij}}),
\end{split}
\end{align}
\end{subequations}
where
$(a)$ in \eqref{eq:Edge-HetGNN-CB-S} or \eqref{eq:Edge-HetGNN-CB-I} comes from substituting ${d}_{i,S}^{(0)}$ and ${d}_{ij,I}^{(0)}$ into ${\bf a}_{i,S}^{(1)}$ and ${\bf a}_{ij,I}^{(1)}$, $(b)$ in \eqref{eq:Edge-HetGNN-CB-S} or \eqref{eq:Edge-HetGNN-CB-I} is obtained by exchanging the order of matrix multiplication and summation operations. $J_{ij}(\cdot)$ is expressed as a function of only $\alpha_{ij}$, since we are concerned whether or not the output of GNN depends on individual interference channel gains.

\vspace{-1mm}If $\sigma_{I}(\cdot)$ is omitted, then ${\sf CB}_{I}^{(1)}(\cdot)$ and $J_{ij}(\cdot)$ in \eqref{eq:Edge-HetGNN-CB-I} become linear functions. Then, when $l=2$, the first term in \eqref{eq:Edge_HetGNN com}  becomes $\sum_{k=1,k\neq i}^K{\bf Q}_{T}^{(2)}{\bf d}_{ik,I}^{(1)}={\bf Q}_{T}^{(2)}\sum_{k=1,k\neq i}^KJ_{ik}(\alpha_{ik})$ that depends on ${\alpha}_{\rm{R/i}}$, and the second term in \eqref{eq:Edge_HetGNN com} becomes $\sum_{k=1,k\neq i}^K{\bf Q}_{R}^{(2)}{\bf d}_{ki,I}^{(1)}$ that depends on ${\alpha}_{\rm{C/i}}$. Hence, ${\bf d}_{i,S}^{(2)}$ depends on ${\alpha}_{\rm{R/i}}$ and ${\alpha}_{\rm{C/i}}$.
With similar analysis, it can be shown that the action on edge $(i,i)$ (i.e., $\hat x_{i,S}$) still depends on ${\mathcal A}_{\rm CR}$,
%${\bm\alpha}_{\rm{R/i}}$ and ${\bm\alpha}_{\rm{C/i}}$,
i.e., the corresponding Edge-GNN cannot distinguish $\alpha_{ij}, i \neq j$.

With $\sigma_{I}(\cdot)$, ${\sf CB}_{I}^{(1)}(\cdot)$ and $J_{ij}(\cdot)$ in \eqref{eq:Edge-HetGNN-CB-I} are non-linear. Since $\sum J_{ij}(\cdot)\neq J_{ij}(\sum(\cdot))$, $\sum_{k=1,k\neq i}^K{\bf Q}_{T}^{(2)}{\bf d}_{ik,I}^{(1)}$ depends on $\alpha_{ik}, k\neq i$ rather than ${\alpha}_{\rm{R/i}}$, and $\sum_{k=1,k\neq i}^K{\bf Q}_{R}^{(2)}{\bf d}_{ki,I}^{(1)}$ depends on $\alpha_{ki}$ rather than ${\alpha}_{\rm{C/i}}$.
Hence, ${\bf d}_{i,S}^{(2)}$ depends on $\mathcal{A}_{i*}$ and $\mathcal{A}_{*i}$ no matter if $\sigma_S(\cdot)$ is omitted. Similarly, we can show that the action of edge $(i,i)$ depends on $\mathcal{A}_{\sf ind}$, i.e., the corresponding GNN can distinguish $\alpha_{ij}, i \neq j$.

With similar analysis, we can show that it is $\sigma_{I}(\cdot)$ (instead of $\sigma_{S}(\cdot)$) in the Edge-GNN$_{\sf ls}^{\sf het}$ that enables the GNN to distinguish $\alpha_{ij}, i \neq j$ and enhances its expressive power.
%This result tells us if we want to improve the expressive ability of the vanilla Edge-GNN$_{\sf ls}^{\sf het}$, it is better to replace $\sigma_I(\cdot)$ rather than $\sigma_S(\cdot)$ by FNN which can approximate arbitrary non-linear functions.

In a nutshell, non-linear processors help improve the expressive power of the vanilla Vertex-GNNs (i.e., Vertex-GNN$_{\sf ls}^{\sf het}$, Vertex-GNN$_{\sf ls}^{\sf undir}$, and Vertex-GNN$_{\sf p}^{\sf het}$). Non-linear combiners for updating the $\rm int$ edges (say $\sigma_{I}(\cdot)$) help the vanilla Edge-GNN$_{\sf ls}^{\sf het}$ to distinguish $\alpha_{ij}, i \neq j$. Since there is only one type of combination functions in the vanilla Edge-GNN$_{\sf ls}^{\sf undir}$ and Edge-GNN$_{\sf p}^{\sf het}$, the non-linearity of all combination functions helps improve the expressive power of the two Edge-GNNs. The expressive power of these GNNs is summarized in Table \ref{table: impact_linearity}.

\begin{table}[!htb]
	\centering
	\small
	\vspace{-6mm}
	\caption{expressive power of GNNs}\label{table: impact_linearity}\vspace{-3mm}
	\begin{threeparttable}
	\begin{tabular}{|c|c|c|c|c|}
		\hline
		\multicolumn{2}{|c|}{\textbf{GNNs}}&\textbf{Processing functions}&\textbf{Combination functions}&\textbf{expressive power}\\
		\hline
		\multicolumn{2}{|c|}{\multirow{4}{*}{Vertex-GNNs}}&\multirow{2}{*}{linear}&linear&\multirow{2}{*}{weak}\\
		\cline{4-4}
		\multicolumn{2}{|c|}{}&&non-linear&\\
		\cline{3-5}
		\multicolumn{2}{|c|}{}&\multirow{2}{*}{non-linear}&linear&\multirow{2}{*}{strong}\\
		\cline{4-4}
		\multicolumn{2}{|c|}{}&&non-linear&\\
		\hline
		\multirow{4}{*}[-3.5ex]{Edge-GNNs}&Edge-GNN$_{\sf ls}^{\sf undir}$&\multirow{2}{*}{linear}&\multirow{2}{*}{non-linear}&\multirow{2}{*}{strong}\\
		\cline{2-2}
		&Edge-GNN$_{\sf p}^{\sf het}$&&& \\
		\cline{2-5}
		&\multirow{2}{*}[-1.5ex]{Edge-GNN$_{\sf ls}^{\sf het}$}&\multirow{2}{*}[-1.5ex]{linear}&\tabincell{c}{${\sf CB}_{S}\tnote{1}$ is non-linear \\ ${\sf CB}_{I}\tnote{1}$ is linear}& weak\\
		\cline{4-5}
		&&&\tabincell{c}{${\sf CB}_{S}$ is linear or non-linear \\ ${\sf CB}_{I}$ is non-linear}& strong\\
		\hline
	\end{tabular}
\begin{tablenotes}
	\footnotesize
	\item[1]${\sf CB}_S$ and ${\sf CB}_I$ are respectively the combination functions for updating the hidden representations of $\rm sig$ and $\rm int$ edges.
\end{tablenotes}\vspace{-6mm}
\end{threeparttable}
\end{table}

\vspace{-4mm}
\subsection{Impact of Dimension}\label{sec:dimension}\vspace{-2mm}
%Now, we analyze how the dimensions of the processing and combination functions affect the expressive power of the Vertex-GNNs with FNN for processing and the vanilla Edge-GNNs.

When a Vertex-GNN learns the link scheduling or power control policy over ${\cal G}_{\sf{ls}}^{\sf{het}}$ that consists of $2K$ vertices, there are $2K$ update equations, where two of them for updating ${\bf d}_{i,T}^{(l)} \in \mathbb{R}^{M^{(l)}}$ and ${\bf d}_{i,R}^{(l)} \in \mathbb{R}^{M^{(l)}}$ are shown in \eqref{eq:HetGNN_detail} and $M^{(l)}$ is an integer that is a hyper-parameter. The representation of all vertices of the Vertex-GNN$_{\sf{ls}}^{\sf{het}}$ in the $l$th layer is $[{\bf d}_{1,T}^{(l)},\cdots,{\bf d}_{K,T}^{(l)},{\bf d}_{1,R}^{(l)},\cdots,{\bf d}_{K,R}^{(l)}]\in \mathbb{R}^{M^{(l)}\times2K}$, whose dimension is higher than the dimension of action vector ${\bf\hat x}_T\in{\mathbb{R}^{K}}$. Analogously, the representation of all vertices of the Vertex-GNN$_{\sf{ls}}^{\sf{undir}}$ in the $l$th layer is $[{\bf d}_{1,V}^{(l)},\cdots,{\bf d}_{K,V}^{(l)}]\in\mathbb{R}^{M^{(l)}\times K}$, whose dimension is no less than the dimension of action vector ${\bf\hat x}_V\in{\mathbb{R}^{K}}$. When Edge-GNNs are used to learn the two policies, the dimension of the representation of all edges is higher than $K$.
%the representations of all edges are with higher dimension than ${\bf\hat x}_V$.
Since channel information is not compressed with the high representation dimension, the dimensions of the GNNs for these two policies do not affect their expressive power. In the sequel, we only analyze the impact of the dimensions for the GNNs learning the precoding policy.

%The precoding policy ${\bf V}^{*}= F_{\sf p}({\bf H})$ is a mapping from a $N \times K$-dimensional complex matrix to another $N \times K$-dimensional complex matrix.
%Due to the ``phase invariance property'' found in  \cite{phase_precoding2022}, more than one channel matrix can yield the same optimal precoding matrix. When the channel matrix is pre-processed by setting the phases of elements in its first row and first column as zeros, the optimal precoding policy becomes a one-to-one mapping.

When learning over ${\cal G}_{\sf{p}}^{\sf{het}}$ consisting of $N+K$ vertices and $NK$ edges, the hidden representations in the $l$th layer of the Vertex-GNN are $[{\bf d}_{1,A}^{(l)},\cdots,{\bf d}_{N,A}^{(l)}]\in{\mathbb R}^{M_{A}^{(l)}\times N}$ and $[{\bf d}_{1,U}^{(l)},\cdots,{\bf d}_{K,U}^{(l)}]\in{\mathbb R}^{M_{U}^{(l)}\times N}$, and that of the Edge-GNN is $[{\bf d}_{11,E}^{(l)},\cdots,{\bf d}_{NK,E}^{(l)}]\in{\mathbb R}^{M_{E}^{(l)}\times NK}$,
%respectively with dimension $NM_A^{(l)}+KM_U^{(l)}$ and $NKM_E^{(l)}$,
where $M_A^{(l)}$, $M_U^{(l)}$, and $M_E^{(l)}$ are hyper-parameters respectively representing the dimensions of the hidden representations for each antenna vertex, each user vertex, and each edge.
Since the precoding policy maps ${\bf H} \in \mathbb{C}^{N\times K}$ into ${\bf V}^* \in \mathbb{C}^{N\times K}$, the channel information will be lost if the vertex representation is with low dimension.
%respectively the output dimensions of the combiners for updating the representations of antenna and user vertices, and $M_E^{(l)}$ is the output dimension of the combiner for updating the representations of edges.
%If the dimension of hidden representation of the GNN is less than the dimension of the channel matrix, then the channel information will be lost and not recoverable.
%In the sequel, the vectors, matrices and tensors reside in the real number field.

%Before analyzing the impact of output dimensions of the processors and combiners on the expressive power of Vertex-GNN$_{\sf p}^{\sf het}$ and Edge-GNN$_{\sf p}^{\sf het}$,
%we first introduce two properties.

%{ \bf Property 1:} For a composite function $\phi(\cdot)=\varphi_2(\varphi_1(\cdot))$ where $\varphi_1(\cdot)$ and $\varphi_2(\cdot)$ are functions, if $\phi(\cdot)$ is one-to-one mapping, then $\varphi_1(\cdot)$ must be one-to-one  mapping \cite{set_theory2016}.

%{\bf Property 2:} If there exists a continuous function that is a one-to-one mapping from ${\mathbb R}^n$ to ${\mathbb R}^m$, then $n\leq m$ \cite{Aggreration2020Cambridge}.

\subsubsection{Vertex-GNN$_{\sf p}^{\sf het}$}
\begin{figure}[!htb]\vspace{-7mm}
	\centering
	\begin{minipage}[t]{0.95\linewidth}
		\subfigure[Vertex-GNN$_{\sf p}^{\sf het}$]{
			\includegraphics[width=\textwidth]{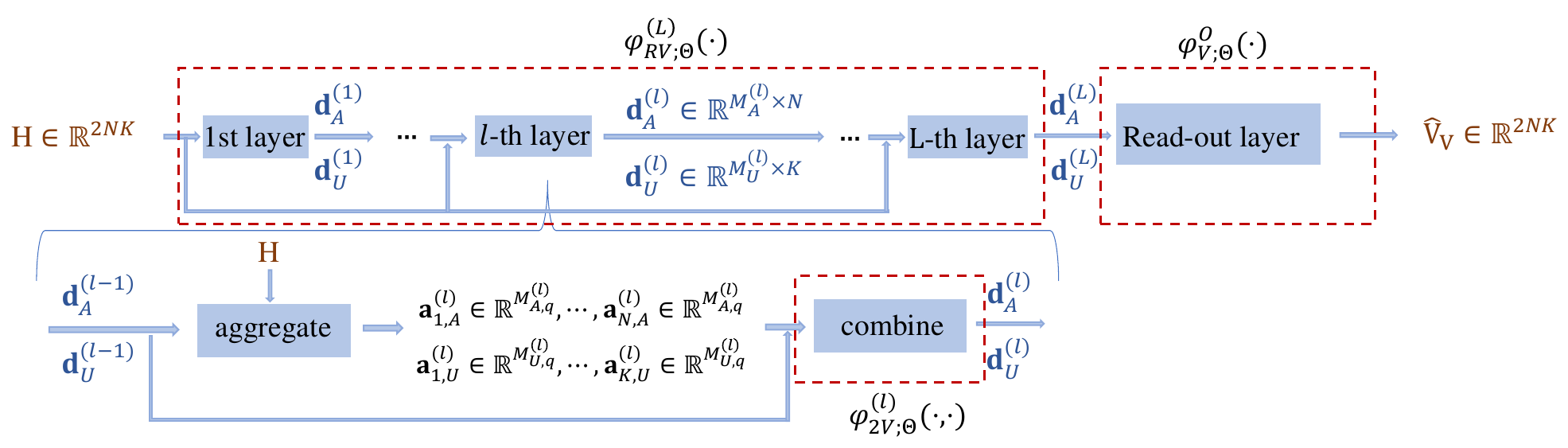}}
	\end{minipage}
	\begin{minipage}[t]{0.83\linewidth}	
		\subfigure[Edge-GNN$_{\sf p}^{\sf het}$]{
			\includegraphics[width=\textwidth]{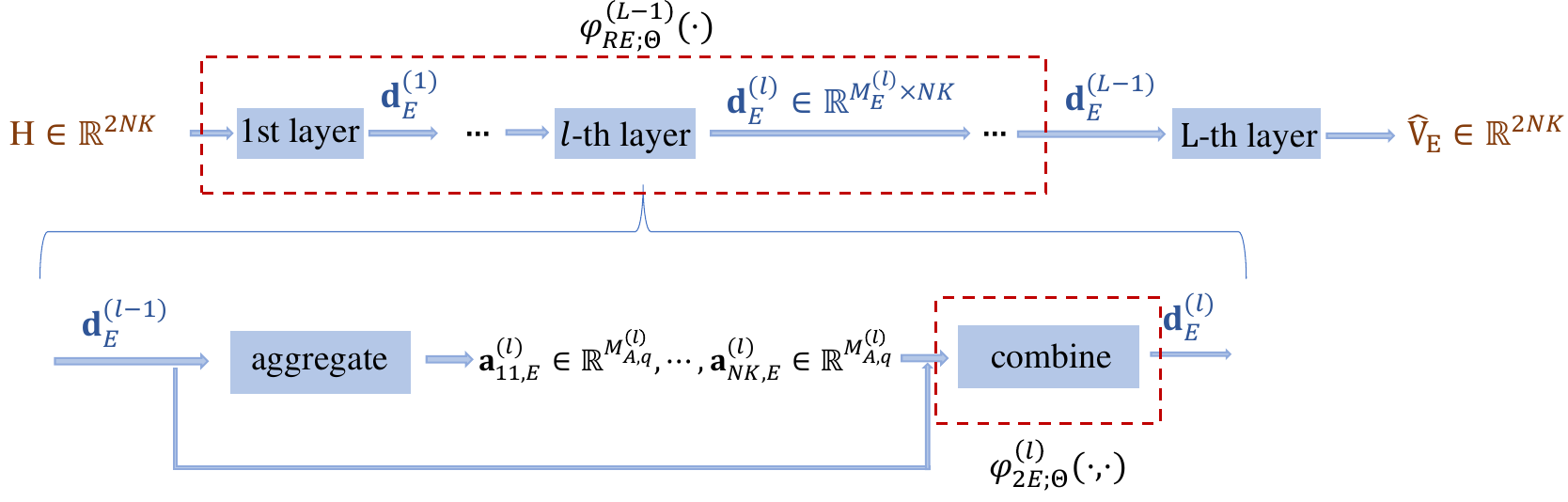}}
	\end{minipage}\vspace{-2mm}
	\caption{Structure of (a) Vertex-GNN$_{\sf p}^{\sf het}$ and (b) Edge-GNN$_{\sf p}^{\sf het}$.}\label{fig:GNN-p}
	\vspace{-3mm}
\end{figure}

%When a Vertex-GNN learns the precoding policy over ${\cal G}_{\sf{p}}^{\sf{het}}$ that consists of $N+K$ vertices, there are $N+K$ update equations. The representation of all vertices of the Vertex-GNN$_{\sf p}^{\sf het}$ in the $l$th layer is $[{\bf d}_{1,A}^{(l)},\cdots,{\bf d}_{N,A}^{(l)},{\bf d}_{1,U}^{(l)},\cdots,{\bf d}_{K,U}^{(l)}]\in\mathbb{R}^{NM_A^{(l)}+KM_U^{(l)}}$, where $M_A^{(l)}$ and $M_U^{(l)}$ are hyper-parameters.
As shown in Fig. \ref{fig:GNN-p}(a), the combination outputs of all antenna and user vertices in the $l$th layer are ${\bf d}_A^{(l)}\triangleq [{\bf d}_{1,A}^{(l)},\cdots,{\bf d}_{N,A}^{(l)}] \in \mathbb{R}^{M_A^{(l)}\times N}$ and ${\bf d}_U^{(l)}\triangleq [{\bf d}_{1,U}^{(l)},\cdots,{\bf d}_{K,U}^{(l)}] \in \mathbb{R}^{M_U^{(l)}\times K}$, the aggregation outputs of $A_i$ and $U_j$ in the $l$th layer are ${\bf a}_{i,A}^{(l)}  \in \mathbb{R}^{M_{A,q}^{(l)}}$, and ${\bf a}_{j,U}^{(l)} \in \mathbb{R}^{M_{U,q}^{(l)}}$, respectively, where
$M_{A,q}^{(l)}$ and $M_{U,q}^{(l)}$ are hyper-parameters.
%, $M_A^{(l)}$, and $M_U^{(l)}$ are respectively the output dimensions of the processing functions and the combination functions for updating the representations of antenna and user vertices in the $l$th layer.

The GNN can be expressed as composite functions $\varphi_{V;\Theta}^O(\varphi_{RV;\Theta}^{(L)}({\bf H}))$, where $\varphi(\cdot)$ with sub-script $\Theta$ denotes the functions with trainable parameters, $\varphi_{RV;\Theta}^{(L)}({\bf H})$ maps ${\bf H}$ (i.e., the input feature of the GNN) to $[{\bf d}_{A}^{(L)},{\bf d}_{U}^{(L)}]$ (i.e., the vertex representations in the $L$th layer), and $\varphi_{V;\Theta}^O(\cdot)$ maps $[{\bf d}_{A}^{(L)},{\bf d}_{U}^{(L)}]$ to ${\bf\hat V}_{V} $ (i.e., the actions on edges).
%According to {\bf Property 1}, if the GNN can represent one-to-one mappings, then $\varphi_{RV;\Theta}^{(L)}({\bf H})$ should also be able to represent one-to-one mappings. 	
In order for the GNN not compressing channel information, $\varphi_{RV;\Theta}^{(L)}({\bf H})$ should not compress the channel dimension. Otherwise, the GNN may not differentiate all channel matrices.

Since matrix ${\bf X}=[x_{ij}]_{m\times n}$ can be vectorized as a $(mn)$-dimensional vector $\overline{\bf x} \in {\mathbb R}^{mn}$, a block matrix $[{\bf X}_1,{\bf X}_2]$ with ${\bf X}_1=[x_{1,ij}]_{m_1\times n_1}$ and ${\bf X}_2=[x_{2,ij}]_{m_2\times n_2}$ seems able to be vectorized to a $(m_1+m_2)(n_1+n_2)$-dimensional vector, because the block matrix before vectorization can be expressed as $\begin{pmatrix} {\bf X}_1&{[0]}_{m_1\times n_2}\\{[0]}_{m_2\times n_1}&{\bf X}_2\end{pmatrix}$. However, since zero matrices do not occupy space, the dimension of the vectorized block matrix is $(m_1n_1+m_2n_2)$.
Hence, the output dimension of $\varphi_{RV;\Theta}^{(L)}(\bf H)$ is ${(NM_A^{(L)}+K M_U^{(L)})}$. ${\bf H} \in\mathbb{C}^{N\times K}$ can be vectorized as a $2NK$-dimensional real vector.
%According to {\bf Property 2}, if $\varphi_{RV;\Theta}^{(L)}({\bf H})$ can represent one-to-one mappings, then the dimensions should satisfy $NM_A^{(L)}+KM_U^{(L)}\geq 2NK$.
In order for $\varphi_{RV;\Theta}^{(L)}({\bf H})$ not compressing channel information, the output dimension of $\varphi_{RV;\Theta}^{(L)}(\bf H)$ should be no less than the dimension of $\bf H$, i.e., $NM_A^{(L)}+KM_U^{(L)}\geq 2NK$.
%If $NM_A^{(L)}+KM_U^{(L)}\geq 2NK$, then $\varphi_{RV;\Theta}^{(L)}({\bf H})$ will not compress channel information.

%\vspace{-18mm}
$\varphi_{RV;\Theta}^{(L)}({\bf H})$ can further be expressed as the following composite functions $\varphi_{RV;\Theta}^{(L)}({\bf H})\triangleq {\varphi}_{2V;\Theta}^{(L)}({\varphi}_{RV;\Theta}^{(L-1)}({\bf H}),{\varphi}_{AV;\Theta}^{(L)}({\bf H}))\triangleq \varphi_{2V;\Theta}^{(L)}(\varphi_{1V;\Theta}^{(L)}({\bf H}))$, where $\varphi_{2V;\Theta}^{(L)}(\cdot,\cdot)$ is the combination function to output the vertex representations in the $L$th layer,
${\varphi}_{AV;\Theta}^{(L)}(\bf H)$ denote the functions that map $\bf H$ to the aggregated output in the $L$th layer (i.e., $[{\bf a}_{1,A}^{(L)},\cdots,{\bf a}_{N,A}^{(L)},{\bf a}_{1,U}^{(L)},\cdots,{\bf a}_{K,U}^{(L)}]$), and $\varphi_{1V;\Theta}^{(L)}({\bf H})$ denotes the functions that map $\bf H$ to $[{\bf d}_{A}^{(L-1)},{\bf d}_{U}^{(L-1)}]$ and $[{\bf a}_{1,A}^{(L)},\cdots,{\bf a}_{N,A}^{(L)},{\bf a}_{1,U}^{(L)},\cdots,{\bf a}_{K,U}^{(L)}]$.
%According to {\bf Property 1}, if $\varphi_{RV;\Theta}^{(L)}({\bf H})$  can represent one-to-one mappings, then $\varphi_{1V;\Theta}^{(L-1)}({\bf H})$ (and hence ${\varphi}_{RV;\Theta}^{(L-1)}(\bf H)$ and ${\varphi}_{AV;\Theta}^{(L)}(\bf H)$) should also be able to represent one-to-one mappings.
%According to {\bf Property 2}, if ${\varphi}_{RV;\Theta}^{(L-1)}({\bf H})$ and ${\varphi}_{AV;\Theta}^{(L)}({\bf H})$ can represent one-to-one mappings, then the dimensions should satisfy $NM_A^{(L-1)}+KM_U^{(L-1)}\geq 2NK$ and $NM_{A,q}^{(L)}+KM_{U,q}^{(L)}\geq 2NK$.
In order for $\varphi_{RV;\Theta}^{(L)}({\bf H})$ not compressing channel information,  $\varphi_{1V;\Theta}^{(L-1)}({\bf H})$ (and hence ${\varphi}_{RV;\Theta}^{(L-1)}(\bf H)$ and ${\varphi}_{AV;\Theta}^{(L)}(\bf H)$) should not compress channel information. Hence, the output dimensions of ${\varphi}_{RV;\Theta}^{(L-1)}(\bf H)$ and ${\varphi}_{AV;\Theta}^{(L)}(\bf H)$ should not be less than the dimension of ${\bf H}$, i.e., $NM_A^{(L-1)}+KM_U^{(L-1)}\geq 2NK$ and $NM_{A,q}^{(L)}+KM_{U,q}^{(L)}\geq 2NK$.
%Analogously, ${\varphi}_{RV;\Theta}^{(L-1)}({\bf H})\triangleq {\varphi}_{2V;\Theta}^{(L-1)}({\varphi}_{RV;\Theta}^{(L-2)}({\bf H}),{\varphi}_{AV;\Theta}^{(L-1)}({\bf H}))\triangleq {\varphi}_{2V;\Theta}^{(L-1)}({\varphi}_{1V;\Theta}^{(L-1)}({\bf H}))$.
%With similar analysis, we can show that the GNN can represent one-to-one mappings if the following conditions are satisfied,
With similar analysis, we can show that the following conditions should be satisfied for the GNN not losing channel information,
\begin{equation} \label{eq:equation-2}
NM_A^{(l)}+KM_U^{(l)}\geq 2NK,~NM_{A,q}^{(l)}+KM_{U,q}^{(l)}\geq 2NK,~ l=1,2,\cdots,L.
\end{equation}
When one sets $M_A^{(l)}=M_U^{(l)} \triangleq M_d^{(l)}$
and $M_{A,q}^{(l)}=M_{U,q}^{(l)} \triangleq M_{q}^{(l)}$, then the necessary conditions in \eqref{eq:equation-2} can be simplified as $M_d^{(l)}\geq 2NK/(N+K),~M_q^{(l)} \geq 2NK/(N+K),~ l=1,2,\cdots,L$.
%\begin{equation} \label{eq:equation-21}
%M_d^{(l)}\geq 2NK/(N+K),~M_q^{(l)} \geq 2NK/(N+K),~ l=1,2,\cdots,L.
%\end{equation}
%This suggests that the vertex-GNNs are possible to learn one-to-one mappings if these conditions are satisfied.
Further considering the analysis in Section \ref{sec:Influence-q-CB}, the Vertex-GNNs with linear processors cannot differentiate ${\bf H}_{[1]}$ and ${\bf H}_{[2]}$. %Hence, to design a Vertex-GNN$_{\sf p}^{\sf het}$ that can learn one-to-one mappings, the processing functions should be non-linear and the conditions in \eqref{eq:equation-2} should be satisfied.
Hence, to design a Vertex-GNN$_{\sf p}^{\sf het}$ that does not lose channel information, the processing functions should be non-linear and the output dimensions of the processing and combination functions in \eqref{eq:equation-2} should be satisfied.

%If $L_{A,p}\geq K$ and $L_{U,p}\geq N$, \eqref{eq:equation-2} is satisfied. Then, when \eqref{eq:equation-1} is satisfied, by choosing appropriate pooling functions\footnote{For example, the output of the $i$th pooling function outputs the element-wise $i$th largest value of the inputs.}, the Vertex-HetGNN-P can learn the precoding policy. Hence, we provide another necessary condition for Vertex-HetGNN-P to learn the precoding policy: $L_{A,p}\geq K$, $L_{U,p}\geq N$, and \eqref{eq:equation-1} is satisfied.
\vspace{-1mm}
\subsubsection{Edge-GNN$_{\sf p}^{\sf het}$} \label{sec:dimension-Edge}
The structure of the GNN is shown in Fig. \ref{fig:GNN-p}(b), where ${\bf d}_E^{(l)}\triangleq[{\bf d}_{11,E}^{(l)},\cdots,{\bf d}_{NK,E}^{(l)}]\in{\mathbb R}^{M_{E}^{(l)}\times NK}$ is the combination output of all edges, and ${\bf a}_{ij,E}^{(l)}\in {\mathbb R}^{M_{E,q}^{(l)}}$ is the aggregation output of edge $(i,j)$ in the $l$th layer, and $M_{E,q}^{(l)}$ is a hyper-parameter.

The GNN can be expressed as composite functions ${\varphi}_{2E,\Theta}^{(L)}({\varphi}_{RE;\Theta}^{(L-1)}({\bf H}),{\varphi}_{AE;\Theta}^{(L)}({\bf H}))\triangleq{\varphi}_{2E;\Theta}^{(L)}({\varphi}_{1E;\Theta}^{(L)}({\bf H}))$, where ${\varphi}_{2E;\Theta}^{(L)}(\cdot,\cdot)$ is the combination function to output the edge representations in the $L$th layer (i.e., ${\bf\hat V}_E$), ${\varphi}_{RE;\Theta}^{(L-1)}({\bf H})$ denote the functions that map $\bf H$ to the edge representations in the $(L-1)$th layer (i.e., ${\bf d}_E^{(L-1)}$), ${\varphi}_{AE;\Theta}^{(L)}({\bf H})$ denote the functions that map $\bf H$ to the aggregated output in the $L$th layer (i.e., $[{\bf a}_{11,A}^{(L)},\cdots,{\bf a}_{NK,A}^{(L)}]$), and ${\varphi}_{1E;\Theta}^{(L)}({\bf H})$ denote the functions that map $\bf H$ to ${\bf d}_E^{(L-1)}$ and $[{\bf a}_{11,A}^{(L)},\cdots,{\bf a}_{NK,A}^{(L)}]$. %According to {\bf Property 1}, if the GNN can represent one-to-one mappings, then ${\varphi}_{1E;\Theta}^{(L)}({\bf H})$ (and hence ${\varphi}_{RE;\Theta}^{(L-1)}({\bf H})$ and ${\varphi}_{AE;\Theta}^{(L)}({\bf H})$) should be able to represent one-to-one mappings.
%According to {\bf Property 2}, if ${\varphi}_{RE;\Theta}^{(L-1)}({\bf H})$ and ${\varphi}_{AE;\Theta}^{(L)}({\bf H})$ can represent one-to-one mappings, then the dimensions should satisfy $NKM_E^{(L-1)}\geq 2NK$ and $NKM_{E,q}^{(L)}\geq 2NK$, i.e., $M_E^{(L-1)}\geq 2$ and $M_{E,q}^{(L)}\geq 2$.
If the GNN does not lose channel information, then ${\varphi}_{1E;\Theta}^{(L)}({\bf H})$ (and hence ${\varphi}_{RE;\Theta}^{(L-1)}({\bf H})$ and ${\varphi}_{AE;\Theta}^{(L)}({\bf H})$) should not compress channel information. Hence, the output dimensions of ${\varphi}_{RE;\Theta}^{(L-1)}({\bf H})$ and ${\varphi}_{AE;\Theta}^{(L)}({\bf H})$ should not be less than the dimension of $\bf H$, i.e., $M_E^{(L-1)}\geq 2$ and $M_{E,q}^{(L)}\geq 2$. Analogously,
we can show that the following conditions
should be satisfied for the GNN not losing channel information,
\begin{equation} \label{eq:equation-edge}
	M_E^{(l)}\geq 2, ~ l=1,\cdots,L-1; ~M_{E,q}^{(l)}\geq 2, ~ l=1,\cdots,L.
\end{equation}
The dimensions should be at least 2, because $\bf H$ and $\bf V$ are complex matrices.
Since $M_E^{(L)}$ is the dimension of the action vector defined on each edge, we have $M_E^{(L)}=2$.
%This suggests that the edge-GNNs are possible to learn one-to-one mappings if these conditions are satisfied.
Further recalling the analysis in Section \ref{sec:Influence-q-CB},
to design an Edge-GNN$_{\sf p}^{\sf het}$ that does not compress channel information, the combination functions should be non-linear and the conditions in \eqref{eq:equation-edge} should be satisfied.

\vspace{-3mm}\section{Simulation Results}\label{sec: simulation}\vspace{-1mm}
In this section, we validate the previous analyses, and compare the system performance, space, time, and sample complexity of Vertex-GNNs with Edge-GNNs via simulations.

We consider three optimization problems, i.e, the link scheduling problem in \eqref{eq: max sum-rate}, the power control problem in \eqref{eq: max sum-ratepc}, and the precoding problem in \eqref{eq: max sum-rate precoding}.
For the link scheduling and power control problems,
all the D2D pairs are randomly located in a $500~m$ $\times$ $500~m$ squared area. The wireless network parameters are provided in Table \ref{table: D2D params}. The composite channel consists of the path loss generated with the model in \cite{LS2021Lee}, log-normal shadowing with standard deviation of 8 dB, and Rayleigh fading. For the precoding problem, we set $P_{max}=1~W$, and change $\sigma_0^2$ in \eqref{eq: max sum-rate precoding} to adjust SNR.
These simulation setups are considered unless otherwise specified.

\begin{table}[htp!]
	\centering
	\vspace{-5mm}
	\caption{Parameters in Simulation} \label{table: D2D params}
	\vspace{-3mm}
	\footnotesize
	\renewcommand\arraystretch{0.95}
	\begin{tabular}{c|c}
		\hline\hline
		{\textbf{Parameters}} & {\textbf{Values}} \\
		\hline
		\tabincell{c}{\scriptsize D2D distance} & 2-65 m \\
		\hline
		\tabincell{c}{\scriptsize Number of D2D links} & 50 \\
		\hline
		\tabincell{c}{\scriptsize Noise spectral density} & -169 dBm/Hz \\
		\hline
		\tabincell{c}{\scriptsize Bandwidth, Carrier frequency } & 5 MHz, 2.4 GHz\\
		\hline
		\tabincell{c}{\scriptsize Antenna height, Antenna gain} & 1.5 m, 2.5 dBi\\
		\hline
		\tabincell{c}{\scriptsize Transmit power of activation link} & 40 dBm \\
		\hline\hline
	\end{tabular}
	\vspace{-5mm}
\end{table}

\vspace{-0.2mm}

While the GNNs can be trained in a supervised or unsupervised manner,
%. The supervised learning requires labels generated by iterative algorithms (e.g., FPLinQ \cite{FPLinQ_shen2017} for link scheduling, and WMMSE for power control and precoding). When the number of required training samples or the scale of wireless network is large, generating labels is time-consuming. Hence, unless otherwise specified,
we train the GNNs in an unsupervised manner to avoid generating labels that is time-consuming. Then, each sample only contains a channel matrix ${\bm \alpha}=[\alpha_{ij}]_{K\times K}$ that is generated according to the channel model with randomly located D2D pairs or ${\bf H}=[h_{nk}]_{N\times K}$ where each element follows Rayleigh distribution. We generate $5\times10^5$ samples as the training set (the number of samples used for training may be much smaller), and $10^3$ samples as the test set. Adam is used as the optimizer. The loss function is designed as
%\begin{equation} \label{eq:loss-function}
$Loss=-\frac{1}{N_s}\sum_{n=1}^{N_s}\Big(\sum_{k=1}^{K}r_k^n + w_1\sum_{k=1}^{K}\log(y_k^n)+w_2\sum_{k=1}^{K}\log(1-y_k^n)\Big)$,
%\end{equation}
where $N_s$ is the number of training samples, $r_k^n$ is the data rate of the $k$th user and $y_k^n$ is the activate probability of the $k$th D2D link in the $n$th sample, $w_1$ and $w_2$ are weights that need to be tuned. The second and the third terms in the loss function are respectively the penalty for preventing all the links from being closed and being activated. For the link scheduling problem, we set $w_1=10^{-1}$ and $w_2=10^{-4}$. For the power control and precoding problems, $w_1=w_2=0$.

\vspace{-2mm}
We use {sum rate ratio} as the performance metric. It is the ratio of the sum rate achieved by the learned policy to the sum rate achieved by a numerical algorithm (which is FPLinQ \cite{FPLinQ_shen2017} for link scheduling, and WMMSE for power control and precoding). We train each GNN five times. The results are obtained by averaging the sum rate ratios achieved by the learned policies with the five trained GNNs over all test samples. For the link scheduling and power control problems, we only provide the performance of Vertex- and Edge-GNN$_{\sf ls}^{\sf het}$ in the sequel since Vertex- and Edge-GNN$_{\sf ls}^{\sf undir}$ perform very close to them.

All the simulation results are obtained on a computer with a 28-core Intel i9-10904X CPU, a Nvidia RTX 3080Ti GPU, and 64 GB memory.

All the GNNs in the following use the mean-pooling function.

%\vspace{-4mm}
\subsection{Impact of the Non-distinguishable Channel Matrices} \label{sec:valid}
We first validate that the vanilla Vertex-GNN$_{\sf ls}^{\sf het}$ cannot well learn the link scheduling and power control policies, and the vanilla Vertex-GNN$_{\sf p}^{\sf het}$ cannot well learn the precoding policy, due to their weak expressive power.

\begin{figure}[!htb]
	\centering
	\begin{minipage}[t]{0.327\linewidth}
		\subfigure[Link Scheduling]{
			\includegraphics[width=\textwidth]{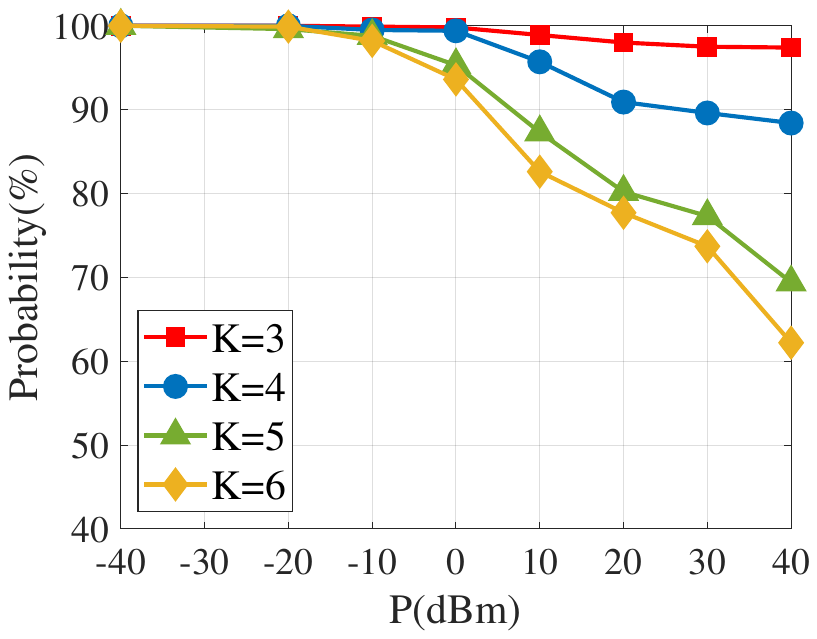}}
	\end{minipage}
	\begin{minipage}[t]{0.327\linewidth}	
		\subfigure[Power Control]{
			\includegraphics[width=\textwidth]{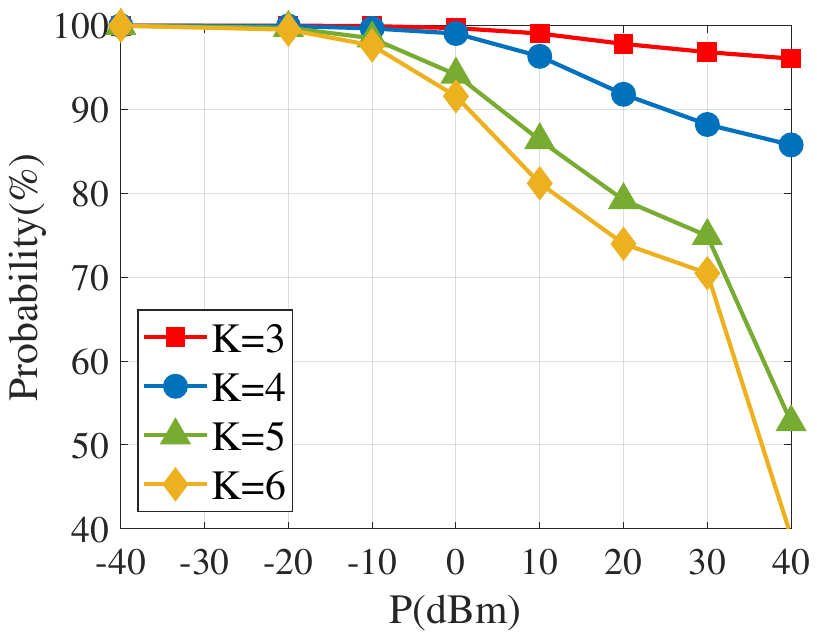}}
	\end{minipage}
    \begin{minipage}[t]{0.325\linewidth}	
	    \subfigure[Precoding]{
		    \includegraphics[width=\textwidth]{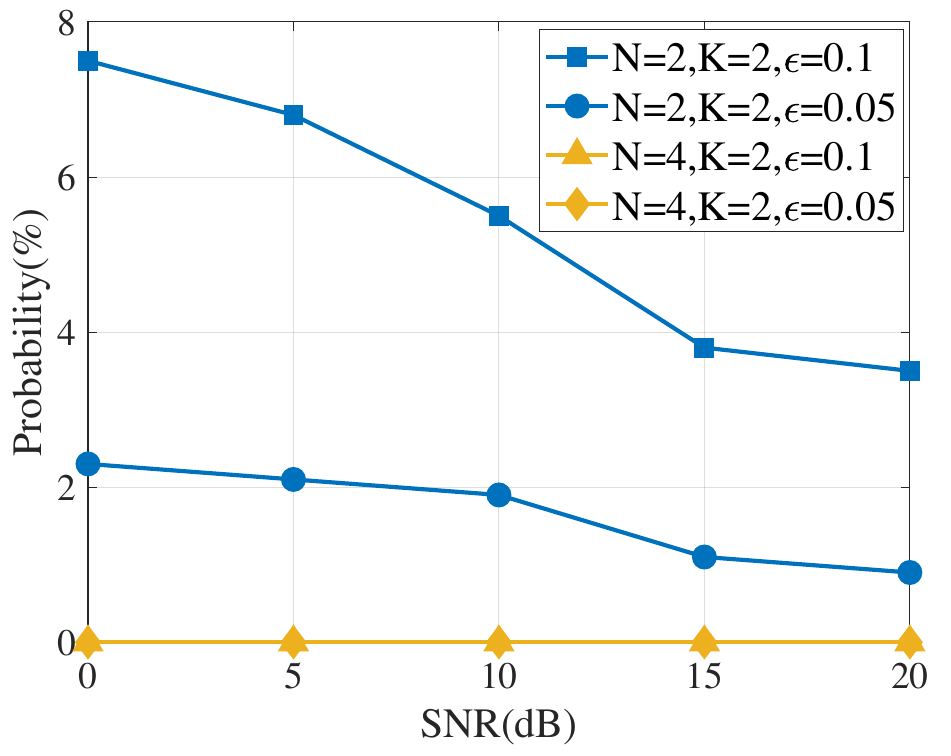}}
    \end{minipage}
	\vspace{-2mm}
	\caption{Probability of (a) $F_{\sf ls}({\bm \alpha}_{[1]})=F_{\sf ls}({\bm \alpha}_{[2]})$, (b) $F_{\sf pc}({\bm \alpha}_{[1]})=F_{\sf pc}({\bm \alpha}_{[2]})$, and (c) $F_{\sf p}({\bf H}_{[1]})=F_{\sf p}({\bf H}_{[2]})$, where ${\bm \alpha}_{[1]}$ and ${\bm \alpha}_{[2]}$ satisfy the conditions in Observation 1, and ${\bf H}_{[1]}$ and ${\bf H}_{[2]}$ satisfy the conditions in Section \ref{sec: Differ-GNN-precoding}.}\label{fig:Prob-SNR} \vspace{-5mm}
\end{figure}

In Fig. \ref{fig:Prob-SNR}(a) and Fig. \ref{fig:Prob-SNR}(b), we show the probability of $F_{\sf ls}({\bm \alpha}_{[1]})=F_{\sf ls}({\bm \alpha}_{[2]})$ and $F_{\sf pc}({\bm \alpha}_{[1]})=F_{\sf pc}({\bm \alpha}_{[2]})$ simulated with different values of $K$ and transmit power, where $p_k=P$ for the link scheduling problem and $P_{\max}=P$ for the power control problem.
${\bm \alpha}_{[1]}$ and ${\bm \alpha}_{[2]}$ are generated by solving the linear equations in \eqref{eq:variable-equation}. Since the computational complexity of solving \eqref{eq:variable-equation} is high when $K$ is large, we take $K\in\{3,4,5,6\}$ as examples. For the link scheduling problem, the optimal solutions are obtained by exhaustive searching. For the power control problem, the sub-optimal solutions are obtained by the WMMSE algorithm. Since the optimized powers are continuous, we regard the frequency of $|F_{\sf pc}({\bm \alpha}_{[1]})-F_{\sf pc}({\bm \alpha}_{[2]})|_{max}<10^{-3}$ as the probability.
We can see that
the probability decreases with $K$ and $P$.
This can be explained as follows. Since the solution spaces of both problems become large with a large value of $K$, the probability that the optimal solutions for ${\bm \alpha}_{[1]}$ and ${\bm \alpha}_{[2]}$ are identical decreases with $K$. When $P$ is low such that the noise power dominates, the optimization for the $K$ D2D pairs in problem \eqref{eq: max sum-rate} and \eqref{eq: max sum-ratepc} is decoupled. In this case, the optimal scheduling policy is to activate all the links, and the optimal power control policy is to transmit with the maximal power to all the receivers. Hence, given any two channel gain matrices, the two optimal solutions are always identical.

In Fig. \ref{fig:Prob-SNR}(c), we show the probability of $F_{\sf p}({\bf H}_{[1]})=F_{\sf p}({\bf H}_{[2]})$ simulated with $N=2$ or $4$ and $K=2$, where  ${\bf H}_{[1]}$ and ${\bf H}_{[2]}$ are generated
by solving the linear equations in \eqref{eq:variable-equation-precoding}. The sub-optimal solutions are also obtained by the WMMSE algorithm. Since the precoding variables are continuous, we regard the frequency of $|F_{\sf p}({\bf H}_{[1]})-F_{\sf p}({\bf H}_{[2]})|_{max}<\epsilon$ as the probability, where $\epsilon$ is a parameter. We can see that the probability is low under different values of $\epsilon$ and SNR. When $\epsilon$ is smaller than 0.05, or the values of $N$ and $K$ are larger, the probability is almost zero, i.e., the optimized precoding matrices for ${\bf H}_{[1]}$ and ${\bf H}_{[2]}$ are different with very high probability.
\vspace{-5mm}

\begin{figure}[!htb]
	\centering
	\begin{minipage}[t]{0.318\linewidth}
		\subfigure[Link Scheduling]{
			\includegraphics[width=\textwidth]{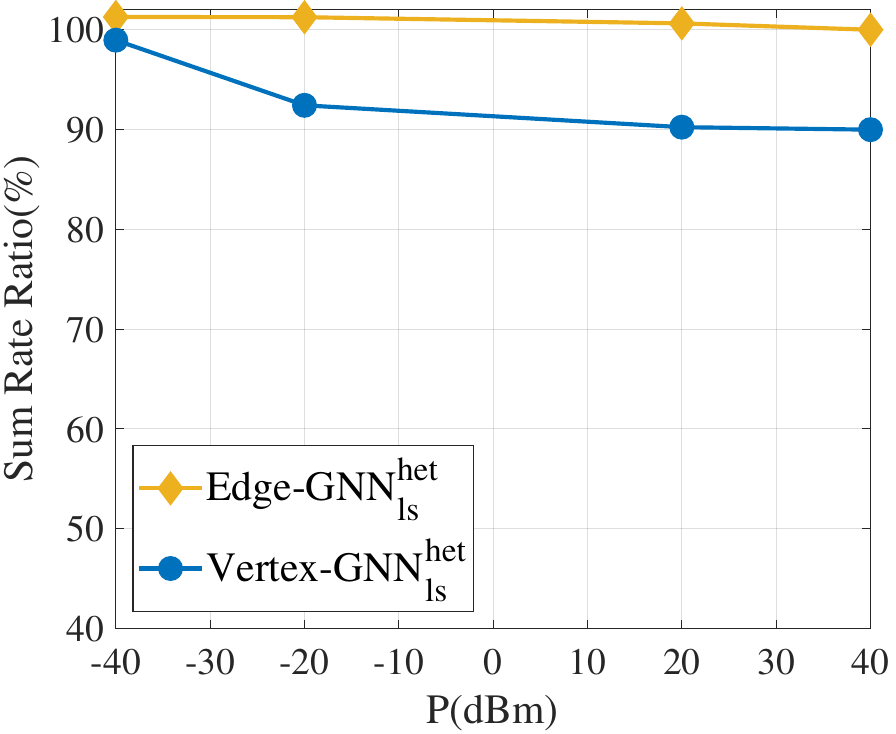}}
	\end{minipage}
	\begin{minipage}[t]{0.318\linewidth}	
		\subfigure[Power Control]{
			\includegraphics[width=\textwidth]{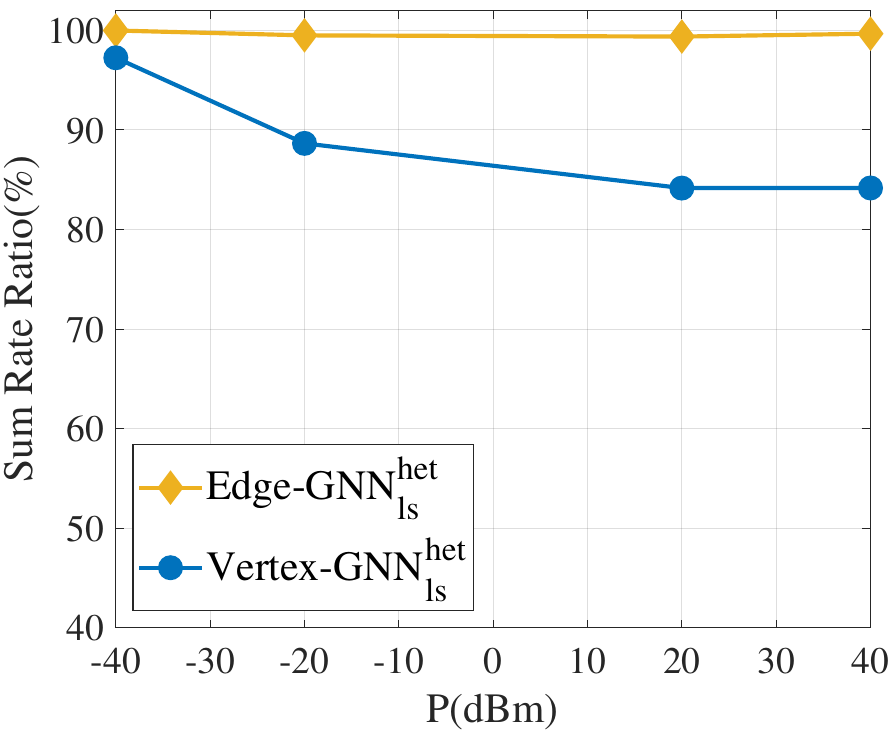}}
	\end{minipage}
    \begin{minipage}[t]{0.32\linewidth}	
	    \subfigure[Precoding]{
		    \includegraphics[width=\textwidth]{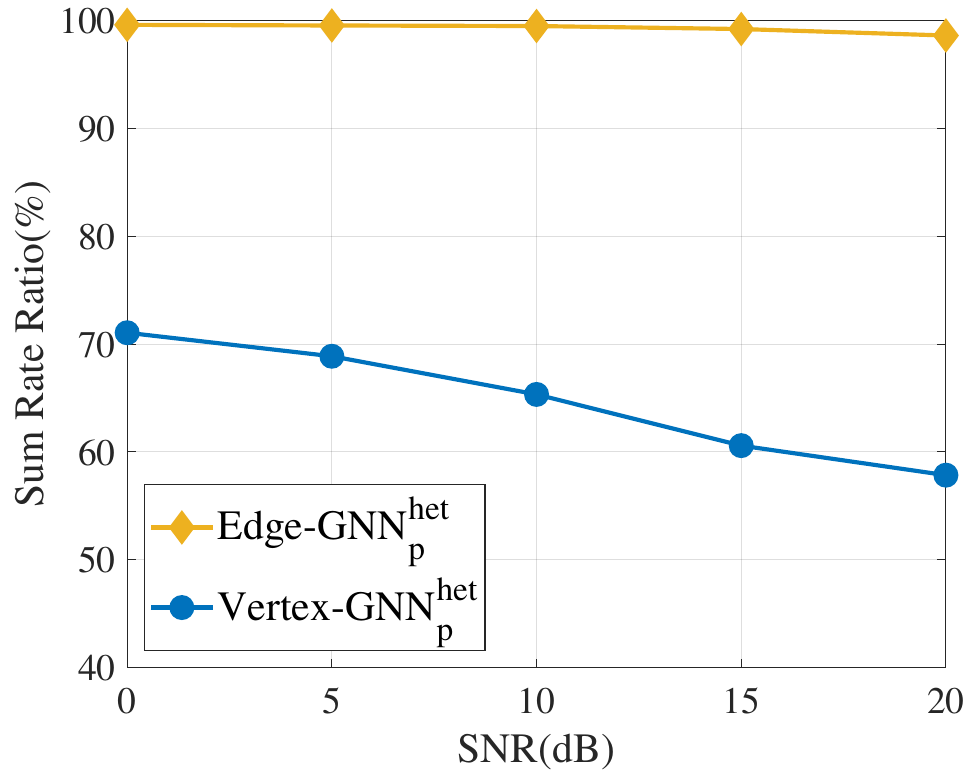}}
    \end{minipage}
	\vspace{-2mm}
	\caption{System performance of the vanilla GNNs versus $P$ or SNR. (a) and (b): $K=50$. (c): $N=4$, $K=2$.}\label{fig:GNN-perf-SNR}
	%\vspace{-4mm}
\end{figure}

%\vspace{-3mm}
In Fig. \ref{fig:GNN-perf-SNR}(a) and Fig. \ref{fig:GNN-perf-SNR}(b), we show the performance of the link scheduling and power control policies learned by the vanilla GNNs versus $P$. The GNNs are trained with 1000 samples, and the fine-tuned hyper-parameters are shown in Table \ref{table: hyper-param-vanilla}.
We can see that the performance of the vanilla Vertex-GNN$_{\sf ls}^{\sf het}$ degrades rapidly with $P$, while the performance of the vanilla Edge-GNN$_{\sf ls}^{\sf het}$ changes little with $P$. This is because when the value of $P$ is higher, the probability of $F_{\sf ls}({\bm \alpha}_{[1]})=F_{\sf ls}({\bm \alpha}_{[2]})$ or $F_{\sf pc}({\bm \alpha}_{[1]})=F_{\sf pc}({\bm \alpha}_{[2]})$ is lower according to Fig. \ref{fig:Prob-SNR}, but the vanilla Vertex-GNN$_{\sf ls}^{\sf het}$ yields the same output with the two channel matrices ${\bm \alpha}_{[1]}$ and ${\bm \alpha}_{[2]}$.
%In other words, the vanilla Vertex-GNN$_{\sf ls}^{\sf het}$ does not have sufficient expressive power to learn the two policies.
Recall that the vanilla Vertex-GNN$_{\sf ls}^{\sf het}$ is with the mean-pooling function. If max-pooling is used as in \cite{2021Shen}, the Vertex-GNN will perform much better when learning the link scheduling or power control policy, which however is still inferior to the vanilla Edge-GNN$_{\sf ls}^{\sf het}$. It is noteworthy that the sum rate ratio achieved by the link scheduling policy learned by the vanilla Edge-GNN exceeds 100\% as shown in Fig. \ref{fig:GNN-perf-SNR}(a). This is because the sum rate ratio
is the ratio of the sum rate achieved by the learned policy to the sum rate achieved by the FPLinQ algorithm that can only learn suboptimal solutions.

In Fig. \ref{fig:GNN-perf-SNR}(c), we show the performance of the precoding policies learned by the vanilla GNNs versus SNR. The GNNs are trained with 10$^5$ samples, and the fine-tuned hyper-parameters are shown in Table \ref{table: hyper-param-vanilla}. We can see that the vanilla Vertex-GNN$_{\sf p}^{\sf het}$ performs poor under different SNRs. This is because the probability of $F_{\sf p}({\bf H}_{[1]})=F_{\sf p}({\bf H}_{[2]})$ is low under different SNRs as shown in Fig. \ref{fig:Prob-SNR}(c), but the vanilla Vertex-GNN$_{\sf p}^{\sf het}$ yields the same output for ${\bf H}_{[1]}$ and ${\bf H}_{[2]}$. As expected, the vanilla Edge-GNN$_{\sf p}^{\sf het}$ performs very well.
%Hence, the expressive capabilities of the vanilla Vertex-GNNs degrade with $P$.
\begin{table}[htb!]
	\centering
	\vspace{-6mm}
	\caption{Fine-tuned hyper-parameters for the vanilla GNNs}\label{table: hyper-param-vanilla}
	\vspace{-2mm}
	\footnotesize
	\renewcommand\arraystretch{0.95}
	\begin{threeparttable}
	\begin{tabular}{c|c|c|c}
		\hline\hline
		\multirow{2}{*}{\textbf{Tasks}}&\multirow{2}{*}{\textbf{Parameters}} & \multicolumn{2}{c}{\textbf{Values}}\\
		\cline{3-4}
		~&~&\tabincell{c}{Vertex-GNN}&\tabincell{c}{Edge-GNN}\\
		\hline
		\multirow{2}{*}{\textbf{Link scheduling}}&\tabincell{c}{Dimension of representation  vectors in each layer} &8,8,8,8,8,1&8,8,8,8,8,1\\
      \cline{2-4}
		~&\tabincell{c}{Learning rate} & 0.001&0.01\\
		\hline
		\multirow{2}{*}{\textbf{Power control}}&\tabincell{c}{Dimension of representation vectors in each layer}  &8,8,8,8,8,1&16,16,16,16,16,1\\
		\cline{2-4}
		~&\tabincell{c}{Learning rate} &  0.001&0.01\\
		\hline
		\multirow{3}{*}{\textbf{Precoding}}&\tabincell{c}{Dimension of representation vectors in each layer}  &32,32,32,32,32&32,32,32,32,32,2\\
		\cline{2-4}
		~&\tabincell{c}{Number of neurons in the read-out layer} &  256,2&-\tnote{1}\\
		\cline{2-4}
		~&\tabincell{c}{Learning rate} &  0.001&0.01\\
		\hline\hline
	\end{tabular}
\begin{tablenotes}
	\footnotesize
	\item[1] In the vanilla Edge-GNN$_{\sf p}^{\sf het}$, there is no read-out layer.
\end{tablenotes}\vspace{-7mm}
    \end{threeparttable}
%\vspace{-10mm}
\end{table}

\vspace{-4mm}
\subsection{Impact of the Linearity of Processing and Combination Functions}
To validate the analysis in Section \ref{sec:Influence-q-CB},
%we compare the performance of Vertex-GNN$_{\sf ls}^{\sf het}$ and Edge-GNN$_{\sf ls}^{\sf het}$ with the vanilla Vertex-GNN$_{\sf ls}^{\sf het}$ and vanilla Edge-GNN$_{\sf ls}^{\sf het}$.
we compare the performance of Vertex-GNNs and Edge-GNNs with vanilla Vertex-GNNs and vanilla Edge-GNNs.

 The hyper-parameters for the vanilla GNNs are the same as Table \ref{table: hyper-param-vanilla}. The fine-tuned hyper-parameters of the Vertex-GNNs with FNN as processor or combiner are shown in Table \ref{table: hyper params-Vertex-NN}. The GNNs for learning the link scheduling policy and the power control policy are trained with 1000 samples. The GNNs for learning the precoding policy with $N=4$, $K=2$, and ${\sf SNR}=10~ {\sf dB}$ are trained with 10$^5$ samples. In Table \ref{table: Performance-q-CB}, we provide the simulation results, where the performance of the vanilla GNNs is marked with bold font.

\begin{table}[htb!]
	\centering
	\vspace{-6mm}
	\caption{Fine-tuned hyper-parameters for the Vertex-GNNs with FNN as processor or combiner}\label{table: hyper params-Vertex-NN}
	\vspace{-2mm}
	\footnotesize
	\renewcommand\arraystretch{0.95}
		\begin{tabular}{c|c|c}
			\hline\hline
			{\textbf{Tasks}}&{\textbf{Parameters}} & {\textbf{Values}}\\
			\hline
			\multirow{3}{*}{\textbf{Link scheduling}}&\tabincell{c}{Dimension of representation  vectors in each layer} &8,8,8,8,8,1\\
			\cline{2-3}
			~&\tabincell{c}{Number of neurons in each hidden layer of FNN} & 32 \\
			\cline{2-3}
			~&\tabincell{c}{Learning rate} & 0.01\\
			\hline
			\multirow{3}{*}{\textbf{Power control}}&\tabincell{c}{Dimension of representation vectors in each layer}  &16,16,16,16,16,1\\
			\cline{2-3}
			~&\tabincell{c}{Number of neurons in each hidden layer of FNN} & 32\\
			\cline{2-3}
			~&\tabincell{c}{Learning rate} &  0.001\\
			\hline
			\multirow{4}{*}{\textbf{Precoding}}&\tabincell{c}{Dimension of representation vectors in each layer}  &16,16,16\\
			\cline{2-3}
			~&\tabincell{c}{Number of neurons in each hidden layer of FNN} & 256\\
			\cline{2-3}
			~&\tabincell{c}{Number of neurons in the read-out layer} &  256,2\\
			\cline{2-3}
			~&\tabincell{c}{Learning rate} &  0.001\\
			\hline\hline
		\end{tabular}
	%\vspace{-10mm}
\end{table}

\vspace{-4mm}
	\begin{table*}[!htb]
        \centering
	\vspace{-4mm}
	\caption{Performance of GNNs with different processing and combination functions}\label{table: Performance-q-CB}
	\vspace{-2mm}
	\footnotesize
	\renewcommand\arraystretch{0.95}
	\begin{threeparttable}
	\begin{tabular}{c|c|c|c|c|c|c|c|c}
	\hline\hline
	\multicolumn{6}{c|}{\textbf {GNNs}} & \multicolumn{2}{c}{\textbf {Performance}}\\
      \hline
        \multirow{2}{*}{}&\multicolumn{2}{c|}{Processing}&\multicolumn{3}{c|}{Combination}&\multirow{2}{*}{Link Scheduling}&\multirow{2}{*}{Power Control}&\multirow{2}{*}{Precoding}\\
        \cline{2-6}
      &Linear&FNN&Linear&$\sigma(\cdot)\tnote{1}$&FNN& & \\
      \hline
%      \multirow{3}{*}{Vertex-DirGNN-LS}&\checkmark&\ding{55}&\ding{55}& \checkmark&\ding{55}&\textbf{89.96\%}&\textbf{84.13\%}\\
%      \cline{2-8}
%      &\checkmark&\ding{55}&\ding{55}&\ding{55}&\checkmark&$89.90\%(-0.07\%)$&$84.43\%(+0.36\%)$\\
%      \cline{2-8}
%      &\ding{55}&\checkmark&\ding{55}&\checkmark&\ding{55}&99.99\%(+11.15\%)&99.75\%(+18.57\%)\\
%      \hline
      \multirow{3}{*}{Vertex-GNNs}&\checkmark&\ding{55}&\ding{55}& \checkmark&\ding{55}&\textbf{89.98\%}&\textbf{84.17\%}&\textbf{64.94\%}\\
      \cline{2-9}
      &\checkmark&\ding{55}&\ding{55}&\ding{55}&\checkmark&90.06\%(+0.09\%)&84.44\%(+0.32\%)&65.12\%(+0.28\%)\\
      \cline{2-9}
      &\ding{55}&\checkmark&\ding{55}&\checkmark&\ding{55}&99.83\%(+10.95\%)&99.65\%(+18.45\%)&99.08\%(+52.57\%)\\
      \hline
%      \multirow{2}{*}{Edge-DirGNN-LS}&\checkmark&\ding{55}&\ding{55}& \checkmark&\ding{55}&\textbf{100.41}\%&\textbf{99.87\%}\\
%      \cline{2-8}
%      &\checkmark&\ding{55}&\checkmark&\ding{55}&\ding{55}&89.60\%(-10.77\%)&82.73\%(-17.16\%)\\
%       %\cline{2-7}
%      %&\ding{55}&\checkmark&\checkmark&\ding{55}&100.42\%(+0.19\%)&99.96\%(+1.23\%)\\
%      \hline
      \multirow{4}{*}{Edge-GNNs}&\checkmark&\ding{55}&\ding{55}& \checkmark&\ding{55}&\textbf{99.99\%}&\textbf{99.67\%}&\textbf{99.16\%}\\
      \cline{2-9}
      &\checkmark&\ding{55}&$CB_I$&$CB_S$&\ding{55}&89.91\%(-10.08\%)&84.13\%(-15.59\%)&-\\
       \cline{2-9}
      &\checkmark&\ding{55}&$CB_S$&$CB_I$&\ding{55}&99.99\%(+0.00\%)&99.22\%(-0.45\%)&-\\
       \cline{2-9}
      &\checkmark&\ding{55}&\checkmark&\ding{55}&\ding{55}&-&-&62.23\%(-37.24\%)\\
      \hline\hline
	\end{tabular}
     \begin{tablenotes}
      \footnotesize
      \item[1]$\sigma(\cdot)$ means that the combination function is a linear function cascaded with an activation function.
      %\item[2] "S" represents $CB_S(\cdot)$, and "I" represents $CB_I(\cdot)$.
     \end{tablenotes}
	\vspace{-6mm}
   \end{threeparttable}
\end{table*}

It is shown that for the Vertex-GNNs learning the same policy, when the processing function is FNN (no matter if the combiner is linear or non-linear, where the results for linear combiner are not shown because combiner is usually non-linear), the performance is much better than the vanilla Vertex-GNN.
%, because the Vertex-GNN$_{\sf ls}^{\sf het}$ can differentiate ${\bm \alpha}_{[1]}$ and ${\bm \alpha}_{[2]}$ but the vanilla Vertex-GNN$_{\sf ls}^{\sf het}$ cannot.
When only the combination function is FNN, the Vertex-GNN performs closely to the vanilla Vertex-GNN, because both of them cannot differentiate ${\bm \alpha}_{[1]}$ and ${\bm \alpha}_{[2]}$ or ${\bf H}_{[1]}$ and ${\bf H}_{[2]}$.
%The Vertex-GNNs with non-linear processing and combination functions perform close to the Vertex-GNNs only with FNN-processor.{\bf ? right? if yes, why remove?}

For the Edge-GNNs learning the link scheduling and power control policies, when ${\sf CB}_S(\cdot)$ is non-linear and ${\sf CB}_I(\cdot)$ is linear, Edge-GNN$_{\sf ls}^{\sf het}$ is inferior to the vanilla Edge-GNN$_{\sf ls}^{\sf het}$.
%This is because this Edge-GNN$_{\sf ls}^{\sf het}$ cannot differentiate ${\bm \alpha}_{[1]}$ and ${\bm \alpha}_{[2]}$.
%Hence this Edge-HetGNN has weaker expressive capability than the vanilla Edge-HetGNN.
When ${\sf CB}_I(\cdot)$ is non-linear and ${\sf CB}_S(\cdot)$ is linear, Edge-GNN$_{\sf ls}^{\sf het}$ performs closely to the vanilla Edge-GNN$_{\sf ls}^{\sf het}$, since both of them can differentiate ${\bm \alpha}_{[1]}$ and ${\bm \alpha}_{[2]}$.
When learning the precoding policy, the Edge-GNN with  linear combination function is inferior to the vanilla Edge-GNN,
since it cannot differentiate ${\bf H}_{[1]}$ and ${\bf H}_{[2]}$.

%These results are consistent with the analysis in Section \ref{sec:Influence-q-CB}.

Next, we show the impact of different activation functions. For the Vertex-GNNs, when the processing function is a non-linear function such as FNN, they can perform well even if a linear combination function is applied. Hence,  non-linear activation functions in the combiner have little impact on the performance of the Vertex-GNNs.
For the vanilla GNNs, only the combination functions contain non-linear activation functions. Thereby, we only compare the performance of the vanilla GNNs with several non-linear activation functions. The simulation results are provided in Table \ref{table: different-activation-function}. It shows that the vanilla GNNs with different activation functions achieve similar performance.

\vspace{-4mm}
\begin{table}[H]
	\centering %\vspace{-6mm}
	\footnotesize
	\caption{Performance of vanilla GNNs under different non-linear activation functions}\label{table: different-activation-function}
	\renewcommand\arraystretch{0.95}\vspace{-2mm}
	\begin{tabular}{c|c|c|c|c|c|c|c}
		\hline\hline
		\multirow{3}{*}{\textbf{Tasks}} &\multirow{3}{*}{\textbf{GNNs}} & \multicolumn{6}{c}{\textbf{Activation functions}}\\
		\cline{3-8}
      &&relu&leaky\_relu&elu&swish&softplus&mish\\
     \cline{3-8}
     &&$\tiny \text{$\max(0,x)$}$&$\tiny \begin{cases}0.2x& \text{$x<0$}\\x& \text{$x\geq0$}
\end{cases}$&$\tiny \begin{cases} e^x-1& \text{$x<0$}\\x& \text{$x\leq 0$}
\end{cases}$&$\tiny \text{$\frac{x}{1+e^{-x}}$}$&$\tiny\text{$ln(1+e^x)$}$&$\tiny \text{$x\cdot tanh(ln(1+e^x))$}$\\
     \hline
		\multirow{2}{*}{\textbf{\scriptsize\makecell{Link \\Scheduling}}}&\tabincell{c}{\scriptsize Vertex-GNN$_{\sf ls}^{\sf het}$} & 89.98\% &89.95\% & 90.08\%&90.05\%&90.21\%&90.10\%\\
		\cline{2-8}
		&\tabincell{c}{\scriptsize Edge-GNN$_{\sf ls}^{\sf het}$} &99.99\% & 99.99\%&99.98\%&99.83\%&99.66\%&100.18\%\\
		\hline
		\multirow{2}{*}{\textbf{\scriptsize\makecell{Power\\ Control}}}&\tabincell{c}{\scriptsize Vertex-GNN$_{\sf ls}^{\sf het}$} & 84.18\% &84.18\% & 84.36\%&84.41\%&84.30\%&84.34\%\\
		\cline{2-8}
		&\tabincell{c}{\scriptsize Edge-GNN$_{\sf ls}^{\sf het}$} &99.67\% & 98.39\%&99.59\%&99.49\%&99.50\%&99.61\%\\
		\hline
		\multirow{2}{*}{\textbf{\scriptsize Precoding}}&\tabincell{c}{\scriptsize Vertex-GNN$_{\sf p}^{\sf het}$} &65.17\% & 65.12\% &65.14\%&65.05\%&65.17\%&65.09\%\\
		\cline{2-8}
		~&\tabincell{c}{\scriptsize Edge-GNN$_{\sf p}^{\sf het}$} &99.16\% & 99.14\% &99.43\%&99.49\%&99.46\%&99.58\%\\
		\hline\hline
	\end{tabular}\vspace{-6mm}
	%\begin{flushleft}
	%	*: ``LS'' and ``PC'' stand for link scheduling and power control, respectively.
	%\end{flushleft}
\end{table}

\vspace{-4mm}
\subsection{Impact of the Output Dimensions of Processing and Combination Functions} \label{sec:result-dimension}\vspace{-2mm}
To validate the analysis in Section \ref{sec:dimension}. we provide the performance of the vanilla Edge-GNN$_{\sf ls}^{\sf het}$, the vanilla Edge-GNN$_{\sf p}^{\sf het}$, as well as Vertex-GNN$_{\sf ls}^{\sf het}$ and Vertex-GNN$_{\sf p}^{\sf het}$ with FNN as processor (denoted as {Vertex-GNN$_{\sf ls}^{\sf het}$+FNN} and {Vertex-GNN$_{\sf p}^{\sf het}$+FNN}, respectively).\footnote{The features and actions of {Vertex-GNN$_{\sf ls}^{\sf het}$+FNN} and {Vertex-GNN$_{\sf p}^{\sf het}$+FNN} are respectively the same as those of Vertex-GNN$_{\sf ls}^{\sf het}$ and Vertex-GNN$_{\sf p}^{\sf het}$ in Table. \ref{table: define-feature-action}.}
%For better dimension influence...

The hyper-parameters of the GNNs for learning the link scheduling and power control policies
are provided in Tables \ref{table: hyper-param-vanilla} and \ref{table: hyper params-Vertex-NN}. The hyper-parameters of the GNNs for learning the precoding policy with $N=4$ and $K=2$ are also shown in Tables \ref{table: hyper-param-vanilla} and \ref{table: hyper params-Vertex-NN}.
When using the GNNs to learn the precoding policy with $N=8$ and $K=4$, the number of training samples needs to increase to achieve acceptable performance, which is set to be $5\times 10^5$. Their hyper-parameters are as follows. For the Vertex-GNN$_{\sf p}^{\sf het}$+FNN, $L=5$, the processing function is a FNN with a hidden layer with 256 neurons, the read-out layer is a FNN with five hidden layers each with 512 neurons. For the vanilla Edge-GNN$_{\sf p}^{\sf het}$, $L=9$.

\iffalse
\vspace{-4mm}
\begin{table*}[!htb]
	\centering
	\caption{Hyper-parameters of GNNs for learning the precoding policy, SNR = 10 dB}\label{table: precoding-hyper}
\vspace{-2mm}
	\footnotesize
	\renewcommand\arraystretch{0.95}
\begin{threeparttable}
	\begin{tabular}{c|c|c|c|c|c|c}
	\hline\hline
	\multicolumn{2}{c|}{\multirow{3}{*}{\textbf{GNNs}}}&\multirow{3}{*}{\textbf{Training samples}}&\multicolumn{4}{c}{\textbf{Hyper-parameters}}\\
	\cline{4-7}
	\multicolumn{2}{c|}{}&&\multirow{2}{*}{\textbf{Learning rate}}&\multirow{2}{*}{$L$}&\multicolumn{2}{c}{\textbf{FNN}}\\
	\cline{6-7}
	\multicolumn{2}{c|}{}&&&&\textbf{processing function}&\textbf{post-processing layer}\\
	\hline
	\multirow{2}{*}{\makecell[c]{\textbf{N=4}\\\textbf{K=2}}}&{Vertex-GNN$_{\sf p}^{\sf het}$+FNN}&\multirow{2}{*}{$10^5$}&0.001&3&[256]&[256]\\
	\cline{2-2}\cline{4-7}
	&{Vanilla Edge-GNN$_{\sf p}^{\sf het}$}&&0.01&6&-\tnote{1}&-\tnote{1}\\
	\hline
	\multirow{2}{*}{\makecell[c]{\textbf{N=8}\\\textbf{K=4}}}&{Vertex-GNN$_{\sf p}^{\sf het}$+FNN}&\multirow{2}{*}{$5\times10^5$}&0.001&5&[256]&[512,512,512,512,512]\\
	\cline{2-2}\cline{4-7}
	&{Vanilla Edge-GNN$_{\sf p}^{\sf het}$}&&0.01&9&-\tnote{1}&-\tnote{1}\\
	\hline\hline
    \end{tabular}
\begin{tablenotes}
	\footnotesize
	\item[1] In the vanilla Edge-GNN$_{\sf p}^{\sf het}$, the processing functions are linear, and there is no read-out layer.
\end{tablenotes}\vspace{-7mm}
\end{threeparttable}
\end{table*}
\fi

\begin{figure}[!htb]
	\centering
	\begin{minipage}[t]{0.4\linewidth}
		\subfigure[Link Scheduling]{
			\includegraphics[width=\textwidth]{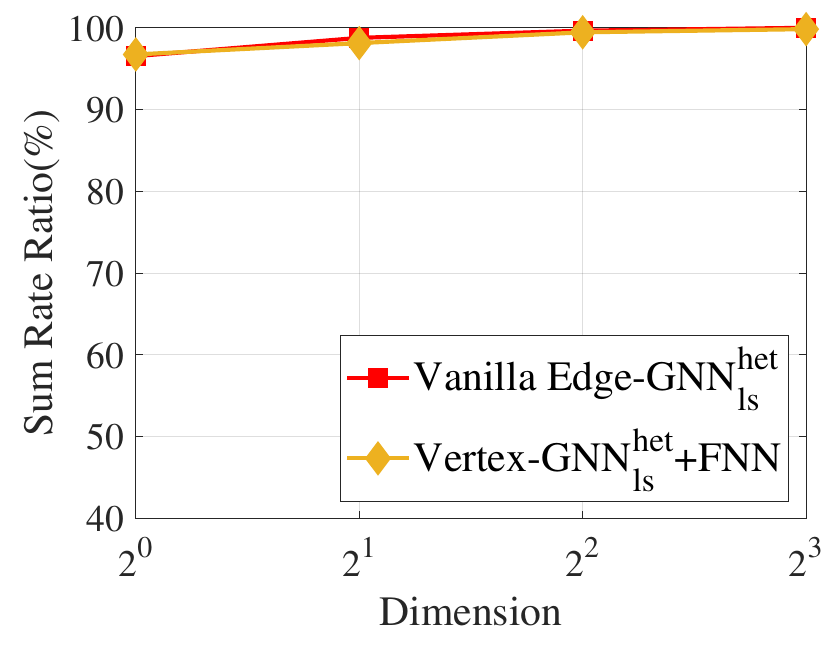}}
	\end{minipage}
	\begin{minipage}[t]{0.4\linewidth}	
		\subfigure[Power Control]{
			\includegraphics[width=\textwidth]{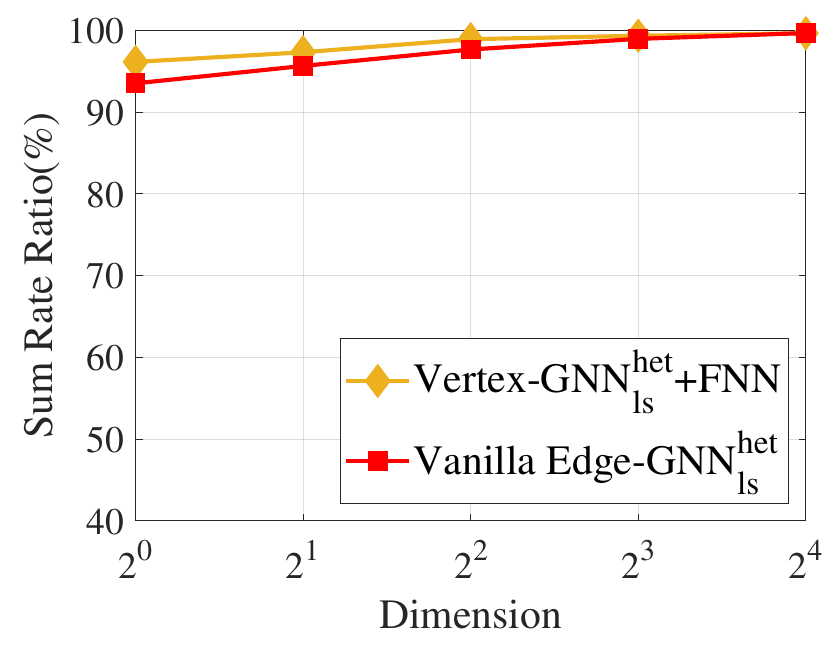}}
	\end{minipage}\vspace{-2mm}
	\caption{Performance of the GNNs with different dimensions of hidden representations, $K=50$.}\label{fig:Dimension_performance_LS_P}
\end{figure}

In Fig. \ref{fig:Dimension_performance_LS_P}, we show the performance of the GNNs with different values of ``Dimension'' for learning the link scheduling and power control policies, where the ``Dimension'' indicates the value of $M^{(l)},~\forall l$. It is shown that the performance of the GNNs grows slowly with ``Dimension''. This is because the dimensions of the representation vectors do not affect the expressive power of the GNNs for learning these two policies, but the representation vectors with higher dimension lead to a wider GNN and hence provide larger hypothesis space.
\vspace{-4mm}

\begin{figure}[!htb]
	\centering
	\begin{minipage}[t]{0.395\linewidth}
		\subfigure[$N=4$, $K=2$]{
			\includegraphics[width=\textwidth]{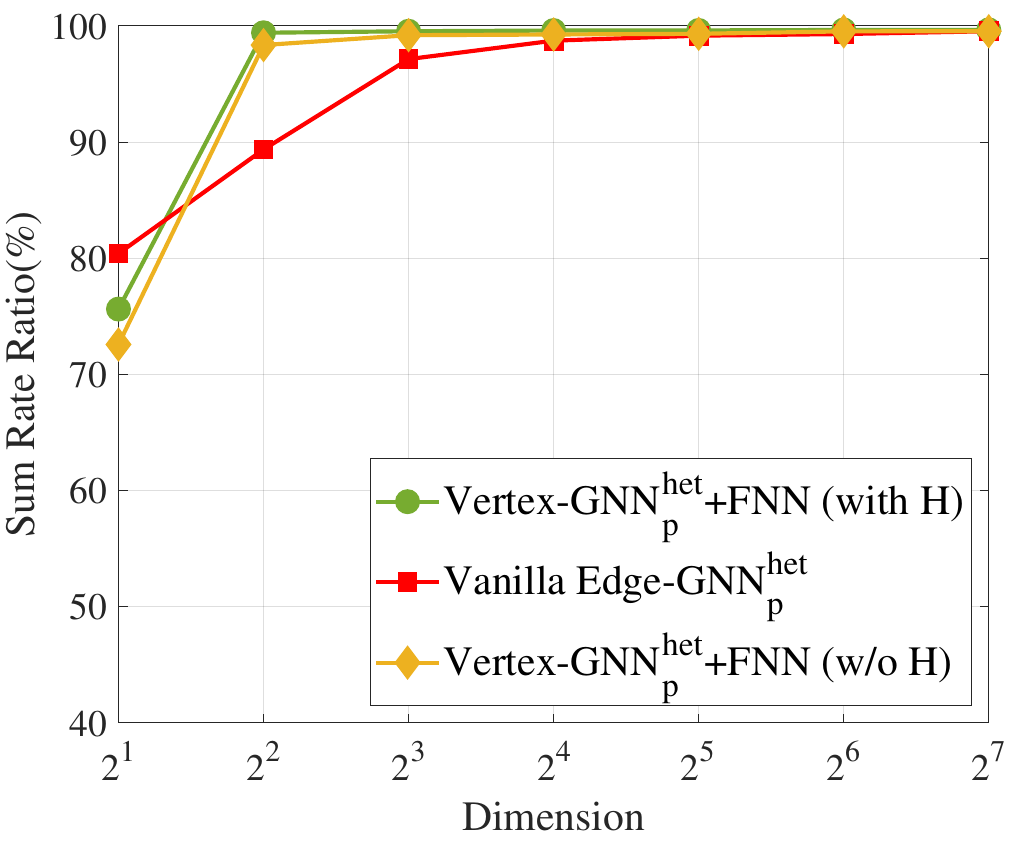}}
	\end{minipage}
	\begin{minipage}[t]{0.405\linewidth}	
		\subfigure[$N=8$, $K=4$]{
			\includegraphics[width=\textwidth]{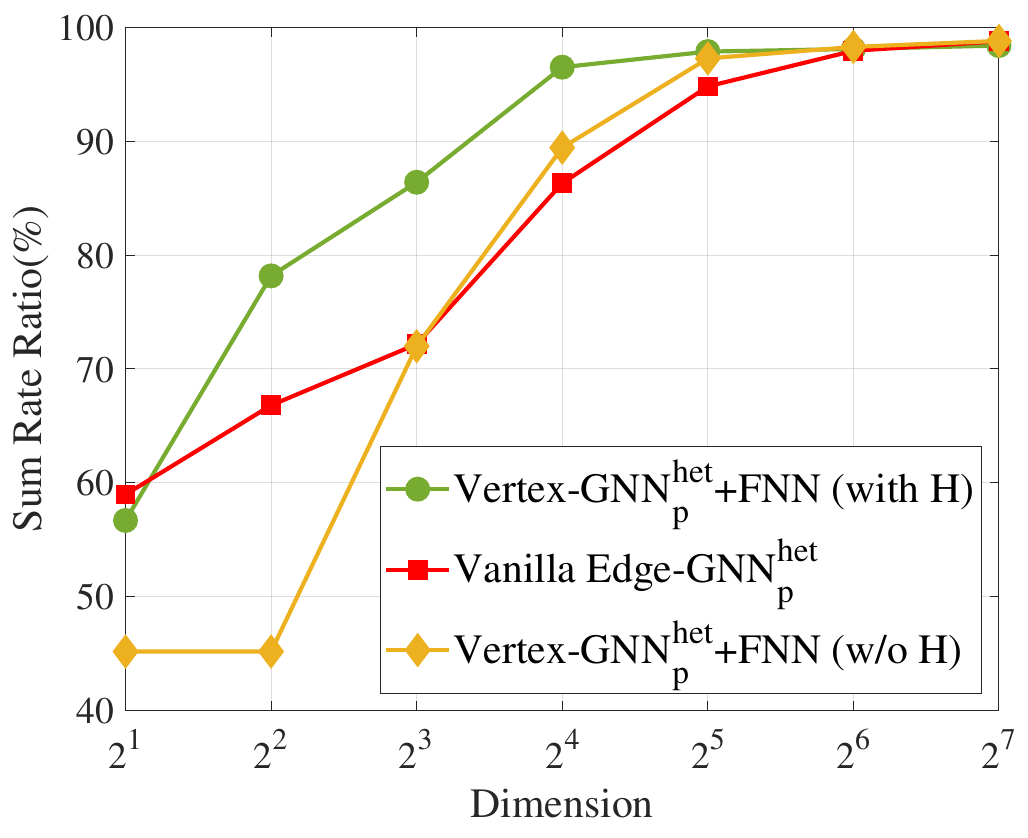}}
	\end{minipage}\vspace{-2mm}
	\caption{Performance of the GNNs with different dimensions of hidden representations, precoding, ${\sf SNR}=10~ {\sf dB}$.}\label{fig:Dimension_performance}
	\vspace{-7mm}
\end{figure}

\vspace{-1mm}
In Fig. \ref{fig:Dimension_performance}, we show the performance of the GNNs with different values of ``Dimension'' for learning the precoding policy where the ``Dimension'' indicates the value of $M_A^{(l)}=M_U^{(l)}=M_{A,q}^{(l)}=M_{U,q}^{(l)}=M_E^{(l)}=M_{E,q}^{(l)},~\forall l$. We show the impact of the output dimensions of the processors and combiners in this simple way, because there are too many combinations of these dimensions that can satisfy \eqref{eq:equation-2} or \eqref{eq:equation-edge}. {Vertex-GNN$_{\sf p}^{\sf het}$+FNN (w/o H)} is a Vertex-GNN$_{\sf p}^{\sf het}$+FNN, where only the representations of vertices are input to the read-out layer (i.e., $\hat v_{ij,{V}}={\sf FNN}_{\sf read}^{\sf het}({\bf d}_{i,A}^{(L)},{\bf d}_{j,U}^{(L)})$ as in Remark \ref{remark: Vertex-GNN-precoding}. {Vertex-GNN$_{\sf p}^{\sf het}$+FNN (with H)} is another Vertex-GNN$_{\sf p}^{\sf het}$+FNN, where the channel matrix $\bf H$ are also input to the read-out layer (i.e., $\hat v_{ij,V}={\sf FNN}_{\sf read}^{\sf het}({\bf d}_{i,A}^{(L)},{\bf d}_{j,U}^{(L)},h_{ij}) $).
It is shown that all the GNNs perform better with larger ``Dimension''. For Edge-GNN, \eqref{eq:equation-edge} is satisfied when ``Dimension''$\geq 2$. For Vertex-GNN, \eqref{eq:equation-2} is satisfied when ``Dimension''$\geq 3$
for $N=4$ and $K=2$ and when ``Dimension''$\geq 6$
for $N=8$ and $K=4$.
However, even when \eqref{eq:equation-edge} or \eqref{eq:equation-2} is satisfied, the vanilla Edge-GNN$_{\sf p}^{\sf het}$ or {Vertex-GNN$_{\sf p}^{\sf het}$+FNN {(w/o H)}} may not perform well.
This is because the conditions in \eqref{eq:equation-edge} or \eqref{eq:equation-2} are only necessary for a GNN to avoid information loss.
%The Vertex-GNN$_{\sf p}^{\sf het}$ with higher \emph{Dimension} can learn more complex mapping, and hence can perform better.
With the same ``Dimension'', the {Vertex-GNN$_{\sf p}^{\sf het}$+FNN {(with H)}} performs better than the {Vertex-GNN$_{\sf p}^{\sf het}$+FNN {(w/o H)}}.
This is because the input of the read-out layer of {Vertex-GNN$_{\sf p}^{\sf het}$+FNN (with H)} contains ${\bf H}$, which is helpful to differentiate channel matrices.

\vspace{-4mm}

\subsection{Space, Time, and Sample Complexities of the GNNs}
Finally, we compare the space, time, and sample complexities of the Vertex-GNNs and the vanilla Edge-GNNs. The space complexity is the number of trainable parameters in a fine-tuned GNN to achieve an expected performance, which is set as 98\% sum rate ratio for the precoding problem with $N=8,K=4$, and 99\% for the other two problems and the precoding problem with $N=4,K=2$. The time complexity includes the training time required to achieve the expected performance and the inference time. The sample complexity is the minimal number of samples required for training a GNN to achieve a given performance.

Since the {Vertex-GNN$_{\sf p}^{\sf het}$+FNN (with H)} outperforms the {Vertex-GNN$_{\sf p}^{\sf het}$+FNN (w/o H)}, we only consider the {Vertex-GNN$_{\sf p}^{\sf het}$+FNN (with H)}.
The Vertex-GNN$_{\sf ls}^{\sf het}+$FNN and the vanilla Edge-GNN$_{\sf ls}^{\sf het}$ are with the hyper-parameters in Table \ref{table: hyper params-Vertex-NN} and Table \ref{table: hyper-param-vanilla}, respectively. The {Vertex-GNN$_{\sf p}^{\sf het}$+FNN (with H)} and {Edge-GNN$_{\sf p}^{\sf het}$} are with the hyper-parameters in Section \ref{sec:result-dimension}. Besides, $M_d=4$ for Vertex-GNN and  $M_d=32$ for Edge-GNN when $N=4$ and $K=2$, while $M_d=64$ for Vertex-GNN and $M_d=128$ for Edge-GNN when $N=8$ and $K=4$.

%\vspace{-10mm}
In Table \ref{table: Complexity-results-NN}, we show the space and time complexities of the GNNs.
%Compared with the Vertex-GNN$_{\sf ls}^{\sf het}$+FNN and the Vertex-GNN$_{\sf p}^{\sf het}$+FNN,
%the training and inference times of the vanilla Edge-GNN$_{\sf ls}^{\sf het}$ and the vanilla Edge-GNN$_{\sf p}^{\sf het}$ are lower.
It is shown that the training time and inference time of the vanilla Edge-GNNs are shorter than the Vertex-GNNs, because using FNN as processor is with higher computational complexity than using linear processor.
%When learning the link scheduling and precoding policies,
The space complexities of the vanilla Edge-GNNs are lower than the Vertex-GNNs.

%\vspace{-3mm}
\begin{table}[H]
	\centering \vspace{-6mm}
	\caption{space and time complexity of GNNs}\label{table: Complexity-results-NN}
	\footnotesize
	\renewcommand\arraystretch{0.95}\vspace{-2mm}
	\begin{tabular}{c|c|c|c|c|c}
		\hline\hline
		\multicolumn{2}{c|}{\textbf{Tasks}} &{\textbf{GNN}} & {\textbf{Space (k)}}&{\textbf{Training Time (min)}}& {\textbf{Inference Time (ms)}}\\
		\hline
		\multicolumn{2}{c|}{\multirow{2}{*}{\textbf{Link Scheduling}}}&\tabincell{c}{\scriptsize Vertex-GNN$_{\sf ls}^{\sf het}$+FNN} & 12.74 &4.43 & 20.58\\
		\cline{3-6}
		\multicolumn{2}{c|}{}&\tabincell{c}{\scriptsize Vanilla Edge-GNN$_{\sf ls}^{\sf het}$} &2.14 & 2.08&5.16\\
		\hline
		\multicolumn{2}{c|}{\multirow{2}{*}{\textbf{Power Control}}}&\tabincell{c}{\scriptsize Vertex-GNN$_{\sf ls}^{\sf het}$+FNN} & 24.71 &25.26 & 19.98\\
		\cline{3-6}
		\multicolumn{2}{c|}{}&\tabincell{c}{\scriptsize Vanilla Edge-GNN$_{\sf ls}^{\sf het}$} &8.37 & 5.62&5.36\\
		\hline
		\multirow{4}{*}{\textbf{Precoding}}&\multirow{2}{*}{$N=4,K=2$}&\tabincell{c}{\scriptsize Vertex-GNN$_{\sf p}^{\sf het}$+FNN (with H)} &149.86 & 12.20 &7.36\\
		\cline{3-6}
		~&~&\tabincell{c}{\scriptsize Vanilla Edge-GNN$_{\sf p}^{\sf het}$} &126.72 & 10.95 &4.04\\
		\cline{2-6}
		~&\multirow{2}{*}{$N=8,K=4$}&\tabincell{c}{\scriptsize Vertex-GNN$_{\sf p}^{\sf het}$+FNN (with H)} &1454.34 & 717.82 &10.87\\
		\cline{3-6}
		~&~&\tabincell{c}{\scriptsize Vanilla Edge-GNN$_{\sf p}^{\sf het}$} &345.60 & 23.13 &4.92\\
		\hline\hline
	\end{tabular}\vspace{-6mm}
	%\begin{flushleft}
	%	*: ``LS'' and ``PC'' stand for link scheduling and power control, respectively.
	%\end{flushleft}
\end{table}

%When learning the power control policy, the training time of the vanilla Edge-HetGNN-LS is the lower than the Vertex-GNN-LSs.
%When learning the power control policy, the training time of the vanilla Edge-HetGNN is the lowest. Since the learning rates of the Vertex-HetGNN+FNN and the vanilla Edge-D-GNN are lower, they have higher training time.

%It can also be seen from the Table \ref{table: Complexity-results-NN} that the space complexity of the Vertex-DirGNN-LS+FNN is lower than the Vertex-HetGNN-LS+FNN, and the space complexity of the vanilla Edge-DirGNN-LS is lower than the vanilla Edge-HetGNN-LS. This can be explained as follows.
%For the Vertex-DirGNN-LS+FNN, each vertex updates its hidden representation with shared aggregation and combination functions. For the Vertex-HetGNN-LS+FNN, two types of vertexes updates their hidden representation with different aggregation and combination functions. Hence, the Vertex-DirGNN-LS+FNN has fewer trainable parameters than the Vertex-HetGNN-LS+FNN. Similarity, the vanilla Edge-DirGNN-LS has fewer trainable parameters than the vanilla Edge-HetGNN-LS.

\begin{figure}[!htb]
	\centering
	\begin{minipage}[t]{0.4\linewidth}
		\subfigure[Link Scheduling]{
			\includegraphics[width=\textwidth]{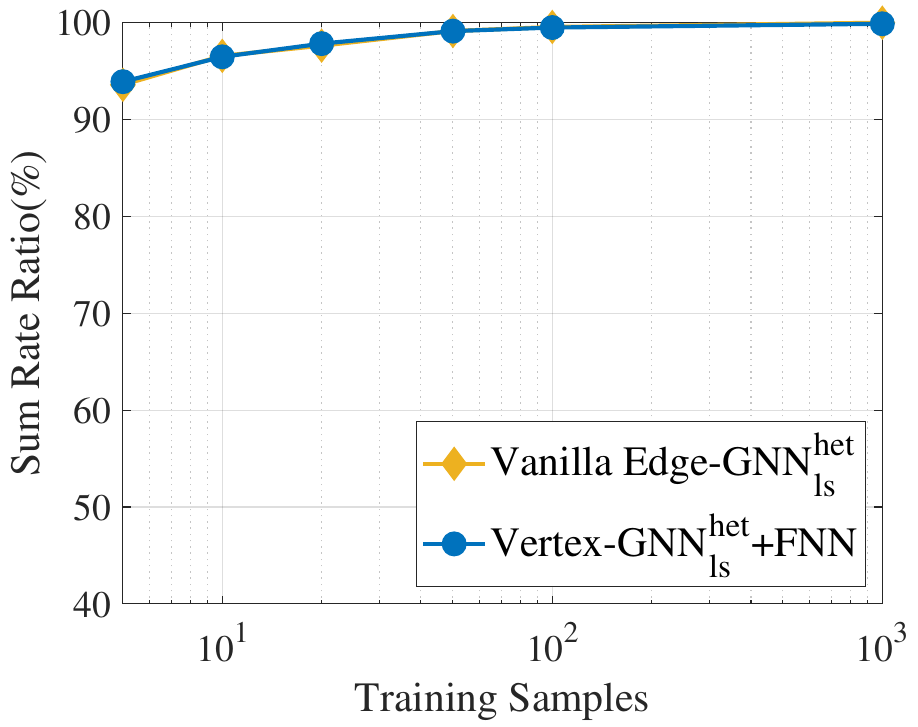}}
	\end{minipage}
	\begin{minipage}[t]{0.4\linewidth}	
		\subfigure[Power Control]{
			\includegraphics[width=\textwidth]{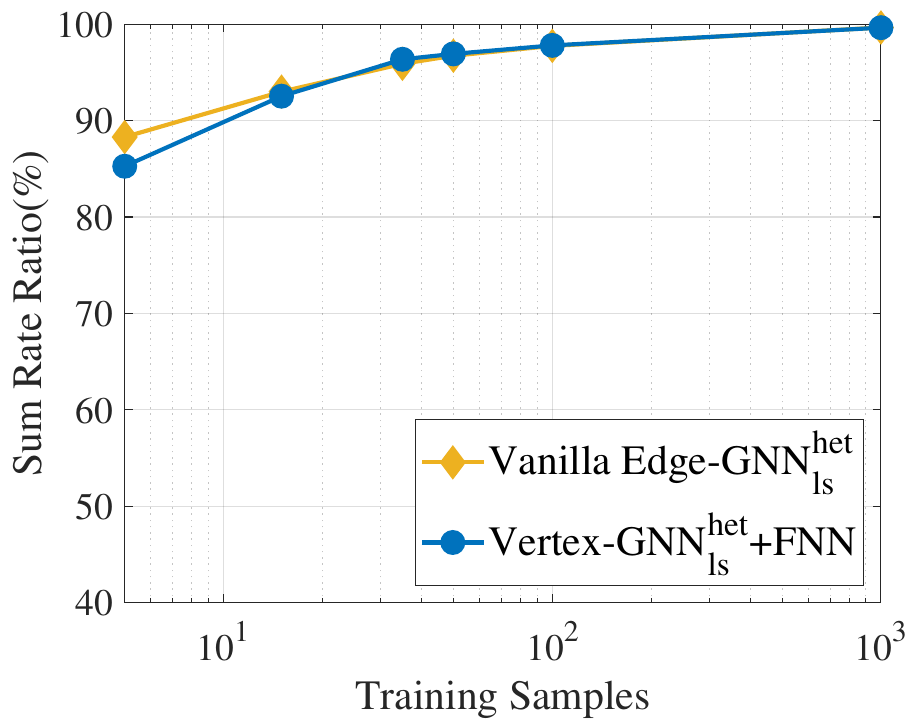}}
	\end{minipage}
	\begin{minipage}[t]{0.42\linewidth}	
		\subfigure[Precoding]{
			\includegraphics[width=\textwidth]{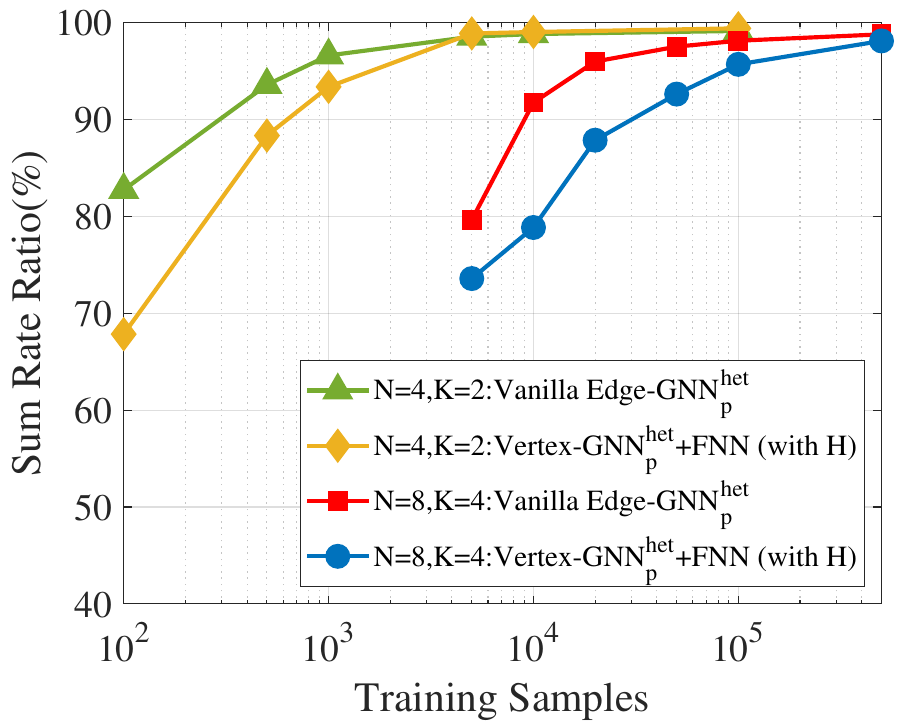}}
	\end{minipage}
	\vspace{-2mm}
	\caption{Performance of the GNNs versus the number of training samples.}\label{fig:Sample_performance}
\end{figure}

In Fig. \ref{fig:Sample_performance}, we compare the performance of the GNNs trained with different numbers of samples.  It is shown that the Edge-GNN and the Vertex-GNN for link scheduling are with almost the same sample complexity, while the Edge-GNN outperforms the Vertex-GNN with few number of training samples when learning the power control policy. When learning the precoding policy, the Edge-GNN (which uses linear processor) is more sample efficient than the Vertex-GNN,
%because the Vertex-GNN is with larger hypothesis space than the Edge-GNN due to using FNN as processor.
because the Vertex-GNN has more trainable parameters than the Edge-GNN due to using FNN as processor and the extra read-out layer.
%\vspace{-2mm}

%\vspace{-4mm}
The main results for the three policies are summarized as follows.
\begin{itemize}
\item The vanilla Edge-GNNs outperform the vanilla Vertex-GNNs and perform closely to the Vertex-GNN with FNN-processor. The performance of the Vertex-GNNs can be improved by using FNN as processor, but cannot be improved by using FNN as combiner.
%\item When learning the precoding policy, if \eqref{eq:equation-2} is satisfied, the performance of \emph{Vertex-HetGNN$_{\sf p}^{\sf het}$-P+FNN without H} can be great. It is better to input $\bf H$ into the read-out layer.
\item The vanilla Edge-GNNs are with lower time complexity than the Vertex-GNN with FNN-processor to achieve the same performance. When learning the precoding policy, the vanilla Edge-GNN is with lower sample complexity than the Vertex-GNN with FNN-processor.
\end{itemize}

	\vspace{-5mm}\section{Conclusion}\label{sec: conclusion}
In this paper, we analyzed the impacts of the linearity and output dimensions of processing and combination functions on the expressive power of the Vertex-GNNs and Edge-GNNs for learning link scheduling, power control, and precoding policies. We demonstrated that all the policies can be learned with either Vertex-GNNs or Edge-GNNs over either homogeneous or heterogeneous graphs. We showed that the Vertex-GNNs with linear processing functions cannot perform well for these policies due to their inability to differentiate all channel matrices. When learning the precoding policy, the expressive power of the Vertex-GNN with non-linear processing functions is still weak, which depends on the output dimensions of processing and combination functions. Simulation results showed that the Edge-GNNs using linear processors can achieve the same performance as the Vertex-GNNs using non-linear processors for learning these policies but with much lower training time and inference time. Our results indicate the advantage of Edge-GNNs for learning wireless policies and provide guidelines for designing efficient and well-performed GNNs. While we focused on two resource allocation policies and a precoding policy, the conclusions are also applicable to other wireless policies such as signal detection, channel estimation, other resource allocation, and other precoding problems whenever the edges of constructed graphs are with features. If both the features and actions of a graph are defined on vertices, then a Vertex-GNN will
be with the same expressive power as an Edge-GNN and may be more sample efficient.

\bibliography{IEEEabrv,ref}

\begin{thebibliography}{10}
\providecommand{\url}[1]{#1}
\csname url@samestyle\endcsname
\providecommand{\newblock}{\relax}
\providecommand{\bibinfo}[2]{#2}
\providecommand{\BIBentrySTDinterwordspacing}{\spaceskip=0pt\relax}
\providecommand{\BIBentryALTinterwordstretchfactor}{4}
\providecommand{\BIBentryALTinterwordspacing}{\spaceskip=\fontdimen2\font plus
\BIBentryALTinterwordstretchfactor\fontdimen3\font minus
  \fontdimen4\font\relax}
\providecommand{\BIBforeignlanguage}[2]{{%
\expandafter\ifx\csname l@#1\endcsname\relax
\typeout{** WARNING: IEEEtran.bst: No hyphenation pattern has been}%
\typeout{** loaded for the language `#1'. Using the pattern for}%
\typeout{** the default language instead.}%
\else
\language=\csname l@#1\endcsname
\fi
#2}}
\providecommand{\BIBdecl}{\relax}
\BIBdecl

\bibitem{WMMSE}
Q.~Shi, M.~Razaviyayn, Z.~Luo, and C.~He, ``An iteratively weighted {MMSE}
  approach to distributed sum-utility maximization for a {MIMO} interfering
  broadcast channel,'' \emph{IEEE Trans. Signal Process.}, vol.~59, no.~9, pp.
  4331--4340, Sept. 2011.

\bibitem{FPLinQ_shen2017}
K.~Shen and W.~Yu, ``F{P}{L}in{Q}: A cooperative spectrum sharing strategy for
  device-to-device communications,'' \emph{IEEE ISIT}, 2017.

\bibitem{DNN_sun2017}
H.~Sun, X.~Chen, Q.~Shi, M.~Hong, X.~Fu, and N.~D. Sidiropoulos, ``Learning to
  optimize: {Training} deep neural networks for wireless resource management,''
  \emph{IEEE SPAWC}, 2017.

\bibitem{REGNN2020}
M.~Eisen and A.~Ribeiro, ``Optimal wireless resource allocation with random
  edge graph neural networks,'' \emph{IEEE Trans. Signal Process.}, vol.~68,
  pp. 2977--2991, 2020.

\bibitem{LS2021Lee}
M.~Lee, G.~Yu, and G.~Y. Li, ``Graph embedding-based wireless link scheduling
  with few training samples,'' \emph{IEEE Trans. Wireless Commun.}, vol.~20,
  no.~4, pp. 2282--2294, Apr. 2021.

\bibitem{2021Shen}
Y.~Shen, Y.~Shi, J.~Zhang, and K.~B. Letaief, ``Graph neural networks for
  scalable radio resource management: Architecture design and theoretical
  analysis,'' \emph{IEEE J. Sel. Areas Commun.}, vol.~39, no.~1, pp. 101--115,
  Jan. 2021.

\bibitem{review2021}
S.~He, S.~Xiong, Y.~Ou, J.~Zhang, J.~Wang, Y.~Huang, and Y.~Zhang, ``An
  overview on the application of graph neural networks in wireless networks,''
  \emph{IEEE Open J. Commun. Soc.}, vol.~2, pp. 2547--2565, 2021.

\bibitem{LYsurvey}
M.~Lee, G.~Yu, H.~Dai, and G.~Li, ``Graph neural networks meet wireless
  communications: Motivation, applications, and future directions,'' \emph{IEEE
  Wireless Commun.}, vol.~29, no.~5, pp. 12--19, Oct. 2022.

\bibitem{Het_PA2022GJ}
J.~Guo and C.~Yang, ``Learning power allocation for multi-cell-multi-user
  systems with heterogeneous graph neural networks,'' \emph{IEEE Trans.
  Wireless Commun.}, vol.~21, no.~2, pp. 884--897, Feb. 2022.

\bibitem{GNN2019power}
X.~Keyulu, H.~Weihua, L.~Jure, and J.~Stefanie, ``How powerful are graph neural
  networks?'' \emph{ICLR}, 2019.

\bibitem{GNNsurvey_Wu2021}
Z.~Wu, S.~Pan, F.~Chen, G.~Long, C.~Zhang, and P.~S. Yu, ``A comprehensive
  survey on graph neural networks,'' \emph{IEEE Trans. Neural Netw. Learn
  Syst.}, vol.~32, no.~1, pp. 4--24, Jan. 2021.

\bibitem{CensNet2019}
X.~Jiang, P.~Ji, and S.~Li, ``{CensNet}: Convolution with edge-node switching
  in graph neural networks,'' \emph{IJCAI}, 2019.

\bibitem{NeuralPS2021}
J.~Jo, J.~Baek, S.~Lee, D.~Kim, M.~Kang, and S.~Hwang, ``Edge representation
  learning with hypergraphs,'' \emph{NeurIPS}, 2021.

\bibitem{skeleton2020}
X.~Zhang, C.~Xu, X.~Tian, and D.~Tao, ``Graph edge convolutional neural
  networks for skeleton-based action recognition,'' \emph{IEEE Trans. Neural
  Netw. Learn. Syst.}, vol.~31, no.~8, pp. 3047--3060, Aug. 2020.

\bibitem{IoT-supervised2022}
T.~Chen, X.~Zhang, M.~You, G.~Zheng, and S.~Lambotharan, ``A {GNN}-based
  supervised learning framework for resource allocation in wireless {IoT}
  networks,'' \emph{IEEE Internet Things J.}, vol.~9, no.~3, pp. 1712--1724,
  Feb. 2022.

\bibitem{GBLink2022}
S.~He, S.~Xiong, W.~Zhang, Y.~Yang, J.~Ren, and Y.~Huang, ``{GBLinks}:
  {GNN}-based beam selection and link activation for ultra-dense {D2D} mm{W}ave
  networks,'' \emph{IEEE Trans. Commun.}, vol.~70, no.~5, pp. 3451--3466, May.
  2022.

\bibitem{Het_PC/BF2021}
X.~Zhang, H.~Zhao, J.~Xiong, X.~Liu, L.~Zhou, and J.~Wei, ``Scalable power
  control/beamforming in heterogeneous wireless networks with graph neural
  networks,'' \emph{IEEE GLOBECOM}, 2021.

\bibitem{BF_RIS2021}
T.~Jiang, H.~V. Cheng, and W.~Yu, ``Learning to reflect and to beamform for
  intelligent reflecting surface with implicit channel estimation,'' \emph{IEEE
  J. Sel. Areas Commun.}, vol.~39, no.~7, pp. 1931--1945, Jul. 2021.

\bibitem{2022ZBC}
B.~Zhao, J.~Guo, and C.~Yang, ``Learning precoding policy: {CNN} or {GNN}?''
  \emph{IEEE WCNC}, 2022.

\bibitem{Bipartite2022}
J.~Kim, H.~Lee, S.-E. Hong, and S.-H. Park, ``A bipartite graph neural network
  approach for scalable beamforming optimization,'' \emph{IEEE Trans. Wireless
  Commun.}, vol.~22, no.~1, pp. 333--347, Jan. 2023.

\bibitem{LSJ2023MGNN}
S.~Liu, J.~Guo, and C.~Yang, ``Multidimensional graph neural networks for
  wireless communications,'' \emph{IEEE Trans. Wireless Commun.}, early
  access,2023.

\bibitem{Access2021}
V.~Ranasinghe, N.~Rajatheva, and M.~{Latva-aho}, ``Graph neural network based
  access point selection for cell-free massive {MIMO} systems,'' \emph{IEEE
  GLOBECOM}, 2021.

\bibitem{SCJ}
C.~Sun, J.~Wu, and C.~Yang, ``{Improving Learning Efficiency for Wireless
  Resource Allocation with Symmetric Prior},'' \emph{IEEE Wireless Commun.},
  vol.~29, no.~2, pp. 162--168, Apr. 2022.

\bibitem{expressive-GNN-Li2022}
P.~Li and J.~Leskovec, \emph{Graph Neural Networks: Foundations, Frontiers, and
  Applications}.\hskip 1em plus 0.5em minus 0.4em\relax Singapore: Springer
  Nature Singapore, 2022, ch. The Expressive Power of Graph Neural Networks,
  pp. 63--98.

\bibitem{survey2020expressive}
S.~Ryoma, ``A survey on the expressive power of graph neural networks,''
  \emph{arXiv:2003.04078}, 2020.

\bibitem{jegelka2022theory}
S.~Jegelka, ``Theory of graph neural networks: Representation and learning,''
  \emph{arXiv:2204.07697}, 2022.

\bibitem{zhang_heterogeneous_2019}
C.~Zhang, D.~Song, C.~Huang, A.~Swami, and N.~V. Chawla, ``Heterogeneous graph
  neural network,'' \emph{{SIGKDD}}, 2019.

\bibitem{Hypergraphs-S2021}
J.~Jaehyeong, B.~Jinheon, L.~Seul, K.~Dongki, W.~Minki, and H.~Sung, Ju, ``Edge
  representation learning with hypergraphs,'' \emph{NeurIPS}, 2021.

\end{thebibliography}
\end{document}